\documentclass[pdflatex,sn-mathphys-num,iicol]{sn-jnl}

\usepackage{graphicx}%
\usepackage{multirow}%
\usepackage{amsmath,amssymb,amsfonts}%
\usepackage{amsthm}%
\usepackage{mathrsfs}%
\usepackage[title]{appendix}%
\usepackage[table]{xcolor}
\usepackage{textcomp}%
\usepackage{manyfoot}%
\usepackage{booktabs}%
\usepackage{makecell}
\usepackage{algorithm}%
\usepackage{algorithmicx}%
\usepackage{algpseudocode}%
\usepackage{listings}%
\usepackage{cleveref}
\usepackage[utf8]{inputenc}
\usepackage[T1]{fontenc}
\usepackage{subcaption}

\theoremstyle{thmstyleone}%
\theoremstyle{thmstyletwo}%

\theoremstyle{thmstylethree}%

\newcommand{\revision}[1]{#1}

\usepackage{comment}
\newcommand\our{{SpaRRTa}}
\newcommand\ourfull{{Spatial Relation Recognition Task}}

\newcommand\cls{\texttt{[CLS]}}

\begin{document}

\title[Article Title]{\our{}: A Synthetic Benchmark for Evaluating Spatial Intelligence in Visual Foundation Models}


\author*[1,2]{\fnm{Turhan Can} \sur{Kargin}}\email{turhancan.kargin@doctoral.uj.edu.pl}

\author[1,4]{\fnm{Wojciech} \sur{Jasi\'nski}}

\author[1,2,6]{\fnm{Adam} \sur{Pardyl}}

\author[1,3]{\fnm{Bartosz} \sur{Zieli\'nski}}

\author[1,5]{\fnm{Marcin} \sur{Przewi\k{e}\.{z}likowski}}

\affil[1]{\orgdiv{Faculty of Mathematics and Computer Science}, \orgname{Jagiellonian University}, \orgaddress{\city{Krak\'ow},  \country{Poland}}}
\affil[2]{\orgdiv{Doctoral School of Exact and Natural Sciences}, \orgname{Jagiellonian University}, \orgaddress{\city{Krak\'ow}, \country{Poland}}}
\affil[3]{\orgname{Jagiellonian Center for Artificial Intelligence}, \orgaddress{\city{Krak\'ow},  \country{Poland}}}
\affil[4]{\orgname{AGH University of Krakow}, \orgaddress{\city{Krak\'ow},  \country{Poland}}}
\affil[5]{\orgname{NASK National Research Institute}, \orgaddress{\city{Warsaw},  \country{Poland}}}
\affil[6]{\orgname{IDEAS NCBR}, \orgaddress{\city{Warsaw},  \country{Poland}}}


\abstract{
Visual Foundation Models (VFMs), such as DINO and CLIP, excel in semantic understanding of images but
exhibit limited spatial reasoning capabilities,
which limits their applicability to embodied systems.
As a result, recent work has incorporated some 3D tasks (such as depth estimation) into VFM training. However, VFM performance remains inconsistent across other spatial tasks, raising the question of whether these models truly possess spatial awareness or overfit to specific 3D objectives.
To address this question, we introduce the \ourfull{} (\our{}) benchmark, which evaluates the ability of VFMs to identify \revision{static directional relations between visible objects}.
Unlike traditional 3D objectives that focus on precise metric prediction (e.g., surface normal estimation), \our{} probes a fundamental capability underpinning more advanced forms of human-like spatial understanding.
\our{} generates an arbitrary number of photorealistic images with diverse scenes and fully controllable object arrangements, along with freely accessible spatial annotations. Evaluating a range of state-of-the-art VFMs, we reveal significant disparities in their
\revision{ability to encode directional spatial relations.}
Through our analysis, we provide insights into the mechanisms that support or hinder \revision{this form of} spatial awareness in modern VFMs.
We hope that \our{} will serve as a useful tool for guiding the development of future spatially aware visual models. Our project page can be found at \href{https://sparrta.gmum.net/}{sparrta.gmum.net}
}


\maketitle

\begin{figure*}[!t]
    \centering
    \includegraphics[width=1\linewidth]{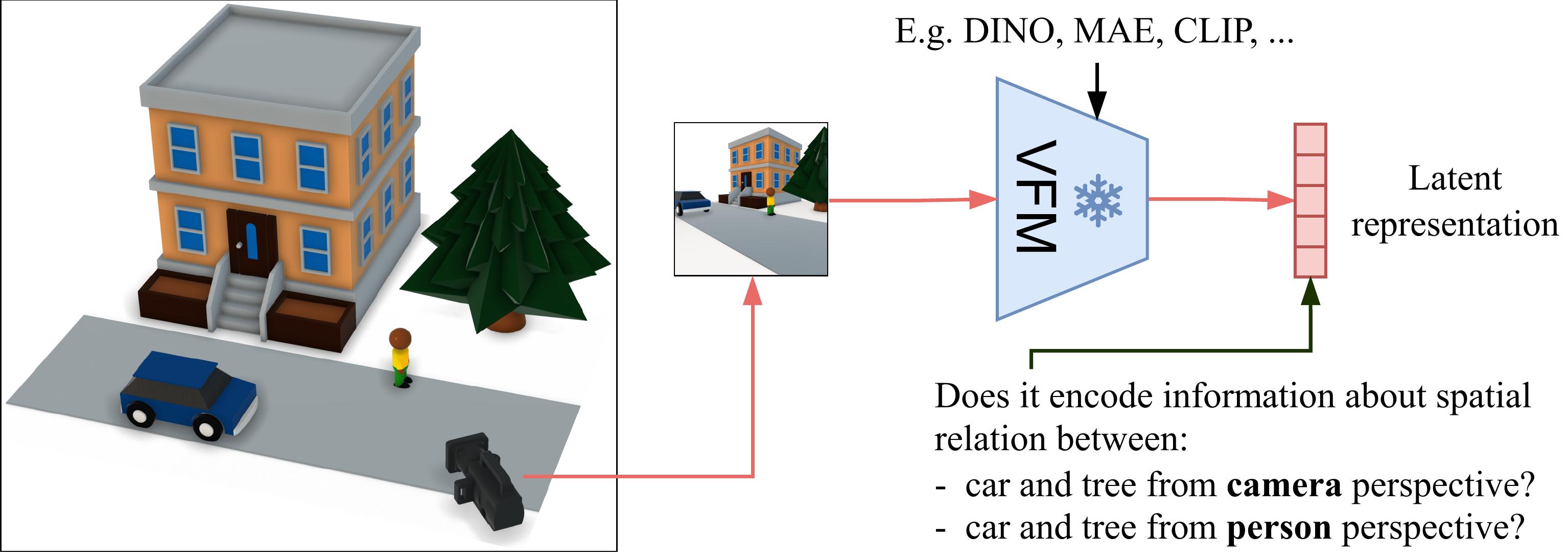}
    \smallskip
    \caption{
    \revision{A crucial component of spatial reasoning is the ability to encode spatial relations between visible objects.}
    To analyze such spatial awareness of Visual Foundation Models (VFMs), we introduce \ourfull{} (\our{}).
    We generate \revision{static} images showing different spatial layouts of several objects, encode them with VFMs, and probe whether their latent representations reliably encode the information about the spatial relations between the objects from a selected perspective.
    }

    \label{fig:initial_teaser}
\end{figure*}

\section{Introduction}

Visual Foundation Models (VFMs) such as DINO~\cite{caron2021emerging,oquab2023dinov2,siméoni2025dinov3} and CLIP~\cite{radford2021clip} demonstrate remarkable performance in semantic visual understanding.
Hence, they are widely used for semantic perception tasks including image classification~\cite{deng2009imagenet}, object recognition~\cite{tsung2014cocodataset}, multimodal learning~\cite{bai2025qwen2,li2024llavaonevisioneasyvisualtask}, and scene understanding, with applications in domains such as retail~\cite{srivastava2025hypervlm} and medical imaging~\cite{srokaoleksiak2025aidrivenrapididentificationbacterial}.
As these models move beyond \revision{passive} perception and are increasingly deployed in embodied and interactive settings, their role expands from recognizing visual content to supporting downstream decision-making and interaction~\cite{venkataramanan2024dora,bardes2024vjepa}.

However, effective deployment of VFMs in physical environments requires more than just semantic recognition.
To act and navigate, embodied agents must be aware of the spatial properties of their surroundings and reason about three-dimensional spatial relations between objects, such as their relative position and perspective (e.g., \texttt{the truck is to the left of the tree from the human's point of view})~\cite{RoboTHOR,linsley20243dpc}.
As a result, recent years have seen an increased interest in evaluating VFMs in terms of their performance in tasks that require spatial awareness, such as depth prediction, camera pose estimation, and surface normal estimation~\cite{banani_probing_2024,chen2024probingmidlevelvisioncapabilities}.
Moreover, recent VFM training objectives increasingly incorporate geometric cues~\cite{weinzaepfel2022croco,weinzaepfel2022crocov2,zhu_spa_2024,wang2025vggt}.
While this has yielded improvements in tasks most related to cues used in pretraining, the overall improvements in 3D perception remain inconsistent~\cite{banani_probing_2024,chen2024probingmidlevelvisioncapabilities}.
This raises a fundamental question: do contemporary VFMs acquire a general notion of spatial awareness, or do they instead overfit to specific, quantifiable geometric objectives without capturing abstractions that underpin human spatial understanding?


\revision{
To address this question, we introduce \ourfull{} (\our{}), a benchmark designed to evaluate relational spatial awareness in visual representations.
Rather than focusing on precise metric 3D predictions such as distances or surface normals, \our{} targets a controlled form of relational spatial understanding: recognizing the relative direction between objects from a specified viewpoint (see~\Cref{fig:initial_teaser}).
}
At a conceptual level, identifying relative object positions is a fundamental capability underlying more advanced forms of spatial reasoning~\cite{johnson_moore_2020,Jechura2006AnimalSC,BURGESS2002625}, providing a basis for detailed use cases like navigation, planning, and object manipulation.
In practice, \our{} evaluates how effectively spatial relations between given objects can be decoded from frozen VFM latent representations.
By decoupling \revision{relational spatial understanding} from precise geometric predictions, \our{} provides a principled way to assess whether VFMs represent transferable spatial structure rather than task-specific geometric shortcuts.


\begin{figure*}[!t]
    \centering
    \includegraphics[width=\textwidth]{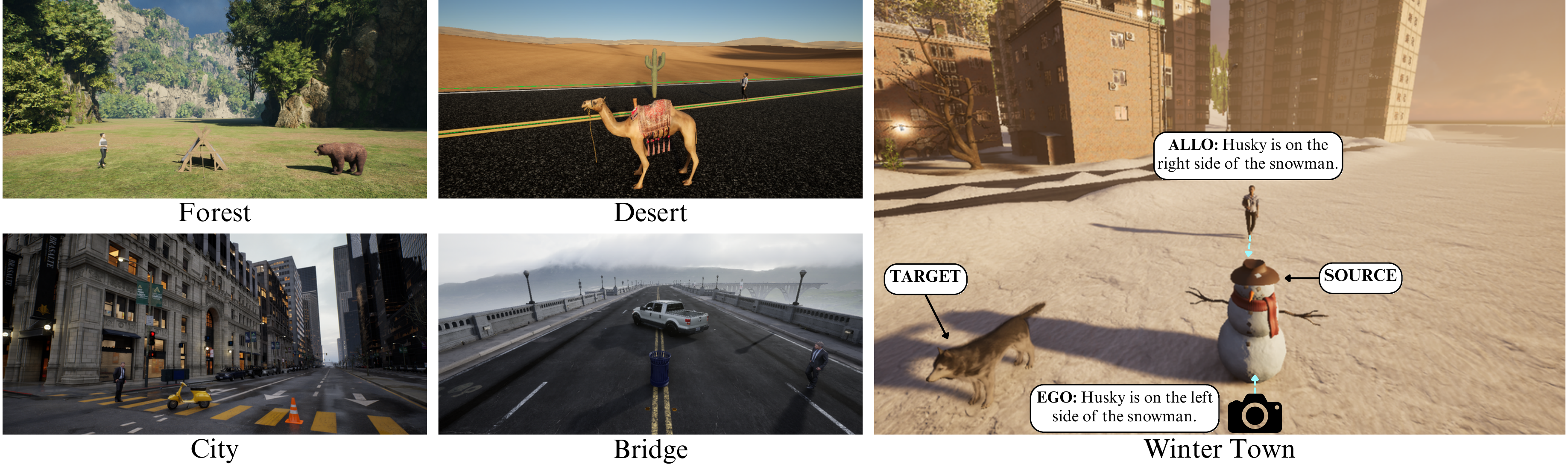}
    \caption{
    \our{} evaluates how effectively a VFM encodes spatial relations between objects in the image, either from the camera (\textbf{Ego}centric) or an arbitrary object's (\textbf{Allo}centric) perspective (see the right-most image).
    We use Unreal Engine 5 to produce photorealistic test images that are in-distribution for contemporary VFMs, using a variety of scene environments and object assets.
    }
    \label{fig:environments_table}
\end{figure*}

\our{} is built with Unreal Engine~5~\cite{unrealengine}, enabling the generation of photorealistic scenes with a wide range of object categories, layouts, and background contexts (see~\Cref{fig:environments_table}).
This synthetic setup provides full control over object placement and precise annotation of spatial relations, allowing evaluation at a scale and diversity that would be prohibitively expensive to obtain from real-world data.
By systematically varying object configurations and viewpoints across diverse environments, \our{}
enables a controlled and robust evaluation of spatial relation understanding in realistic visual settings, remaining in-distribution for common VFMs trained on natural images.

Using \our{}, we evaluate a broad set of VFMs trained under different supervision regimes and analyze how they encode spatial relations.
Our results show that self-supervised VFMs consistently outperform supervised and vision-language models, yet even the strongest models exhibit limitations in perspective-dependent and cluttered settings.
By comparing different probing mechanisms, we further reveal that spatial information is primarily stored at the patch level and is largely obscured by global pooling, highlighting the importance of structured aggregation for uncovering spatial knowledge.
We hope that \our{} will serve as a useful diagnostic tool for studying spatial representations and for guiding the development of future models with stronger spatial awareness, which is essential for embodied and interactive applications.

Our contributions can be summarized as follows:
\begin{itemize}
        \item \textbf{The Tool:} We introduce \our{}, a photorealistic, synthetically generated evaluation environment built on Unreal Engine 5 for probing
        \revision{directional spatial relation recognition in VFMs.}
        \item \textbf{The Benchmark:} We define two standardized challenge sets, \our{}\textbf{-ego} (camera-centric) and \our{}\textbf{-allo} (allocentric/perspective-taking), designed to disentangle spatial logic from semantic recognition.
        \item \textbf{The Insights:} We benchmark varying families of VFMs (MIM, Joint-Embedding, Supervised, Vision-Language) and characterize which models provide spatial information in their representations and how these representations are structured.
\end{itemize}

\section{Related Work}

\paragraph{Spatial awareness in visual representation learning.}
Mainstream self-supervised vision transformers, including DINO~\cite{caron2021emerging, oquab2023dinov2,siméoni2025dinov3}, MAE~\cite{he2021masked}, and I-JEPA~\cite{assran2023ijepa}, are trained almost exclusively on 2D imagery. They achieve high accuracy on high-level perception tasks, but they can lag behind on depth and multiview geometry when probed or fine-tuned~\cite{banani_probing_2024, chen2024probingmidlevelvisioncapabilities}.
To bridge this gap, several recent variants inject explicit 3D cues during pre-training:
MultiMAE~\cite{bachmann_multimae_2022} and EmbodiedMAE~\cite{dong_embodiedmae_2025} reconstruct heterogeneous RGB, depth, and semantic segmentation maps;
CroCo enforces cross-view completion~\cite{weinzaepfel2022croco, weinzaepfel2022crocov2};
SPA~\cite{zhu_spa_2024} trains with differentiable novel-view rendering;
Finally, V-JEPA~\cite{bardes2024vjepa} learns to predict future image representations conditioned on masked video frames. In contrast to these self-supervised approaches, VGGT~\cite{wang2025vggt} adopts a fully supervised paradigm by training a large-scale transformer on a diverse collection of 3D-annotated datasets (e.g., Co3D\cite{reizenstein_2021_ICCV}, MegaDepth\cite{li_2018_CVPR}). This approach forces the encoder to internalize a precise geometric structure by directly regressing camera parameters, depth maps, and point tracks.
A parallel line of work develops feed-forward 3D vision models that explicitly recover scene geometry from images. DUSt3R~\cite{Wang_2024_CVPR} introduced calibration- and pose-free pointmap regression, followed by MASt3R~\cite{leroy2024grounding}, which adds dense 3D-aware features for image matching. Spann3R~\cite{spann3r_wang} and CUT3R~\cite{wang2025continuous} further incorporate spatial memory and persistent state for globally consistent and continuous reconstruction, while MUSt3R~\cite{Cabon_2025_CVPR} extends the approach to multiple views and large image collections. RayZer~\cite{jiang2025rayzerselfsupervisedlargeview} explores self-supervised 3D representation learning from unposed and uncalibrated images through view synthesis. These methods primarily target explicit geometric reconstruction, camera estimation, or view synthesis.
While results still vary across tasks, these methods generally improve depth, pose, and policy-transfer benchmarks~\cite{majumdar_where_2023}, suggesting that geometric and temporal cues are important inductive biases for the next generation of foundation vision encoders.
In this work, we introduce a measure of spatial awareness that is more direct than standard metric evaluations (e.g., depth or pose estimation) and conduct a systematic evaluation of the above encoders, providing broader insights about their representations.

\paragraph{Benchmarks for 3D awareness.}
3D awareness is usually measured through frozen-encoder probe tasks or task-driven control suites.
Probe3D~\citep{banani_probing_2024} adds decoders for depth, surface normals, and cross-view matching, while the Mid-Level Vision Probe~\citep{chen2024probingmidlevelvisioncapabilities} extends this with object segmentation and image similarity.
Together, these studies show that higher ImageNet accuracy does not guarantee stronger 3D reasoning and that performance correlates only moderately between probes, highlighting the need for \emph{task-agnostic} diagnostics.
Task-driven suites such as VC-1~\citep{majumdar_where_2023} and SPA~\citep{zhu_spa_2024} utilize the same encoders in navigation and object-search policies, measuring downstream success and thereby tying representation quality to action.
Complementing these, several intuitive-physics benchmarks~\citep{riochet2018intphys, bear2021physion, bordes2025intphys2} test whether a model predicts physically plausible outcomes, an ability adjacent to spatial awareness and vital for embodied agents.
Unlike the task-driven or physics-focused evaluations above, our benchmark targets representational 3D knowledge, specifically evaluating
\revision{a controlled form of abstract relational spatial reasoning.}

\paragraph{Synthetic Environments for Benchmarking.}
Synthetic datasets and simulators provide dense, perfectly aligned ground-truth labels that are hard to capture in the wild.
For scene-scale perception, RGB-D reconstructions such as ScanNet~\cite{dai2017scannet} offer thousands of annotated indoor rooms, though their visual diversity remains limited.
Interactive robot benchmarks, including Meta-World~\cite{yu2020metaworld} and Adroit~\cite{rajeswaran2018adroit} built on MuJoCo~\cite{todorov2012mujoco}, and RLBench~\cite{james2020rlbench} built on CoppeliaSim~\cite{rohmer2013vrep}, target control dynamics rather than appearance and therefore lack photorealism.
At the opposite end of this spectrum, Habitat-Sim~\cite{habitat19iccv} renders Matterport3D-style scans~\cite{Matterport3D} at near-photorealistic quality and has become a widely adopted platform for embodied navigation evaluation.
Extending this direction to large-scale outdoor settings, recent works such as FlySearch~\cite{pardyl2025flysearchexploringvisionlanguagemodels} make use of an environment based on Unreal Engine 5~\cite{unrealengine} to evaluate vision-language models on highly photorealistic synthetic data. In this paper, we explore the use of such environments for evaluating 3D awareness in visual representations.

\paragraph{Probing mechanisms for Vision Transformers.}
Standard evaluation of Vision Transformer (ViT)-based Visual Foundation Models (VFMs) typically relies on linear probing (LP) trained on global descriptors, such as the \cls{} token or the Global Average Pooling (GAP) of patch embeddings~\cite{he2020momentum,he2021masked,przewiezlikowski2025cls}. This avoids the high computational cost of full fine-tuning and, more importantly, allows us to explore the full potential of the pre-trained representations rather than the model's adaptability.
Recent works have introduced selective pooling techniques that recover local information lost by global pooling, such as Attention-Based Multiple Instance Learning Pooling (AbMILP)~\cite{ilse2018attention,rymarczyk2021kernel, przewiezlikowski2025cls}, efficient probing~\cite{psomas2025attention}, and SimPool~\cite{psomas2023simpool}. The AbMILP method~\cite{ilse2018attention} serves as a selective aggregation mechanism, learning a lightweight regressor to assign scalar importance weights to each patch, effectively filtering out background noise and focusing on relevant regions~\cite{przewiezlikowski2025cls}. Unlike simple pooling, efficient probing learns a set of distinct query vectors that attend directly to the input feature space, allowing the probe to generate diverse and complementary attention maps~\cite{psomas2023simpool,psomas2025attention}. For instance, attention maps can separately attend to different objects or different parts of the same object.
While the above probing techniques have proved their effectiveness in high-level recognition tasks, we use them to solve fundamental spatial recognition tasks. This allows us to rigorously evaluate the intrinsic spatial awareness of existing VFMs and determine whether spatial information is present in their global or local latent space.

\section{Methodology for Evaluating Spatial Awareness}
\label{sec:methodology}

\begin{figure*}[!t]
    \begin{center}
        \includegraphics[width=\textwidth]{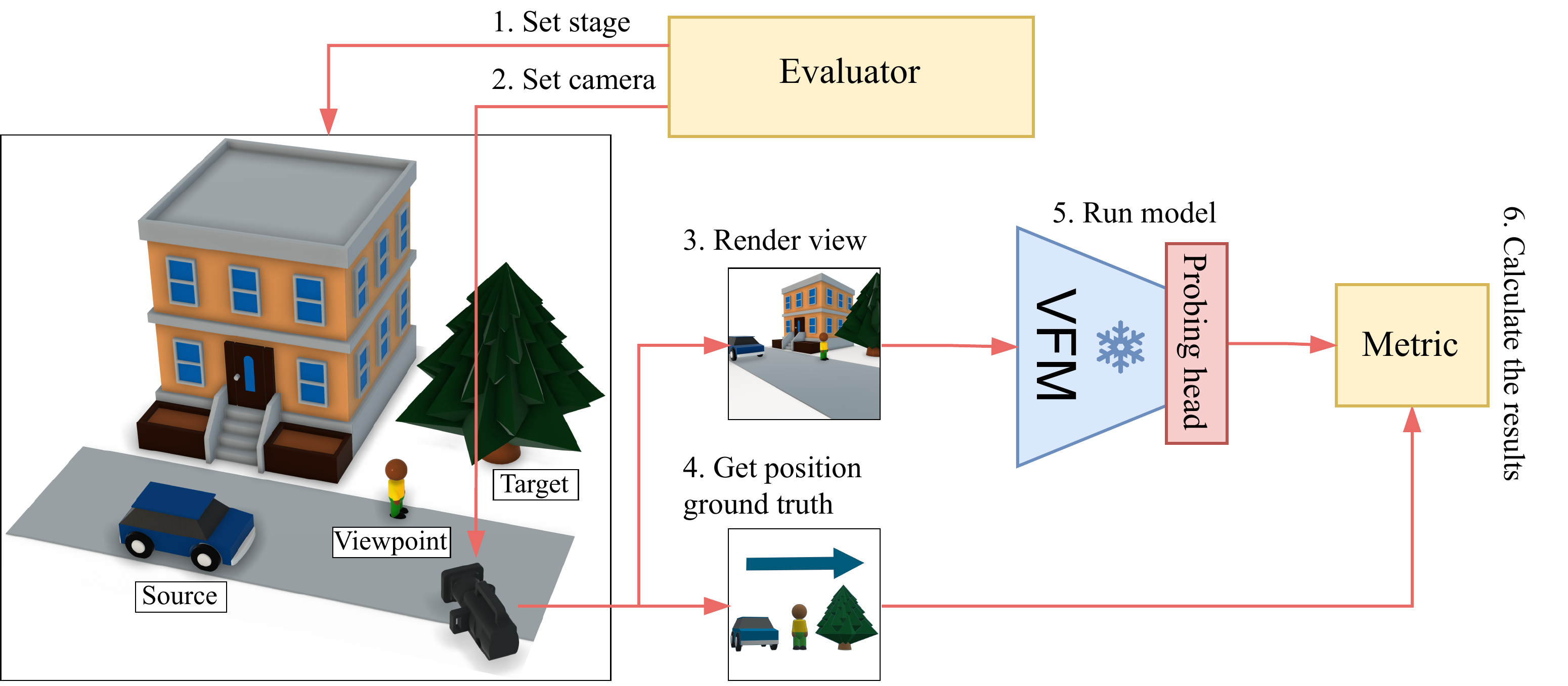}
    \end{center}
    \caption{\textbf{Evaluation pipeline:} The \our{} evaluation setup consists of a data-generation system built on Unreal Engine 5. The pipeline begins with an evaluation controller that sets the scene in a precisely controlled manner using a diverse set of realistic assets~\textbf{(1)} and places the camera~\textbf{(2)} for capture. Next, a near-photorealistic image is rendered by Unreal~\textbf{(3)}, and ground-truth positional information is acquired~\textbf{(4)}. The image is then passed through the evaluated VFM, and the resulting representation is fed to the probing head to predict the spatial relation~\textbf{(5)}. Finally, the accuracy is computed~\textbf{(6)}.}
    \label{fig:pipeline}
\end{figure*}

In this section, we introduce our \our{} methodology for evaluating the abstract spatial awareness of visual models.
\revision{From a high-level perspective, \our{} measures the accuracy of recovering the relative directional configuration of two objects in a static rendered image based on an image representation.}
We utilize a two-stage evaluation framework comprising synthetic dataset generation and offline model probing.
The entire pipeline, illustrated in~\Cref{fig:pipeline}, and detailed below, follows a structured protocol:

\textbf{Synthetic Data Generation.} The process begins with the experimental design defined by the evaluator.
\begin{enumerate}
    \item \textbf{Set Stage:} The evaluator establishes the scene configuration by selecting a specific environment and assets and randomly positioning the source, target, and viewpoint objects (see \Cref{subsec:assets} for more details on the 3D assets used).
    \item \textbf{Set Camera:} A viewpoint is selected to capture the scene, ensuring that the necessary spatial context is visible from the sensor's perspective. At this stage, the geometric relationship is validated to filter out ambiguous configurations (e.g., where the target lies on a decision boundary), ensuring that only clear cases remain in the dataset (see~\Cref{subsec:data_gen_pipeline} for a detailed description of data generation).
    \item \textbf{Render View:} The Unreal Engine rendering pipeline synthesizes a high-fidelity 2D RGB image of the configured scene.
    \item \textbf{Get Ground Truth Position:}
    Based on the exact positions of the objects and a given variant of the task, the image is annotated with the spatial relation to be predicted (see~\Cref{subsec:task_description}).
\end{enumerate}
\textbf{Representation Probing.} Once the data are generated, we proceed to the evaluation of the visual models.
\begin{enumerate}
    \setcounter{enumi}{4}
    \item \textbf{Run Model:} The images are fed into the VFM. Crucially, the VFM backbone remains frozen to assess its intrinsic pre-trained representations without further weight adaptation. The extracted features are then passed to a lightweight probing head, which predicts the discrete spatial relation (we detail the benchmarked VFMs and probing techniques in~\Cref{sec:exp_setup}).
    \item \textbf{Calculate Results:} Finally, the model's predicted spatial relation class is compared to the ground-truth geometric label derived in Step 4 to compute the classification accuracy.
\end{enumerate}

\begin{figure*}[h]
    \centering
    \begin{subfigure}[t]{0.48\textwidth}
        \includegraphics[width=\linewidth]{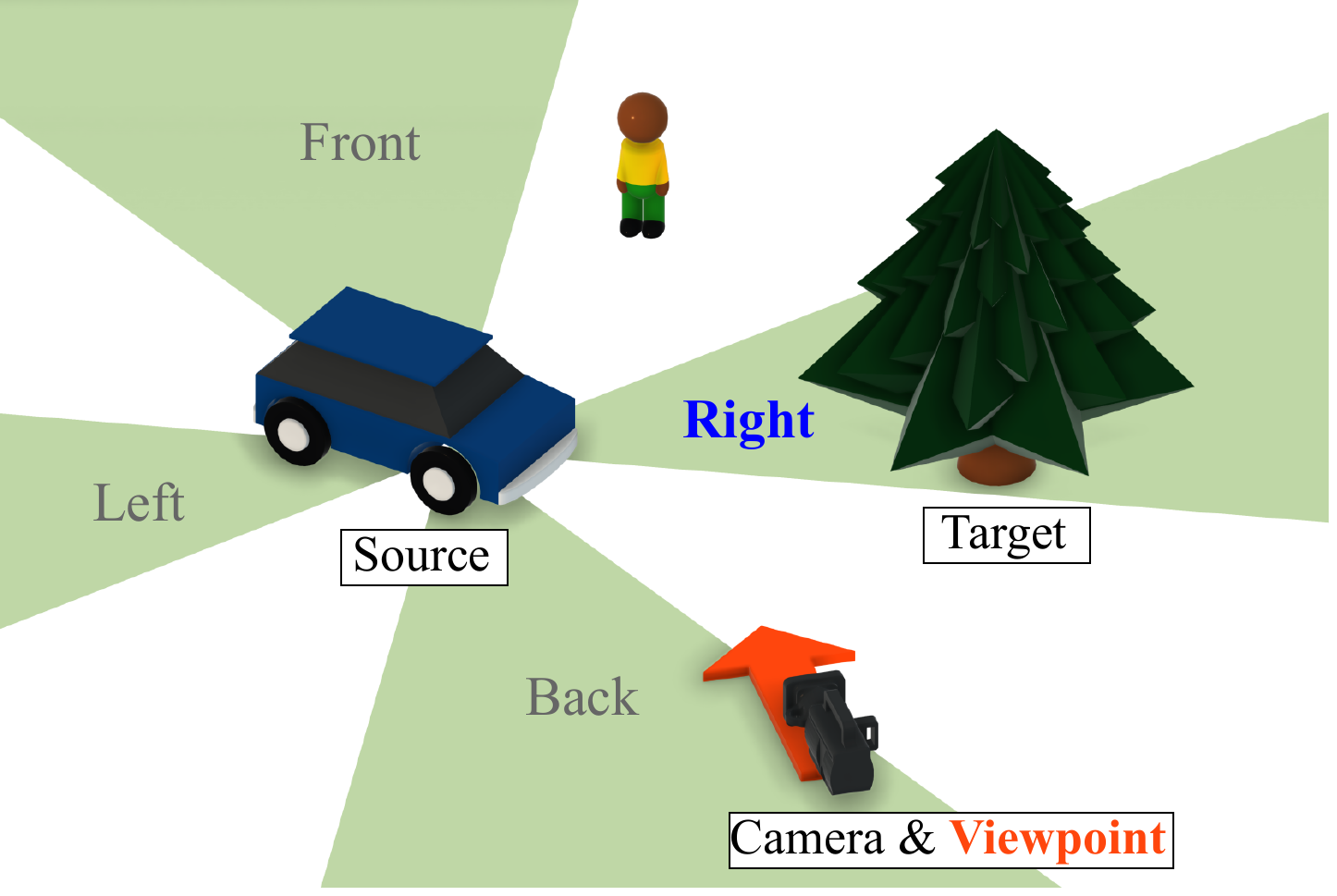}
        \caption{Egocentric \ourfull{} (\textbf{\our{}-ego}). In this variant, the camera position defines the viewpoint reference for the spatial directions. The correct relative direction from the \textbf{car} (source) to the \textbf{tree} (target) is \textbf{right}.
        }
        \label{fig:ego_example}
    \end{subfigure}
    \hfill
    \begin{subfigure}[t]{0.48\textwidth}
        \includegraphics[width=\linewidth]{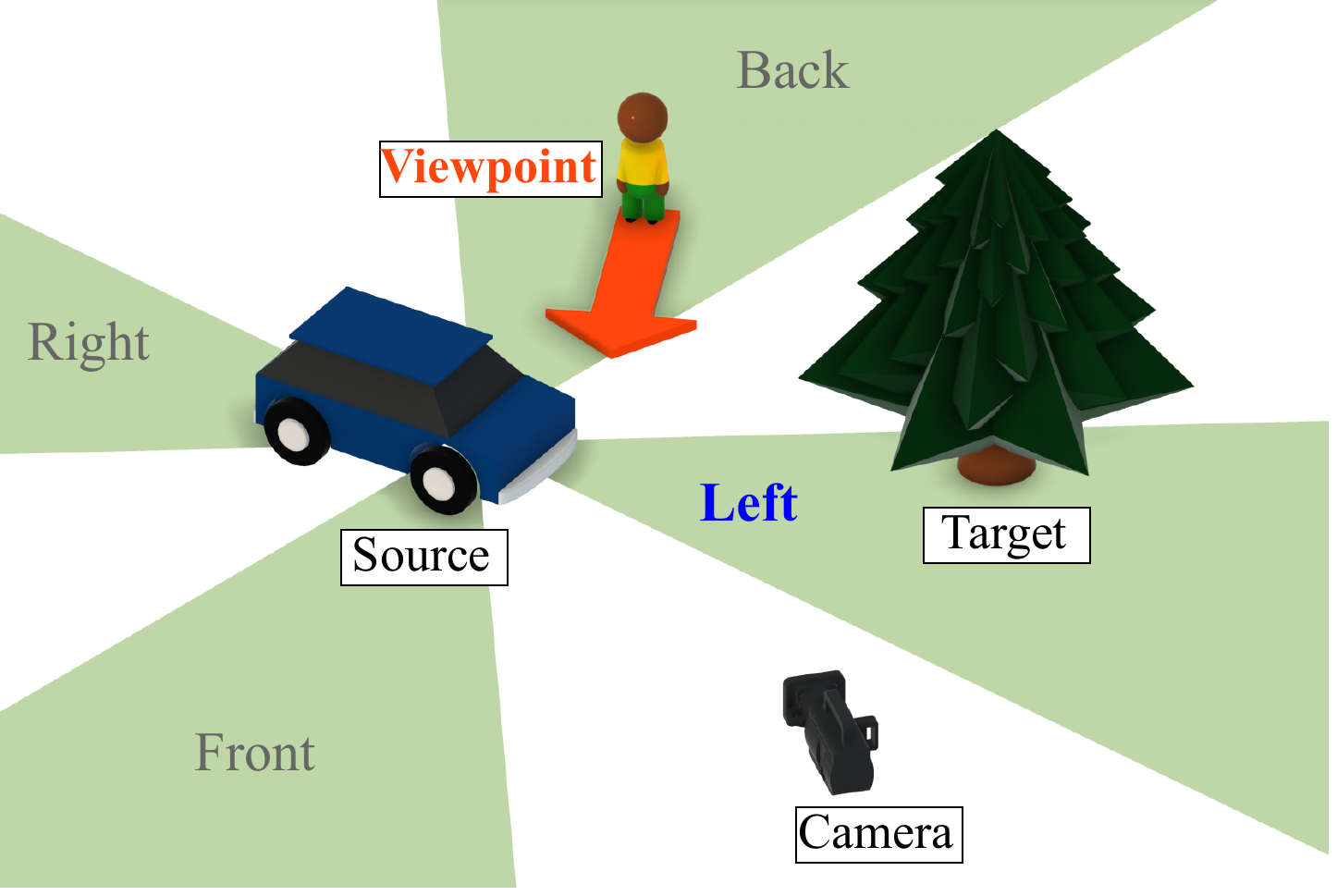}
    \caption{Allocentric \ourfull{} (\textbf{\our{}-allo}). In this variant, the spatial directions from the reference are defined by the \textbf{viewpoint} object (human).
        The correct relative direction from the \textbf{car} (source) to the \textbf{tree} (target) is \textbf{left}.
        }
        \label{fig:allo_example}
    \end{subfigure}
    \caption{Visualization of the \ourfull{} (\our{}). The task is to predict the spatial relationship between the \textbf{source} (i.e., car) and \textbf{target} (i.e., tree) objects with respect to a given \textbf{viewpoint} (i.e., the camera in the egocentric variant \textbf{(a)} and the human in the allocentric variant \textbf{(b)}). To eliminate ambiguity, we generate scenes in which the spatial relationships between objects are clearly recognizable; the green area denotes valid locations for target placement.
    }
    \label{fig:ego_allo_example}
\end{figure*}

\subsection{\ourfull{}}
\label{subsec:task_description}

To evaluate spatial awareness of visual models, we define a family of tasks where the objective is to determine the relative spatial relation between objects present in a single image.
Given two objects, denoted as \textbf{source} and \textbf{target}, the task is to classify the direction from the source to the target.
We formulate the prediction as a four-way classification problem with discrete labels, i.e. \texttt{\{left, right, front, back\}}.
A key aspect of spatial direction is that it is not defined in absolute image coordinates, but with respect to a chosen reference \textbf{viewpoint}. In this work, we define a viewpoint as the observer that determines how spatial relations are perceived. We formulate the \textbf{egocentric} and \textbf{allocentric} variants of the task that differ in the viewpoint used to express the direction. We visualize both variants in~\Cref{fig:ego_allo_example}, and describe them below.

\textbf{Egocentric variant (\Cref{fig:ego_example})} defines the camera as the viewpoint.
The question is therefore where the target lies relative to the source as seen from the camera.
This variant captures spatial relations directly observable from the input image.

\textbf{Allocentric variant (\Cref{fig:allo_example})} defines the selected third object (in our case, the human) as a viewpoint. In this scenario, the spatial relation observed by the camera differs from the relation observed by the viewpoint object. 
Consequently, the allocentric variant is more challenging, as it requires the model to decouple the spatial relation from its own visual input. This demands an implicit viewpoint transformation known as perspective-taking. The model must ignore the apparent positions of objects in the image and infer their geometric arrangement relative to the third object's perspective.

\subsection{Data generation pipeline}
\label{subsec:data_gen_pipeline}
In contrast to standard vision benchmarks such as ImageNet~\cite{deng2009imagenet}, which rely on images gathered from the Internet, \our{} uses a modern rendering engine to create near-photorealistic images for evaluation. This allows us to retain full control over image content while remaining in-distribution for the common VFMs.

\textbf{Renderer.} As the rendering engine, we chose Unreal Engine~5~\cite{unrealengine}, which provides state-of-the-art rendering quality and performance through real-time ray tracing with dynamic global illumination and reflections, as well as automatically adjusted levels of detail. The engine is highly extensible, allowing us to adapt it for this study and provide interfaces for automated environment-map creation, image capture, and ground-truth acquisition. Moreover, Unreal Engine's vast online marketplace of free graphical assets, many of which were created using photogrammetry, provides a rich source of diverse evaluation scenarios while ensuring high-fidelity, realistic results. Finally, it requires relatively modest hardware, such as a consumer-grade GPU, while supporting deep-learning workloads through the Vulkan API.

\textbf{Scenario generator.} The scenario generation part of the pipeline is responsible for procedural generation of diverse testing environments using the official Unreal Engine Editor Python API~\cite{unreal_python_api}. First, it randomly selects test objects from a map-specific asset list and places them on the ground at positions sampled from a Gaussian distribution. Next, the camera position is sampled from a uniform distribution over an area surrounding the map center and oriented toward the placed objects. Finally, the system verifies that all assets are correctly loaded and that the scene is ready for image capture. Additional implementation details are provided in~\Cref{subsec:render_details}.

\textbf{Geometric ambiguity control.} A fundamental challenge in spatial classification lies in defining precise boundaries between semantic classes (e.g., determining the exact threshold where \textit{Front} transitions to \textit{Left}). To eliminate label noise and ensure mathematical rigor, we implement a strict procedural rejection sampling strategy. We define ambiguity zones as $\pm 15^\circ$ angular regions centered along the diagonals ($45^\circ, 135^\circ, 225^\circ, 315^\circ$) relative to the viewpoint's forward direction. These diagonals serve as the dividing lines between the four main directions. The pipeline computes the precise angular relationship between the target and source objects, and automatically discards any scenario in which the target falls within these ambiguity zones. This guarantees that all retained samples possess indisputable ground-truth labels, ensuring that evaluation metrics reflect genuine spatial
\revision{relation recognition}
capabilities rather than boundary confusion. We provide a comprehensive description and visualization of these exclusion zones in~\Cref{subsec:design_principles}.

\textbf{Image and ground-truth capture.} View capture is performed using the adapted UnrealCV library~\cite{qiu2017unrealcv}. We capture both the ray-traced RGB image and a ground-truth segmentation mask. The mask is used together with the 3D coordinates of spawned objects to assert the validity of the scenario generation and calculate the ground-truth label for the view (Front, Back, Left, or Right). All captured images and metadata are stored for model evaluation. Detailed statistics on the dataset size are provided in~\Cref{subsec:design_principles}.

\begin{figure*}[ht]
    \centering
    \includegraphics[width=0.925\linewidth]{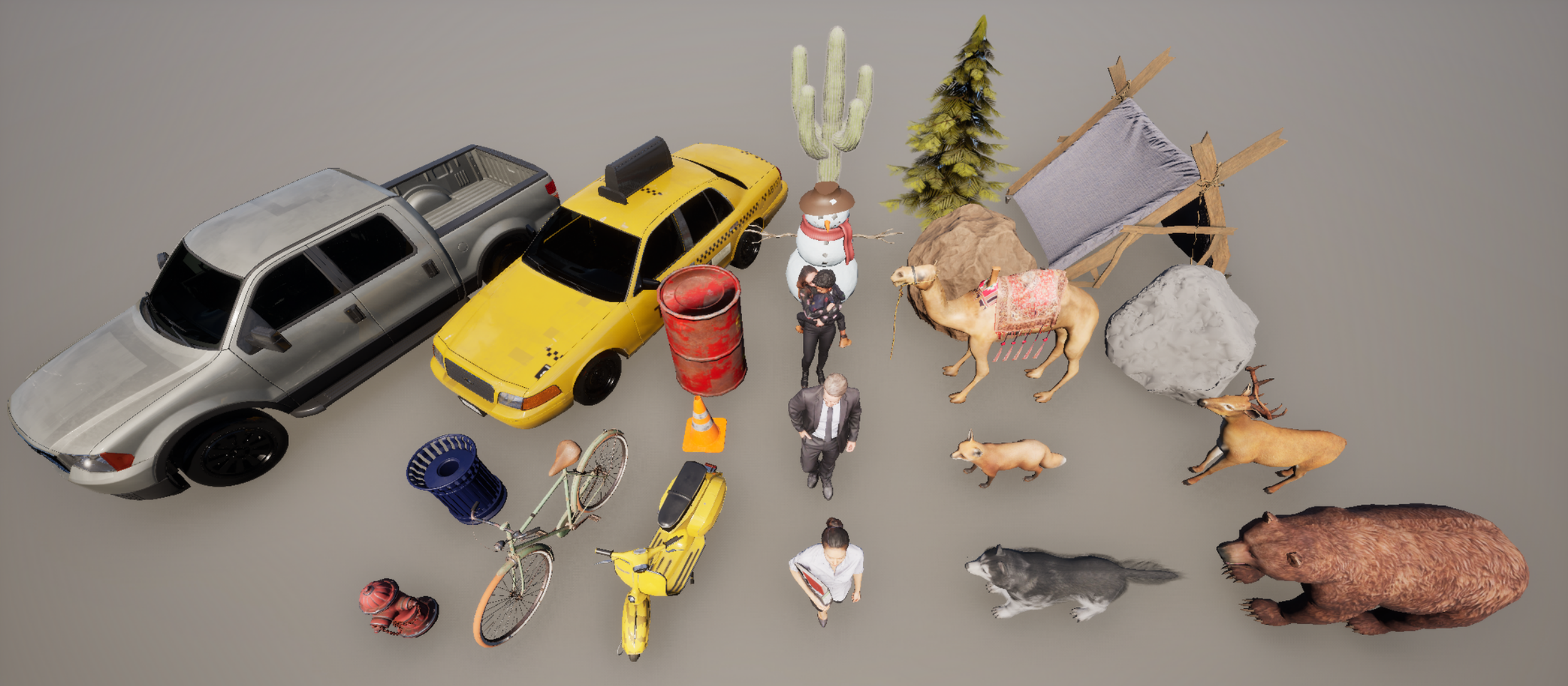}
    \caption{\textbf{\our{} Asset Library.}
    \our{} constructs test examples using a curated set of diverse, high-fidelity 3D assets selected based on common ImageNet classes~\cite{deng2009imagenet}.
    }
    \label{fig:assets}
\end{figure*}

\subsection{Evaluation Environments}
\label{subsec:assets}

To ensure that \our{} measures robust spatial
\revision{relation recognition}
rather than overfitting to specific visual domains, we construct a diverse suite of five distinct high-fidelity environments (see~\Cref{fig:environments_table}). These environments range from organic, unstructured landscapes to dense, structured urban settings, challenging the model to generalize its geometric understanding across vastly different textures, lighting conditions, and contextual layouts.

\textbf{Asset Classes.} To minimize semantic ambiguity and prioritize spatial
\revision{relation recognition}
performance, we used a fixed set of distinguishable objects for each environment.
Each scene includes a human object that serves as the consistent viewpoint for the allocentric tasks. The source and target objects are selected to be semantically coherent with their respective environments (see~\Cref{fig:environments_table}). This selection ensures that the objects occur naturally within the scene context across all evaluation trials. Crucially, our asset selection strategy prioritized objects that are not only natural within their respective scene contexts (e.g., camels in a desert and taxis in a city) but are also statistically prominent in standard pre-training datasets like ImageNet~\cite{ridnik2021imagenet21k}.
This way, we ensure that any failure in spatial
\revision{relation recognition}
is due to geometric understanding rather than a lack of semantic recognition by selecting assets that align with common ImageNet super-categories: (1) \textit{Animals}, (2) \textit{Everyday Objects}, (3) \textit{Nature}, and (4) \textit{Humans}. We provide a catalog of all 3D assets, including their sources, in~\Cref{tab:asset_classes} and~\Cref{fig:assets}.

\textbf{Environmental diversity.} We leverage high-quality environments from the Unreal Engine ecosystem to create photorealistic scenes that mimic real-world complexity. The benchmark consists of five distinct environment types.
\begin{itemize}
    \item \textbf{Forest environment} is based on the Electric Dreams Environment Unreal Engine product demo~\cite{electric_dreams}. This scene depicts a sparse forest landscape with complex foliage, uneven terrain, and natural rock formations.
    \item \textbf{Desert environment} is a vast, arid landscape~\cite{desert_env} characterized by open terrain, sand dunes, and high-contrast lighting. This environment is very sparse and texture-homogeneous compared to other environments.
    \item \textbf{Winter Town environment} is a snow-covered setting reflecting a typical small Eastern European town~\cite{winter_town}. This environment consists of cold lighting, occluding snow textures, and a small number of village-type buildings.
    \item \textbf{Bridge environment} is a valley scene~\cite{bridge_env} centered around a large bridge infrastructure.
    \item \textbf{City environment} is a large-scale, modern American metropolis derived from the City Sample demo~\cite{city_sample}, featuring high-rise architecture, paved roads, and complex urban geometry.
\end{itemize}

\subsection{Implementation details}\label{subsec:implementation_details}

To ensure the reproducibility of our results and the accessibility of the \our{} benchmark to the wider research community, we use standard, open-source libraries and consumer-grade hardware for all experiments.

\textbf{Simulation and Rendering Stack.} The synthetic environment generation is built on Unreal Engine 5.5, leveraging its native Python API~\cite{unreal_python_api} for scene manipulation and automated asset placement. We employ the UnrealCV plugin~\cite{qiu2017unrealcv} to capture synchronized RGB images and segmentation masks. The rendering pipeline runs on a standard Windows workstation equipped with two NVIDIA RTX 2080 Ti GPUs (11 GB of VRAM each), generating high-fidelity data without the need for
enterprise-grade compute clusters.

\textbf{Probing and Evaluation Stack.} All probing experiments are conducted in a Linux environment using an NVIDIA RTX 4090 (24 GB of VRAM). We follow the established protocols for probing frozen features, specifically the methodology outlined in recent mid-level vision benchmarks~\cite{chen2024probingmidlevelvisioncapabilities}. The evaluation framework is implemented in PyTorch, with pre-trained VFMs sourced directly from the official HuggingFace Transformers or timm libraries to ensure standardized weight initialization. The full codebase, including the Unreal Engine scene generation scripts and probing frameworks, will be made publicly available upon publication.

\begin{table*}[ht]
\centering
\small
\setlength{\tabcolsep}{6pt}

\begin{tabular}{l p{4cm} l p{4.2cm}}
\toprule
\textbf{Model} & \textbf{Pre-training Objective} & \textbf{Supervision} & \textbf{Training Dataset} \\
\midrule

\multicolumn{4}{l}{\textit{\textbf{Joint-Embedding (JEA)}}} \\
DINO\cite{caron2021emerging} & Contrastive / Distillation & Self-Sup. & ImageNet-1k \\
DINO-v2$^{\dagger}$\cite{oquab2023dinov2} & DINO + iBOT & Self-Sup. & LVD-142M \\
DINO-v2 reg$^{\ddagger}$\cite{darcet_vision_2023} & DINO-v2 w/ Register Tokens & Self-Sup. & LVD-142M \\
DINOv3\cite{siméoni2025dinov3} & DINO + iBOT & Self-Sup. & LVD-1689M \\

\midrule

\multicolumn{4}{l}{\textit{\textbf{Masked Image Modeling (MIM)}}} \\
MAE\cite{he2021masked} & Pixel Reconstruction & Self-Sup. & ImageNet-1k \\
MaskFeat\cite{wei2023masked} & HOG Feature Prediction & Self-Sup. & ImageNet-1k \\
SPA\cite{zhu_spa_2024} & Masked Volumetric Neural Rendering & Multi-View Self-Sup. & ScanNet, Hypersim, S3DIS \\
CroCo\cite{weinzaepfel2022croco} & Cross-View Completion & Self-Sup. & Habitat \\
CroCov2\cite{weinzaepfel2022crocov2} & Cross-View Completion & Self-Sup. & ARKitScenes, MegaDepth, ... \\

\midrule

\multicolumn{4}{l}{\textit{\textbf{Supervised \& Weakly Supervised}}} \\
VGGT$^{*}$\cite{wang2025vggt} & Multi-Task 3D Regression (Cam, Depth, Tracks) & 3D Sup. & Co3D, MegaDepth, etc. \\
DeiT\cite{touvron2022deit} & Classification + Distillation & Label Sup. & ImageNet-1k \\
CLIP\cite{radford2021clip} & Image-Text Contrastive & Text-Image Pairs & Web Image-Text (WIT) \\

\bottomrule
\end{tabular}

\vspace{4pt}
\footnotesize
\raggedright
\textit{Note: All models utilize a ViT-B/16 backbone unless otherwise marked:} \\
$^{\dagger}$ViT-B/14 \quad $^{\ddagger}$ViT-B/14 \& ViT-L/14 \quad $^{*}$ViT-L/14

\caption{\textbf{Evaluated Visual Foundation Models (VFMs).} We consider a range of visual backbones, specifically focusing on Self-supervised methods spanning various forms of supervision (JEA, MIM), as well as prominent supervised models.}
\label{tab:models}
\end{table*}

\section{Experiments}

In this section, we describe our evaluation of the current prominent Visual Foundation Models (VFMs) on the \ourfull{} (\our{}). \Cref{sec:exp_setup} details the choice of VFMs, as well as the methodology of adapting them for solving \our{}.
In \Cref{sec:exp_main_results}, we conduct the evaluation of VFMs on \our{}, revealing which models encode spatial relations.
In \Cref{sec:exp_correlation}, we compare VFM performance on \our{} with performance on other spatial-awareness and high-level image-recognition tasks to understand the relationship among these capabilities.
Finally, in~\Cref{sec:analysis_attention}, we analyze the inner representations of VFMs to better understand how
\revision{representations useful for spatial relation recognition}
are formed and where they are expressed.

\subsection{Experimental setup}
\label{sec:exp_setup}

\begin{figure*}[!ht]
    \centering
    \begin{subfigure}[b]{0.48\textwidth}
        \centering
        \includegraphics[width=\linewidth]{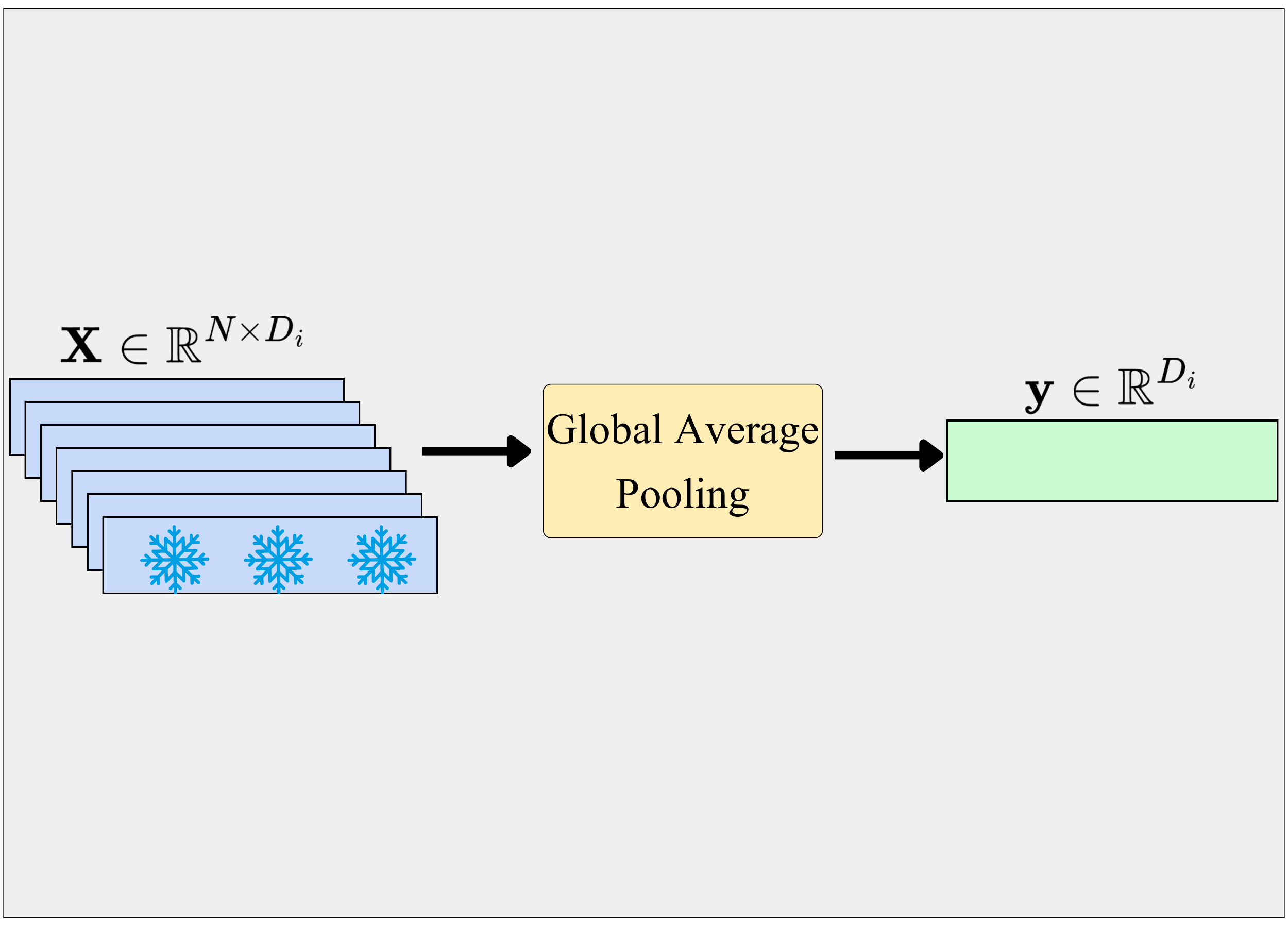}
        \caption{Linear Probing with GAP}
        \label{fig:probing_linear}
    \end{subfigure}
    \begin{subfigure}[b]{0.48\textwidth}
        \centering
        \includegraphics[width=\linewidth]{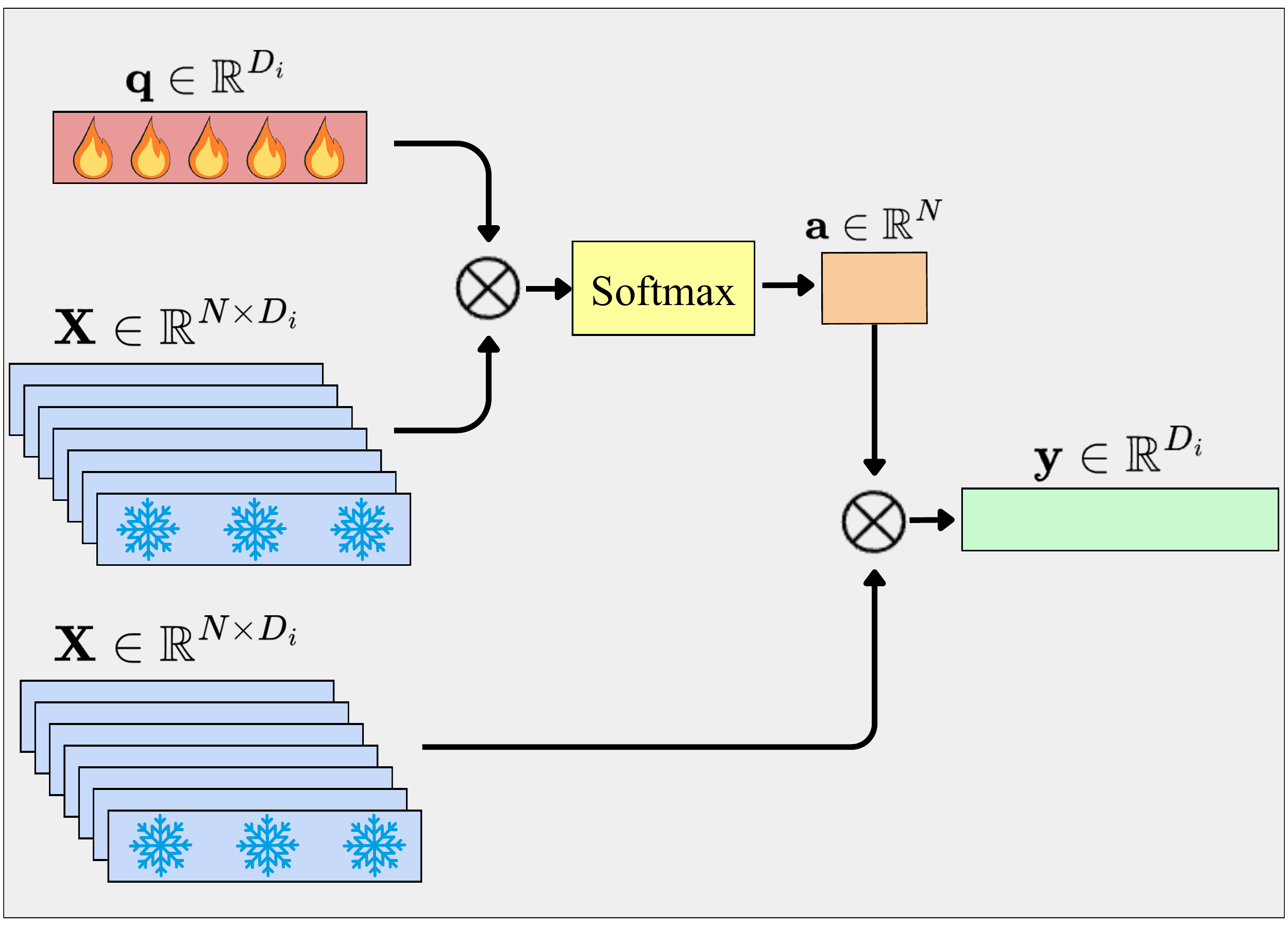}
        \caption{AbMILP}
        \label{fig:probing_abmilp}
    \end{subfigure}
    \begin{subfigure}[b]{0.48\textwidth}
        \centering
        \includegraphics[width=\linewidth]{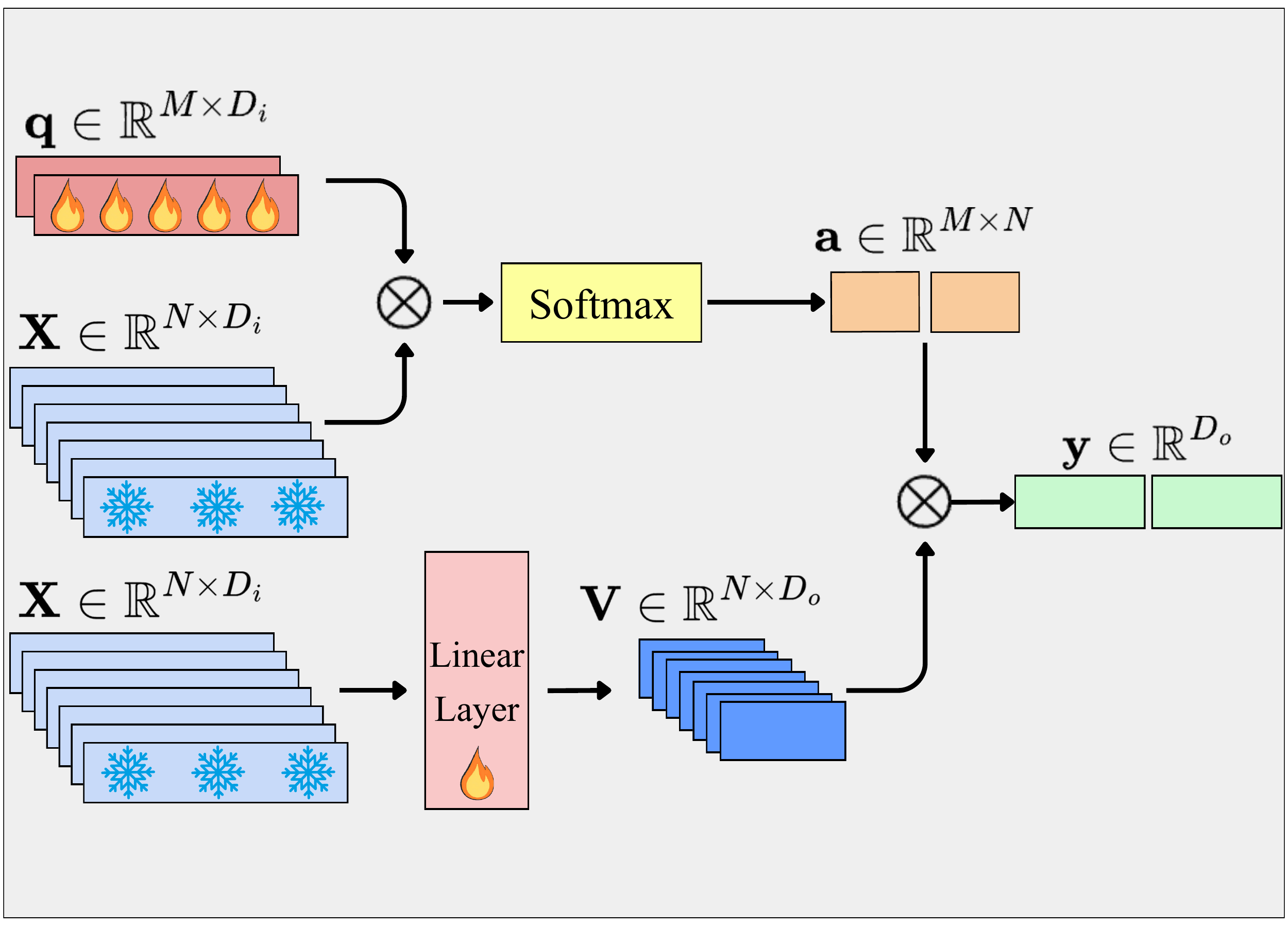}
        \caption{Efficient Probing}
        \label{fig:probing_efficient}
    \end{subfigure}

    \caption{\textbf{Architectural comparison of probing protocols on VFMs.}
    A frozen VFM backbone processes the input image. We then apply three different transformations to form a global representation, on top of which we train a linear probe for spatial relation prediction: \textbf{Linear Probing with Global Average Pooling (GAP)} aggregates all tokens into a single vector regardless of their relevance (\textbf{a}); \textbf{AbMILP} learns a scalar attention map to selectively weight patches (\textbf{b}); \textbf{Efficient Probing} uses a multi-query cross-attention mechanism with a set of learnable queries (\textbf{c}).}
    \label{fig:probing_methods}
\end{figure*}

\textbf{Evaluated Visual Foundation Models (VFMs).} To systematically evaluate the
\revision{encoding of directional spatial relations,}
we curate a diverse suite of VFMs spanning distinct learning paradigms, as detailed in \Cref{tab:models}. Our selection prioritizes self-supervised learning (SSL) methods, which we categorize into Joint-Embedding Architectures (JEA)~\cite{caron2021emerging,oquab2023dinov2,siméoni2025dinov3} and
Masked Image Modeling (MIM)~\cite{he2021masked,wei2023masked,zhu_spa_2024,weinzaepfel2022croco,weinzaepfel2022crocov2}.
Moreover, we compare to several VFMs prominent in the community, trained in a supervised and weakly supervised manner~\cite{wang2025vggt,touvron2022deit,radford2021clip}.
To ensure a fair comparison of learned representations, we evaluate all models using a ViT-Base backbone with an input resolution standardized to $224 \times 224$. We note two necessary exceptions regarding model size. VGGT~\cite{wang2025vggt} provides only ViT-Large checkpoints.
We explicitly include VGGT, which is trained with direct 3D supervision (cameras, depth, tracks), to serve as a 3D-native baseline. This allows us to determine if explicit 3D training signals are strictly necessary for the abstract spatial reasoning tasks in \our{}, or if such capabilities can emerge from 2D self-supervision alone.
For a fair comparison on the ViT-L backbone, we also add the DINO-v2 reg ViT-Large model~\cite{oquab2023dinov2,darcet_vision_2023}, a JEA model whose parameters were also used to initialize VGGT during training~\cite{wang2025vggt}.
By comparing these two models, we directly analyze the shift in representations of a model pre-trained for high-level recognition resulting from adaptation to 3D perception tasks.

\textbf{Adapting VFMs to solve \our{} via probing.}
At a conceptual level, our approach follows a simple principle: if a pre-trained VFM representation contains spatial information, it should be possible to extract it via lightweight probing.
By default, we extract features from the final transformer block (we provide a detailed analysis of intermediate layers in~\Cref{sec:layer-wise}).
Following the protocols established in recent mid-level vision benchmarks~\cite{chen2024probingmidlevelvisioncapabilities}, we evaluate representations by attaching different types of probing heads to the VFMs: \textbf{(i)} linear probing with Global Average Pooling of the patch tokens, as well as Selective Aggregation via \textbf{(ii)} Attention-Based Multiple Instance Learning Pooling (AbMILP)~\cite{ilse2018attention,przewiezlikowski2025cls}, and \textbf{(iii)} Efficient probing~\cite{psomas2025attention}. We depict these approaches in~\Cref{fig:probing_methods}.
Probe \textbf{(i)} evaluates a global representation formed by naively aggregating dense VFM features, while probes \textbf{(ii)} and \textbf{(iii)} additionally account for their spatially local nature.

\begin{table}[ht]
    \centering
\resizebox{0.5\textwidth}{!}{
\begin{tabular}{@{}lccc@{}}
    \toprule
    \textbf{Hyperparameter} & \textbf{Linear Probing} & \textbf{AbMILP} & \textbf{Efficient Probing} \\
    \midrule
    Optimizer & \multicolumn{3}{c}{AdamW} \\
    Weight Decay & \multicolumn{3}{c}{0.001} \\
    Scheduler & \multicolumn{3}{c}{Cosine Decay} \\
    Learning Rate & \multicolumn{3}{c}{$10^{-2}, 10^{-3}, 10^{-4}$} \\
    Dropout & \multicolumn{3}{c}{$0.2, 0.4, 0.6$} \\
    Batch Size & \multicolumn{3}{c}{256} \\
    Linear Warmup & 200 & 100 & 100 \\
    Training Epochs & 1000 & 500 & 500 \\
    \midrule
    \multicolumn{4}{c}{Method-Specific Architectures} \\
    \midrule
    Hidden Dimension & 1 linear layer & 2-layer MLP & 1 linear layer \\
    Queries ($N_q$) & - & - & 4 \\
    Output Dim ($D_{o}$) & $D_{i}$ & $D_{i}$ & $D_{i}$ / 8 \\
    \bottomrule
\end{tabular}
}
\caption{\textbf{Hyperparameters for Spatial Probing.} We list the configuration used to train the probing heads (Linear, AbMILP, Efficient Probing) on top of frozen VFM backbones. Training settings are consistent across both egocentric and allocentric tasks.}
\label{tab:probing_hyperparams}
\vspace{-1cm}
\end{table}

\textbf{Evaluation protocol.}
For a given VFM, we evaluate \our{} performance using the accuracy of probes trained to recognize spatial relations for specified \texttt{source, target, viewpoint} object triples.
Performance on \our{} across images belonging to a given semantic triple therefore reflects the ability to resolve spatial relations for those particular object types, while averaging over diverse semantic triples provides an aggregate measure of spatial relation recognition.
\revision{
For each such triple of objects, we curate a dataset of images depicting different object layouts and split it into training, validation, and test folds in an 80/10/10 proportion.
The split is performed at the level of rendered samples: no identical rendered image or metadata instance appears in more than one fold.
Within each environment and object triple, train, validation, and test samples differ by independently resampled object placements, camera poses, and viewpoint configurations, subject to the geometric validity constraints described in \Cref{subsec:data_gen_pipeline}.
Importantly, this protocol evaluates generalization to unseen spatial configurations within the same environment and object triple, rather than generalization to held-out environments or unseen object categories. As a result, no identical samples appear in more than one split. However, because training, validation, and test sets are drawn from the same environment, they may still share environment-specific patterns, such as background appearance, lighting, or scene structure.
}
We train triple-specific probes for the number of epochs specified in Table~\ref{tab:probing_hyperparams}, using the validation set to select the best-performing probe parameters. We report the test accuracies of the probes and repeat this procedure with two random seeds and three distinct object triples from each of the five environments (see~\Cref{subsec:assets}) to achieve a robust evaluation.

\begin{table*}[t]
\centering
\small

\begin{tabular}{l ccc @{\hskip 12pt} ccc @{\hskip 12pt} ccc}
\toprule
\multirow{2}{*}{\textbf{Model}} & \multicolumn{3}{c}{\textbf{FOREST}} & \multicolumn{3}{c}{\textbf{DESERT}} & \multicolumn{3}{c}{\textbf{Winter Town}} \\
\cmidrule(r){2-4} \cmidrule(lr){5-7} \cmidrule(l){8-10}
 & Lin. & ABM. & Eff. & Lin. & ABM. & Eff. & Lin. & ABM. & Eff. \\
\midrule
DINO & \cellcolor[rgb]{0.755,0.995,0.750}\underline{64.76} & \cellcolor[rgb]{0.782,0.968,0.750}85.53 & \cellcolor[rgb]{0.793,0.957,0.750}89.28 & \cellcolor[rgb]{0.896,0.854,0.750}55.14 & \cellcolor[rgb]{0.759,0.991,0.750}89.34 & \cellcolor[rgb]{0.805,0.945,0.750}92.10 & \cellcolor[rgb]{0.821,0.929,0.750}61.97 & \cellcolor[rgb]{0.763,0.987,0.750}87.05 & \cellcolor[rgb]{0.798,0.952,0.750}89.32 \\
DINO-v2 (B/14) & \cellcolor[rgb]{0.798,0.952,0.750}59.91 & \cellcolor[rgb]{0.776,0.974,0.750}86.36 & \cellcolor[rgb]{0.777,0.973,0.750}91.91 & \cellcolor[rgb]{0.840,0.910,0.750}64.57 & \cellcolor[rgb]{0.750,1.000,0.750}\textbf{90.64} & \cellcolor[rgb]{0.790,0.960,0.750}93.94 & \cellcolor[rgb]{0.860,0.890,0.750}58.03 & \cellcolor[rgb]{0.750,1.000,0.750}\textbf{88.41} & \cellcolor[rgb]{0.768,0.982,0.750}93.26 \\
DINO-v2 reg (B/14) & \cellcolor[rgb]{0.790,0.960,0.750}60.81 & \cellcolor[rgb]{0.795,0.955,0.750}83.81 & \cellcolor[rgb]{0.779,0.971,0.750}91.60 & \cellcolor[rgb]{0.811,0.939,0.750}\underline{69.40} & \cellcolor[rgb]{0.776,0.974,0.750}86.73 & \cellcolor[rgb]{0.784,0.966,0.750}94.63 & \cellcolor[rgb]{0.786,0.964,0.750}65.38 & \cellcolor[rgb]{0.767,0.983,0.750}86.59 & \cellcolor[rgb]{0.781,0.969,0.750}91.52 \\
DINO-v2 reg (L/14) & \cellcolor[rgb]{0.750,1.000,0.750}\textbf{65.37} & \cellcolor[rgb]{0.774,0.976,0.750}\underline{86.59} & \cellcolor[rgb]{0.764,0.986,0.750}93.92 & \cellcolor[rgb]{0.750,1.000,0.750}\textbf{79.68} & \cellcolor[rgb]{0.761,0.989,0.750}89.03 & \cellcolor[rgb]{0.767,0.983,0.750}96.70 & \cellcolor[rgb]{0.750,1.000,0.750}\textbf{69.02} & \cellcolor[rgb]{0.780,0.970,0.750}85.15 & \cellcolor[rgb]{0.755,0.995,0.750}\underline{94.92} \\
DINOv3 & \cellcolor[rgb]{0.776,0.974,0.750}62.41 & \cellcolor[rgb]{0.788,0.962,0.750}84.71 & \cellcolor[rgb]{0.764,0.986,0.750}\underline{93.93} & \cellcolor[rgb]{0.835,0.915,0.750}65.34 & \cellcolor[rgb]{0.774,0.976,0.750}86.96 & \cellcolor[rgb]{0.761,0.989,0.750}\underline{97.47} & \cellcolor[rgb]{0.779,0.971,0.750}\underline{66.14} & \cellcolor[rgb]{0.795,0.955,0.750}83.56 & \cellcolor[rgb]{0.768,0.982,0.750}93.18 \\
VGGT (L/14) & \cellcolor[rgb]{0.866,0.884,0.750}52.10 & \cellcolor[rgb]{0.750,1.000,0.750}\textbf{89.90} & \cellcolor[rgb]{0.750,1.000,0.750}\textbf{96.18} & \cellcolor[rgb]{0.847,0.903,0.750}63.42 & \cellcolor[rgb]{0.754,0.996,0.750}\underline{90.03} & \cellcolor[rgb]{0.750,1.000,0.750}\textbf{98.78} & \cellcolor[rgb]{0.848,0.902,0.750}59.24 & \cellcolor[rgb]{0.761,0.989,0.750}\underline{87.24} & \cellcolor[rgb]{0.750,1.000,0.750}\textbf{95.53} \\
SPA & \cellcolor[rgb]{0.859,0.891,0.750}52.91 & \cellcolor[rgb]{0.874,0.876,0.750}73.01 & \cellcolor[rgb]{0.862,0.888,0.750}78.33 & \cellcolor[rgb]{0.930,0.820,0.750}49.39 & \cellcolor[rgb]{0.868,0.882,0.750}72.58 & \cellcolor[rgb]{0.816,0.934,0.750}90.80 & \cellcolor[rgb]{0.945,0.805,0.750}49.62 & \cellcolor[rgb]{0.884,0.866,0.750}74.01 & \cellcolor[rgb]{0.786,0.964,0.750}90.89 \\
CroCo & \cellcolor[rgb]{0.917,0.833,0.750}46.25 & \cellcolor[rgb]{0.858,0.892,0.750}75.19 & \cellcolor[rgb]{0.795,0.955,0.750}88.98 & \cellcolor[rgb]{0.934,0.816,0.750}48.77 & \cellcolor[rgb]{0.815,0.935,0.750}80.75 & \cellcolor[rgb]{0.823,0.927,0.750}89.88 & \cellcolor[rgb]{0.948,0.802,0.750}49.32 & \cellcolor[rgb]{0.799,0.951,0.750}83.17 & \cellcolor[rgb]{0.809,0.941,0.750}87.95 \\
CroCo v2 & \cellcolor[rgb]{0.848,0.902,0.750}54.14 & \cellcolor[rgb]{0.851,0.899,0.750}76.18 & \cellcolor[rgb]{0.786,0.964,0.750}90.48 & \cellcolor[rgb]{0.911,0.839,0.750}52.53 & \cellcolor[rgb]{0.807,0.943,0.750}81.91 & \cellcolor[rgb]{0.815,0.935,0.750}90.88 & \cellcolor[rgb]{0.905,0.845,0.750}53.56 & \cellcolor[rgb]{0.780,0.970,0.750}85.20 & \cellcolor[rgb]{0.796,0.954,0.750}89.62 \\
CLIP & \cellcolor[rgb]{1.000,0.750,0.750}36.66 & \cellcolor[rgb]{1.000,0.750,0.750}55.85 & \cellcolor[rgb]{1.000,0.750,0.750}56.33 & \cellcolor[rgb]{1.000,0.750,0.750}37.65 & \cellcolor[rgb]{0.975,0.775,0.750}56.29 & \cellcolor[rgb]{1.000,0.750,0.750}68.44 & \cellcolor[rgb]{1.000,0.750,0.750}44.09 & \cellcolor[rgb]{1.000,0.750,0.750}61.57 & \cellcolor[rgb]{1.000,0.750,0.750}63.21 \\
DeiT & \cellcolor[rgb]{0.856,0.894,0.750}53.23 & \cellcolor[rgb]{0.960,0.790,0.750}61.33 & \cellcolor[rgb]{0.880,0.870,0.750}75.49 & \cellcolor[rgb]{0.898,0.852,0.750}54.75 & \cellcolor[rgb]{1.000,0.750,0.750}52.45 & \cellcolor[rgb]{0.919,0.831,0.750}78.22 & \cellcolor[rgb]{0.898,0.852,0.750}54.24 & \cellcolor[rgb]{0.992,0.758,0.750}62.42 & \cellcolor[rgb]{0.901,0.849,0.750}75.98 \\
MAE & \cellcolor[rgb]{0.772,0.978,0.750}62.82 & \cellcolor[rgb]{0.791,0.959,0.750}84.26 & \cellcolor[rgb]{0.769,0.981,0.750}93.10 & \cellcolor[rgb]{0.877,0.873,0.750}58.36 & \cellcolor[rgb]{0.772,0.978,0.750}87.34 & \cellcolor[rgb]{0.792,0.958,0.750}93.71 & \cellcolor[rgb]{0.812,0.938,0.750}62.80 & \cellcolor[rgb]{0.774,0.976,0.750}85.84 & \cellcolor[rgb]{0.768,0.982,0.750}93.26 \\
MaskFeat & \cellcolor[rgb]{0.908,0.842,0.750}47.20 & \cellcolor[rgb]{0.862,0.888,0.750}74.60 & \cellcolor[rgb]{0.791,0.959,0.750}89.66 & \cellcolor[rgb]{0.969,0.781,0.750}42.87 & \cellcolor[rgb]{0.810,0.940,0.750}81.52 & \cellcolor[rgb]{0.799,0.951,0.750}92.87 & \cellcolor[rgb]{0.963,0.787,0.750}47.80 & \cellcolor[rgb]{0.841,0.909,0.750}78.63 & \cellcolor[rgb]{0.783,0.967,0.750}91.21 \\
\bottomrule
\end{tabular}

\vspace{5pt} 

\begin{tabular}{l ccc @{\hskip 12pt} ccc @{\hskip 12pt} ccc}
\toprule
\multirow{2}{*}{\textbf{Model}} & \multicolumn{3}{c}{\textbf{Bridge}} & \multicolumn{3}{c}{\textbf{City}} & \multicolumn{3}{c}{\textbf{Mean Rank}} \\
\cmidrule(r){2-4} \cmidrule(lr){5-7} \cmidrule(l){8-10}
 & Lin. & ABM. & Eff. & Lin. & ABM. & Eff. & Lin. & ABM. & Eff. \\
\midrule
DINO & \cellcolor[rgb]{0.814,0.936,0.750}63.34 & \cellcolor[rgb]{0.751,0.999,0.750}\underline{89.48} & \cellcolor[rgb]{0.800,0.950,0.750}89.86 & \cellcolor[rgb]{0.885,0.865,0.750}53.01 & \cellcolor[rgb]{0.762,0.988,0.750}83.89 & \cellcolor[rgb]{0.851,0.899,0.750}82.45 & \cellcolor[rgb]{0.829,0.921,0.750}4.80 & \cellcolor[rgb]{0.786,0.964,0.750}3.00 & \cellcolor[rgb]{0.921,0.829,0.750}9.20 \\
DINO-v2 (B/14) & \cellcolor[rgb]{0.780,0.970,0.750}66.69 & \cellcolor[rgb]{0.750,1.000,0.750}\textbf{89.63} & \cellcolor[rgb]{0.769,0.981,0.750}93.75 & \cellcolor[rgb]{0.882,0.868,0.750}53.24 & \cellcolor[rgb]{0.750,1.000,0.750}\textbf{85.31} & \cellcolor[rgb]{0.762,0.988,0.750}93.00 & \cellcolor[rgb]{0.833,0.917,0.750}5.00 & \cellcolor[rgb]{0.750,1.000,0.750}\textbf{1.40} & \cellcolor[rgb]{0.825,0.925,0.750}4.60 \\
DINO-v2 reg (B/14) & \cellcolor[rgb]{0.775,0.975,0.750}67.15 & \cellcolor[rgb]{0.760,0.990,0.750}88.49 & \cellcolor[rgb]{0.771,0.979,0.750}93.52 & \cellcolor[rgb]{0.837,0.913,0.750}\underline{57.15} & \cellcolor[rgb]{0.770,0.980,0.750}82.98 & \cellcolor[rgb]{0.762,0.988,0.750}93.00 & \cellcolor[rgb]{0.792,0.958,0.750}3.00 & \cellcolor[rgb]{0.835,0.915,0.750}5.20 & \cellcolor[rgb]{0.833,0.917,0.750}5.00 \\
DINO-v2 reg (L/14) & \cellcolor[rgb]{0.750,1.000,0.750}\textbf{69.59} & \cellcolor[rgb]{0.752,0.998,0.750}89.40 & \cellcolor[rgb]{0.750,1.000,0.750}\underline{96.11} & \cellcolor[rgb]{0.750,1.000,0.750}\textbf{64.79} & \cellcolor[rgb]{0.775,0.975,0.750}82.38 & \cellcolor[rgb]{0.752,0.998,0.750}\underline{94.27} & \cellcolor[rgb]{0.750,1.000,0.750}\textbf{1.00} & \cellcolor[rgb]{0.812,0.938,0.750}4.20 & \cellcolor[rgb]{0.779,0.971,0.750}\underline{2.40} \\
DINOv3 & \cellcolor[rgb]{0.774,0.976,0.750}\underline{67.29} & \cellcolor[rgb]{0.772,0.978,0.750}87.11 & \cellcolor[rgb]{0.760,0.990,0.750}94.82 & \cellcolor[rgb]{0.840,0.910,0.750}56.93 & \cellcolor[rgb]{0.781,0.969,0.750}81.63 & \cellcolor[rgb]{0.767,0.983,0.750}92.40 & \cellcolor[rgb]{0.787,0.963,0.750}\underline{2.80} & \cellcolor[rgb]{0.862,0.888,0.750}6.40 & \cellcolor[rgb]{0.800,0.950,0.750}3.40 \\
VGGT (L/14) & \cellcolor[rgb]{0.860,0.890,0.750}58.92 & \cellcolor[rgb]{0.760,0.990,0.750}88.49 & \cellcolor[rgb]{0.750,1.000,0.750}\textbf{96.11} & \cellcolor[rgb]{0.964,0.786,0.750}46.08 & \cellcolor[rgb]{0.761,0.989,0.750}\underline{83.99} & \cellcolor[rgb]{0.750,1.000,0.750}\textbf{94.47} & \cellcolor[rgb]{0.875,0.875,0.750}7.00 & \cellcolor[rgb]{0.772,0.978,0.750}\underline{2.40} & \cellcolor[rgb]{0.750,1.000,0.750}\textbf{1.00} \\
SPA & \cellcolor[rgb]{0.881,0.869,0.750}56.90 & \cellcolor[rgb]{0.860,0.890,0.750}77.06 & \cellcolor[rgb]{0.809,0.941,0.750}88.72 & \cellcolor[rgb]{0.971,0.779,0.750}45.48 & \cellcolor[rgb]{0.880,0.870,0.750}69.75 & \cellcolor[rgb]{0.869,0.881,0.750}80.27 & \cellcolor[rgb]{0.917,0.833,0.750}9.00 & \cellcolor[rgb]{0.964,0.786,0.750}11.00 & \cellcolor[rgb]{0.938,0.812,0.750}10.00 \\
CroCo & \cellcolor[rgb]{0.895,0.855,0.750}55.57 & \cellcolor[rgb]{0.818,0.932,0.750}81.86 & \cellcolor[rgb]{0.777,0.973,0.750}92.68 & \cellcolor[rgb]{0.978,0.772,0.750}44.88 & \cellcolor[rgb]{0.837,0.913,0.750}74.93 & \cellcolor[rgb]{0.805,0.945,0.750}87.95 & \cellcolor[rgb]{0.954,0.796,0.750}10.80 & \cellcolor[rgb]{0.929,0.821,0.750}9.40 & \cellcolor[rgb]{0.921,0.829,0.750}9.20 \\
CroCo v2 & \cellcolor[rgb]{0.888,0.862,0.750}56.25 & \cellcolor[rgb]{0.804,0.946,0.750}83.52 & \cellcolor[rgb]{0.791,0.959,0.750}91.01 & \cellcolor[rgb]{0.989,0.761,0.750}43.90 & \cellcolor[rgb]{0.812,0.938,0.750}77.93 & \cellcolor[rgb]{0.821,0.929,0.750}85.99 & \cellcolor[rgb]{0.917,0.833,0.750}9.00 & \cellcolor[rgb]{0.888,0.862,0.750}7.60 & \cellcolor[rgb]{0.904,0.846,0.750}8.40 \\
CLIP & \cellcolor[rgb]{1.000,0.750,0.750}45.35 & \cellcolor[rgb]{1.000,0.750,0.750}61.13 & \cellcolor[rgb]{1.000,0.750,0.750}64.71 & \cellcolor[rgb]{1.000,0.750,0.750}42.92 & \cellcolor[rgb]{0.941,0.809,0.750}62.58 & \cellcolor[rgb]{1.000,0.750,0.750}64.67 & \cellcolor[rgb]{1.000,0.750,0.750}13.00 & \cellcolor[rgb]{1.000,0.750,0.750}12.60 & \cellcolor[rgb]{1.000,0.750,0.750}13.00 \\
DeiT & \cellcolor[rgb]{0.888,0.862,0.750}56.17 & \cellcolor[rgb]{0.974,0.776,0.750}64.10 & \cellcolor[rgb]{0.867,0.883,0.750}81.40 & \cellcolor[rgb]{0.995,0.755,0.750}43.38 & \cellcolor[rgb]{1.000,0.750,0.750}55.50 & \cellcolor[rgb]{0.901,0.849,0.750}76.50 & \cellcolor[rgb]{0.921,0.829,0.750}9.20 & \cellcolor[rgb]{0.996,0.754,0.750}12.40 & \cellcolor[rgb]{0.979,0.771,0.750}12.00 \\
MAE & \cellcolor[rgb]{0.823,0.927,0.750}62.50 & \cellcolor[rgb]{0.763,0.987,0.750}88.11 & \cellcolor[rgb]{0.768,0.982,0.750}93.83 & \cellcolor[rgb]{0.886,0.864,0.750}52.87 & \cellcolor[rgb]{0.805,0.945,0.750}78.77 & \cellcolor[rgb]{0.796,0.954,0.750}89.00 & \cellcolor[rgb]{0.833,0.917,0.750}5.00 & \cellcolor[rgb]{0.848,0.902,0.750}5.80 & \cellcolor[rgb]{0.825,0.925,0.750}4.60 \\
MaskFeat & \cellcolor[rgb]{0.912,0.838,0.750}53.89 & \cellcolor[rgb]{0.809,0.941,0.750}82.85 & \cellcolor[rgb]{0.780,0.970,0.750}92.30 & \cellcolor[rgb]{0.979,0.771,0.750}44.80 & \cellcolor[rgb]{0.856,0.894,0.750}72.67 & \cellcolor[rgb]{0.891,0.859,0.750}77.64 & \cellcolor[rgb]{0.967,0.783,0.750}11.40 & \cellcolor[rgb]{0.933,0.817,0.750}9.60 & \cellcolor[rgb]{0.900,0.850,0.750}8.20 \\
\bottomrule
\end{tabular}

\caption{\textbf{Results on the Egocentric (Camera-Perspective) Task.} We report Accuracy (\%) across five environments. While most models achieve reasonable performance with Linear probing, efficient probing yields substantial gains, particularly for MIMs (e.g., MAE), which retain rich spatial information in patch tokens that is otherwise lost in global average pooling. Crucially, VGGT achieves state-of-the-art performance across all environments when using efficient probing, demonstrating that explicit 3D supervision yields the richest spatial structure in patch-level representations. \textbf{Bold} indicates the best result; underlining indicates the second-best result. Abbreviations: Lin. (Linear Probing), ABM. (AbMILP), Eff. (Efficient Probing). Each column is colored from the lowest (red color) to the highest value (green color).}
\label{tab:ego_results}
\end{table*}

\subsection{\ourfull{} results}
\label{sec:exp_main_results}

We evaluate the effectiveness of Visual Foundation Models (VFMs) in solving the \ourfull{} (\our{}) with three probing strategies across five environments. We summarize the accuracy on the egocentric and allocentric variants of \our{} in~\Cref{tab:ego_results} and~\Cref{tab:allo_results}, respectively.
Below, we discuss several key findings that characterize how VFMs encode spatial relations.

\begin{table*}[t]
\centering
\small
\setlength{\tabcolsep}{3.5pt} 

\begin{tabular}{l ccc @{\hskip 12pt} ccc @{\hskip 12pt} ccc}
\toprule
\multirow{2}{*}{\textbf{Model}} & \multicolumn{3}{c}{\textbf{FOREST}} & \multicolumn{3}{c}{\textbf{DESERT}} & \multicolumn{3}{c}{\textbf{Winter Town}} \\
\cmidrule(r){2-4} \cmidrule(lr){5-7} \cmidrule(l){8-10}
 & Lin. & ABM. & Eff. & Lin. & ABM. & Eff. & Lin. & ABM. & Eff. \\
\midrule
DINO & \cellcolor[rgb]{0.756,0.994,0.750}\underline{59.09} & \cellcolor[rgb]{0.811,0.939,0.750}70.49 & \cellcolor[rgb]{0.853,0.897,0.750}69.45 & \cellcolor[rgb]{0.867,0.883,0.750}39.78 & \cellcolor[rgb]{0.830,0.920,0.750}56.69 & \cellcolor[rgb]{0.924,0.826,0.750}57.12 & \cellcolor[rgb]{0.757,0.993,0.750}\underline{49.21} & \cellcolor[rgb]{0.802,0.948,0.750}63.98 & \cellcolor[rgb]{0.880,0.870,0.750}64.73 \\
DINO-v2 (B/14) & \cellcolor[rgb]{0.778,0.972,0.750}57.44 & \cellcolor[rgb]{0.766,0.984,0.750}73.69 & \cellcolor[rgb]{0.784,0.966,0.750}74.42 & \cellcolor[rgb]{0.838,0.912,0.750}41.93 & \cellcolor[rgb]{0.750,1.000,0.750}\textbf{64.42} & \cellcolor[rgb]{0.868,0.882,0.750}63.80 & \cellcolor[rgb]{0.824,0.926,0.750}45.59 & \cellcolor[rgb]{0.776,0.974,0.750}\underline{66.11} & \cellcolor[rgb]{0.830,0.920,0.750}70.26 \\
DINO-v2 reg (B/14) & \cellcolor[rgb]{0.791,0.959,0.750}56.45 & \cellcolor[rgb]{0.804,0.946,0.750}70.98 & \cellcolor[rgb]{0.803,0.947,0.750}73.11 & \cellcolor[rgb]{0.841,0.909,0.750}41.77 & \cellcolor[rgb]{0.791,0.959,0.750}60.52 & \cellcolor[rgb]{0.874,0.876,0.750}63.14 & \cellcolor[rgb]{0.777,0.973,0.750}48.12 & \cellcolor[rgb]{0.788,0.962,0.750}65.19 & \cellcolor[rgb]{0.860,0.890,0.750}66.92 \\
DINO-v2 reg (L/14) & \cellcolor[rgb]{0.750,1.000,0.750}\textbf{59.55} & \cellcolor[rgb]{0.762,0.988,0.750}\underline{73.94} & \cellcolor[rgb]{0.753,0.997,0.750}\underline{76.66} & \cellcolor[rgb]{0.750,1.000,0.750}\textbf{48.51} & \cellcolor[rgb]{0.778,0.972,0.750}\underline{61.70} & \cellcolor[rgb]{0.813,0.937,0.750}\underline{70.30} & \cellcolor[rgb]{0.750,1.000,0.750}\textbf{49.58} & \cellcolor[rgb]{0.750,1.000,0.750}\textbf{68.26} & \cellcolor[rgb]{0.776,0.974,0.750}\underline{76.35} \\
DINOv3 & \cellcolor[rgb]{0.808,0.942,0.750}55.20 & \cellcolor[rgb]{0.849,0.901,0.750}67.75 & \cellcolor[rgb]{0.772,0.978,0.750}75.31 & \cellcolor[rgb]{0.821,0.929,0.750}\underline{43.22} & \cellcolor[rgb]{0.796,0.954,0.750}59.99 & \cellcolor[rgb]{0.837,0.913,0.750}67.45 & \cellcolor[rgb]{0.759,0.991,0.750}49.08 & \cellcolor[rgb]{0.831,0.919,0.750}61.63 & \cellcolor[rgb]{0.812,0.938,0.750}72.38 \\
VGGT (L/14) & \cellcolor[rgb]{0.853,0.897,0.750}51.85 & \cellcolor[rgb]{0.750,1.000,0.750}\textbf{74.81} & \cellcolor[rgb]{0.750,1.000,0.750}\textbf{76.91} & \cellcolor[rgb]{0.872,0.878,0.750}39.43 & \cellcolor[rgb]{0.787,0.963,0.750}60.85 & \cellcolor[rgb]{0.750,1.000,0.750}\textbf{77.74} & \cellcolor[rgb]{0.889,0.861,0.750}42.10 & \cellcolor[rgb]{0.779,0.971,0.750}65.91 & \cellcolor[rgb]{0.750,1.000,0.750}\textbf{79.27} \\
SPA & \cellcolor[rgb]{0.950,0.800,0.750}44.57 & \cellcolor[rgb]{0.895,0.855,0.750}64.46 & \cellcolor[rgb]{0.833,0.917,0.750}70.89 & \cellcolor[rgb]{0.965,0.785,0.750}32.55 & \cellcolor[rgb]{0.881,0.869,0.750}51.82 & \cellcolor[rgb]{0.867,0.883,0.750}63.90 & \cellcolor[rgb]{0.954,0.796,0.750}38.59 & \cellcolor[rgb]{0.905,0.845,0.750}55.58 & \cellcolor[rgb]{0.861,0.889,0.750}66.89 \\
CroCo & \cellcolor[rgb]{0.952,0.798,0.750}44.46 & \cellcolor[rgb]{0.888,0.862,0.750}64.95 & \cellcolor[rgb]{0.795,0.955,0.750}73.67 & \cellcolor[rgb]{0.971,0.779,0.750}32.06 & \cellcolor[rgb]{0.863,0.887,0.750}53.52 & \cellcolor[rgb]{0.892,0.858,0.750}60.97 & \cellcolor[rgb]{0.973,0.777,0.750}37.59 & \cellcolor[rgb]{0.819,0.931,0.750}62.59 & \cellcolor[rgb]{0.843,0.907,0.750}68.91 \\
CroCo v2 & \cellcolor[rgb]{0.956,0.794,0.750}44.17 & \cellcolor[rgb]{0.883,0.867,0.750}65.33 & \cellcolor[rgb]{0.794,0.956,0.750}73.74 & \cellcolor[rgb]{0.941,0.809,0.750}34.35 & \cellcolor[rgb]{0.811,0.939,0.750}58.58 & \cellcolor[rgb]{0.895,0.855,0.750}60.64 & \cellcolor[rgb]{0.969,0.781,0.750}37.81 & \cellcolor[rgb]{0.818,0.932,0.750}62.69 & \cellcolor[rgb]{0.846,0.904,0.750}68.53 \\
CLIP & \cellcolor[rgb]{1.000,0.750,0.750}40.87 & \cellcolor[rgb]{1.000,0.750,0.750}56.97 & \cellcolor[rgb]{1.000,0.750,0.750}58.79 & \cellcolor[rgb]{1.000,0.750,0.750}29.93 & \cellcolor[rgb]{0.915,0.835,0.750}48.56 & \cellcolor[rgb]{0.979,0.771,0.750}50.72 & \cellcolor[rgb]{0.999,0.751,0.750}36.21 & \cellcolor[rgb]{0.981,0.769,0.750}49.37 & \cellcolor[rgb]{1.000,0.750,0.750}51.29 \\
DeiT & \cellcolor[rgb]{0.832,0.918,0.750}53.44 & \cellcolor[rgb]{0.979,0.771,0.750}58.48 & \cellcolor[rgb]{0.923,0.827,0.750}64.38 & \cellcolor[rgb]{0.867,0.883,0.750}39.81 & \cellcolor[rgb]{1.000,0.750,0.750}40.39 & \cellcolor[rgb]{1.000,0.750,0.750}48.18 & \cellcolor[rgb]{0.860,0.890,0.750}43.65 & \cellcolor[rgb]{1.000,0.750,0.750}47.83 & \cellcolor[rgb]{0.965,0.785,0.750}55.21 \\
MAE & \cellcolor[rgb]{0.861,0.889,0.750}51.23 & \cellcolor[rgb]{0.803,0.947,0.750}71.06 & \cellcolor[rgb]{0.756,0.994,0.750}76.45 & \cellcolor[rgb]{0.924,0.826,0.750}35.57 & \cellcolor[rgb]{0.787,0.963,0.750}60.88 & \cellcolor[rgb]{0.862,0.888,0.750}64.55 & \cellcolor[rgb]{0.897,0.853,0.750}41.70 & \cellcolor[rgb]{0.791,0.959,0.750}64.92 & \cellcolor[rgb]{0.808,0.942,0.750}72.81 \\
MaskFeat & \cellcolor[rgb]{0.954,0.796,0.750}44.28 & \cellcolor[rgb]{0.850,0.900,0.750}67.68 & \cellcolor[rgb]{0.793,0.957,0.750}73.83 & \cellcolor[rgb]{0.971,0.779,0.750}32.11 & \cellcolor[rgb]{0.810,0.940,0.750}58.69 & \cellcolor[rgb]{0.891,0.859,0.750}61.05 & \cellcolor[rgb]{1.000,0.750,0.750}36.14 & \cellcolor[rgb]{0.861,0.889,0.750}59.17 & \cellcolor[rgb]{0.836,0.914,0.750}69.59 \\
\bottomrule
\end{tabular}

\vspace{5pt} 

\begin{tabular}{l ccc @{\hskip 12pt} ccc @{\hskip 12pt} ccc}
\toprule
\multirow{2}{*}{\textbf{Model}} & \multicolumn{3}{c}{\textbf{Bridge}} & \multicolumn{3}{c}{\textbf{City}} & \multicolumn{3}{c}{\textbf{Mean Rank}} \\
\cmidrule(r){2-4} \cmidrule(lr){5-7} \cmidrule(l){8-10}
 & Lin. & ABM. & Eff. & Lin. & ABM. & Eff. & Lin. & ABM. & Eff. \\
\midrule
DINO & \cellcolor[rgb]{0.783,0.967,0.750}48.98 & \cellcolor[rgb]{0.813,0.937,0.750}60.61 & \cellcolor[rgb]{0.910,0.840,0.750}62.25 & \cellcolor[rgb]{0.789,0.961,0.750}48.29 & \cellcolor[rgb]{0.881,0.869,0.750}57.78 & \cellcolor[rgb]{0.854,0.896,0.750}64.89 & \cellcolor[rgb]{0.796,0.954,0.750}3.20 & \cellcolor[rgb]{0.884,0.866,0.750}7.60 & \cellcolor[rgb]{0.953,0.797,0.750}10.60 \\
DINO-v2 (B/14) & \cellcolor[rgb]{0.793,0.957,0.750}48.38 & \cellcolor[rgb]{0.751,0.999,0.750}\underline{64.96} & \cellcolor[rgb]{0.836,0.914,0.750}69.58 & \cellcolor[rgb]{0.814,0.936,0.750}47.08 & \cellcolor[rgb]{0.750,1.000,0.750}\textbf{66.00} & \cellcolor[rgb]{0.787,0.963,0.750}70.53 & \cellcolor[rgb]{0.819,0.931,0.750}4.20 & \cellcolor[rgb]{0.750,1.000,0.750}\textbf{1.80} & \cellcolor[rgb]{0.823,0.927,0.750}4.60 \\
DINO-v2 reg (B/14) & \cellcolor[rgb]{0.789,0.961,0.750}48.61 & \cellcolor[rgb]{0.759,0.991,0.750}64.42 & \cellcolor[rgb]{0.875,0.875,0.750}65.76 & \cellcolor[rgb]{0.800,0.950,0.750}47.75 & \cellcolor[rgb]{0.798,0.952,0.750}63.00 & \cellcolor[rgb]{0.837,0.913,0.750}66.37 & \cellcolor[rgb]{0.815,0.935,0.750}4.00 & \cellcolor[rgb]{0.810,0.940,0.750}4.40 & \cellcolor[rgb]{0.901,0.849,0.750}8.20 \\
DINO-v2 reg (L/14) & \cellcolor[rgb]{0.750,1.000,0.750}\textbf{51.05} & \cellcolor[rgb]{0.756,0.994,0.750}64.63 & \cellcolor[rgb]{0.756,0.994,0.750}\underline{77.60} & \cellcolor[rgb]{0.760,0.990,0.750}\underline{49.70} & \cellcolor[rgb]{0.754,0.996,0.750}\underline{65.78} & \cellcolor[rgb]{0.750,1.000,0.750}\textbf{73.61} & \cellcolor[rgb]{0.750,1.000,0.750}\textbf{1.20} & \cellcolor[rgb]{0.755,0.995,0.750}\underline{2.00} & \cellcolor[rgb]{0.763,0.987,0.750}\underline{1.80} \\
DINOv3 & \cellcolor[rgb]{0.766,0.984,0.750}\underline{50.04} & \cellcolor[rgb]{0.836,0.914,0.750}59.01 & \cellcolor[rgb]{0.839,0.911,0.750}69.28 & \cellcolor[rgb]{0.750,1.000,0.750}\textbf{50.21} & \cellcolor[rgb]{0.878,0.872,0.750}57.98 & \cellcolor[rgb]{0.807,0.943,0.750}68.83 & \cellcolor[rgb]{0.782,0.968,0.750}\underline{2.60} & \cellcolor[rgb]{0.889,0.861,0.750}7.80 & \cellcolor[rgb]{0.815,0.935,0.750}4.20 \\
VGGT (L/14) & \cellcolor[rgb]{0.821,0.929,0.750}46.64 & \cellcolor[rgb]{0.750,1.000,0.750}\textbf{65.02} & \cellcolor[rgb]{0.750,1.000,0.750}\textbf{78.15} & \cellcolor[rgb]{0.929,0.821,0.750}41.49 & \cellcolor[rgb]{0.803,0.947,0.750}62.66 & \cellcolor[rgb]{0.779,0.971,0.750}\underline{71.19} & \cellcolor[rgb]{0.884,0.866,0.750}7.00 & \cellcolor[rgb]{0.778,0.972,0.750}3.00 & \cellcolor[rgb]{0.750,1.000,0.750}\textbf{1.20} \\
SPA & \cellcolor[rgb]{0.927,0.823,0.750}40.12 & \cellcolor[rgb]{0.865,0.885,0.750}56.99 & \cellcolor[rgb]{0.894,0.856,0.750}63.81 & \cellcolor[rgb]{1.000,0.750,0.750}37.99 & \cellcolor[rgb]{0.957,0.793,0.750}52.95 & \cellcolor[rgb]{0.923,0.827,0.750}59.15 & \cellcolor[rgb]{0.954,0.796,0.750}10.00 & \cellcolor[rgb]{0.958,0.792,0.750}10.80 & \cellcolor[rgb]{0.927,0.823,0.750}9.40 \\
CroCo & \cellcolor[rgb]{0.972,0.778,0.750}37.31 & \cellcolor[rgb]{0.788,0.962,0.750}62.35 & \cellcolor[rgb]{0.859,0.891,0.750}67.35 & \cellcolor[rgb]{0.991,0.759,0.750}38.44 & \cellcolor[rgb]{0.837,0.913,0.750}60.55 & \cellcolor[rgb]{0.841,0.909,0.750}66.01 & \cellcolor[rgb]{0.981,0.769,0.750}11.20 & \cellcolor[rgb]{0.903,0.847,0.750}8.40 & \cellcolor[rgb]{0.892,0.858,0.750}7.80 \\
CroCo v2 & \cellcolor[rgb]{0.966,0.784,0.750}37.69 & \cellcolor[rgb]{0.778,0.972,0.750}63.08 & \cellcolor[rgb]{0.861,0.889,0.750}67.15 & \cellcolor[rgb]{0.986,0.764,0.750}38.67 & \cellcolor[rgb]{0.770,0.980,0.750}64.75 & \cellcolor[rgb]{0.825,0.925,0.750}67.37 & \cellcolor[rgb]{0.963,0.787,0.750}10.40 & \cellcolor[rgb]{0.856,0.894,0.750}6.40 & \cellcolor[rgb]{0.892,0.858,0.750}7.80 \\
CLIP & \cellcolor[rgb]{0.965,0.785,0.750}37.75 & \cellcolor[rgb]{0.942,0.808,0.750}51.65 & \cellcolor[rgb]{1.000,0.750,0.750}53.29 & \cellcolor[rgb]{0.957,0.793,0.750}40.11 & \cellcolor[rgb]{1.000,0.750,0.750}50.28 & \cellcolor[rgb]{1.000,0.750,0.750}52.73 & \cellcolor[rgb]{0.986,0.764,0.750}11.40 & \cellcolor[rgb]{0.995,0.755,0.750}12.40 & \cellcolor[rgb]{1.000,0.750,0.750}12.80 \\
DeiT & \cellcolor[rgb]{0.818,0.932,0.750}46.86 & \cellcolor[rgb]{1.000,0.750,0.750}47.61 & \cellcolor[rgb]{0.968,0.782,0.750}56.47 & \cellcolor[rgb]{0.849,0.901,0.750}45.35 & \cellcolor[rgb]{0.997,0.753,0.750}50.46 & \cellcolor[rgb]{0.911,0.839,0.750}60.19 & \cellcolor[rgb]{0.856,0.894,0.750}5.80 & \cellcolor[rgb]{1.000,0.750,0.750}12.60 & \cellcolor[rgb]{0.983,0.767,0.750}12.00 \\
MAE & \cellcolor[rgb]{0.864,0.886,0.750}43.97 & \cellcolor[rgb]{0.785,0.965,0.750}62.59 & \cellcolor[rgb]{0.826,0.924,0.750}70.61 & \cellcolor[rgb]{0.947,0.803,0.750}40.59 & \cellcolor[rgb]{0.801,0.949,0.750}62.80 & \cellcolor[rgb]{0.807,0.943,0.750}68.88 & \cellcolor[rgb]{0.907,0.843,0.750}8.00 & \cellcolor[rgb]{0.815,0.935,0.750}4.60 & \cellcolor[rgb]{0.797,0.953,0.750}3.40 \\
MaskFeat & \cellcolor[rgb]{1.000,0.750,0.750}35.59 & \cellcolor[rgb]{0.847,0.903,0.750}58.27 & \cellcolor[rgb]{0.856,0.894,0.750}67.58 & \cellcolor[rgb]{0.999,0.751,0.750}38.03 & \cellcolor[rgb]{0.965,0.785,0.750}52.47 & \cellcolor[rgb]{0.858,0.892,0.750}64.61 & \cellcolor[rgb]{1.000,0.750,0.750}12.00 & \cellcolor[rgb]{0.921,0.829,0.750}9.20 & \cellcolor[rgb]{0.879,0.871,0.750}7.20 \\
\bottomrule
\end{tabular}

\caption{\textbf{Results on the Allocentric (Human-Perspective) Task.} This task requires implicit perspective taking, causing a significant performance drop compared to the egocentric task. Linear probing fails for most models, hovering near the random baseline. However, VGGT combined with efficient probing achieves the best performance, demonstrating that 3D-supervised features retain precise geometric structure in patch tokens, which patch-aware probes can successfully extract. \textbf{Bold} indicates the best result; underlining indicates the second-best result. Abbreviations: Lin. (Linear Probing), ABM. (AbMILP), Eff. (Efficient Probing). Each column is colored from the lowest (red color) to the highest value (green color).}
\label{tab:allo_results}
\end{table*}

\begin{figure*}[ht]
        \centering
        \includegraphics[width=\linewidth]{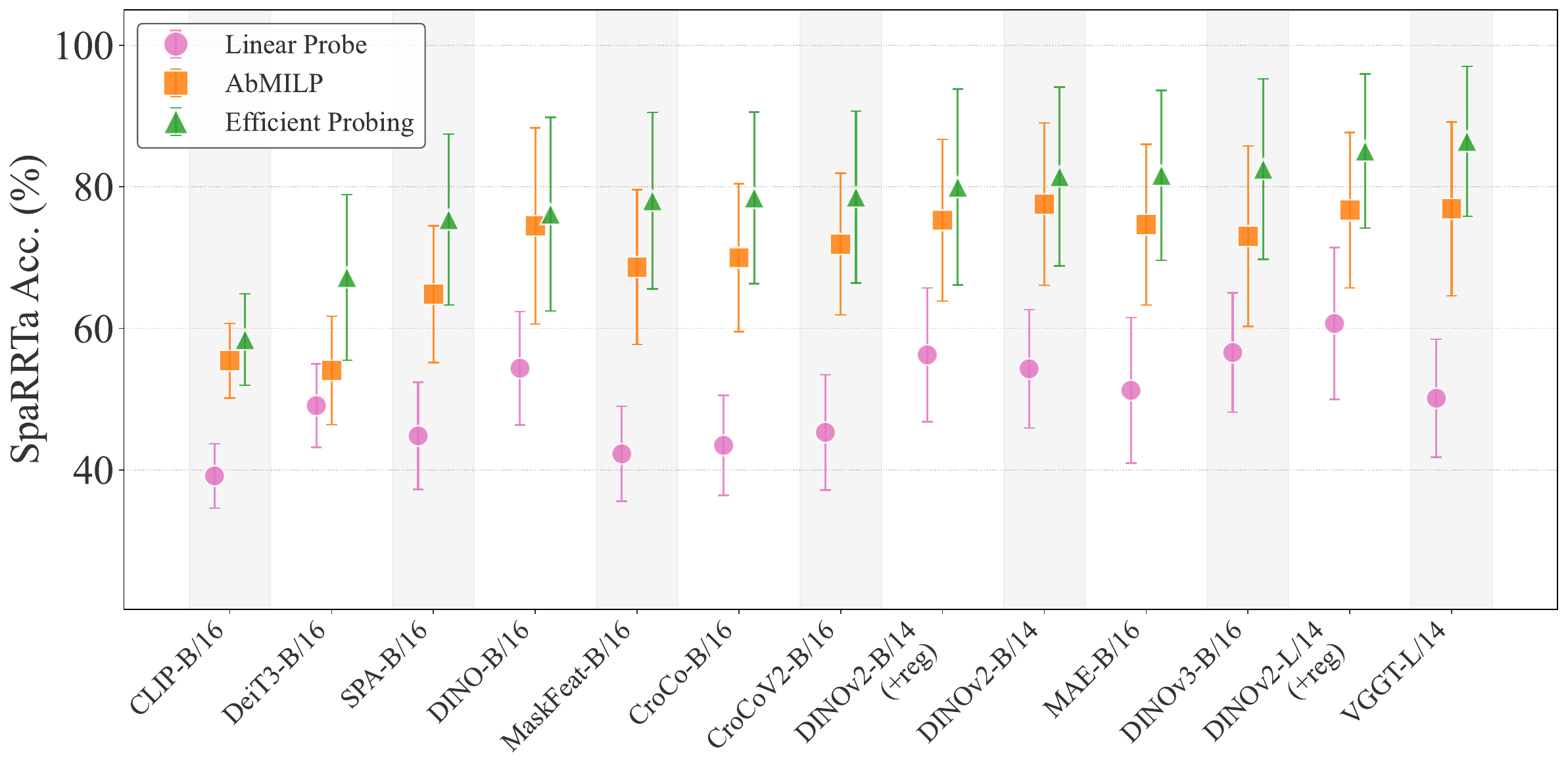}
        \caption{\textbf{Impact of Probing Strategy on Spatial Accuracy.} We report the aggregated mean accuracy across all five evaluation environments and both tasks (\our{}\textbf{-ego} and \our{}\textbf{-allo}) for all VFMs. The same performance hierarchy emerges across all backbones (\textbf{Linear $<$ AbMILP $<$ Efficient Probing}), indicating that spatial information is primarily encoded in local patch features and largely lost during global pooling.}
        \label{fig:scatter_probe_comparison}
\end{figure*}

\textbf{Spatial relation information is more accessible in dense patch representations than in global features.}
Across all evaluated models, we observe a consistent performance hierarchy of \textbf{Linear $<$ AbMILP $<$ Efficient Probing} (\Cref{fig:scatter_probe_comparison}), indicating that spatial aggregation mechanisms capable of selectively emphasizing informative regions are critical for \our{}.
Linear probing based on global average pooling assumes that spatial information is uniformly present across patches; in complex scenes, this assumption breaks down, as only patches corresponding to the source, target, or viewpoint objects are informative, while background patches dilute the signal.
AbMILP partially alleviates this issue by learning patch-wise importance weights, leading to consistent improvements over linear probing, particularly in cluttered environments.

This trend is further amplified by the distinction between single-map and multi-query aggregation.
While AbMILP compresses selected regions into a single weighted representation via one attention map~\cite{przewiezlikowski2025cls}, efficient probing employs multiple learnable queries that can specialize to different spatial entities or scene components~\cite{psomas2025attention}. We provide qualitative visualizations of these learned attention patterns in~\Cref{sec:extend_attention}.
For relation-based tasks that require reasoning about multiple distinct objects, this additional capacity provides a consistent advantage.
\revision{We note that probe complexity in this setting is not determined by parameter count alone. In our ViT-Base setup, the linear probing head with GAP has 4,612 learnable parameters, while AbMILP has 5,381, i.e., only 769 more. Even with this small increase, AbMILP already performs clearly better than linear probing. Efficient Probing is much larger, with 77,380 learnable parameters, but its gains over AbMILP are comparatively small. This suggests that improvements in patch-aware probes are not solely due to a larger parameter budget, and that the way the probe accesses and combines features plays an important role.}

\textbf{3D supervision primarily enhances patch-level spatial structure rather than global representations.}
Different probing mechanisms reveal a counterintuitive divergence when comparing VGGT~\cite{wang2025vggt} with its parent model, DINO-v2 reg~\cite{oquab2023dinov2,darcet_vision_2023}, from which it is initialized and further trained with explicit 3D-awareness supervision.
Under linear probing, VGGT does not outperform DINO-v2 and in some cases performs slightly worse, indicating that 3D supervision does not substantially improve global image-level representations for spatial relation recognition.
In contrast, under AbMILP and efficient probing, VGGT surpasses DINO-v2 on the \our{} task (see~\Cref{fig:scatter_probe_comparison}).
This divergence suggests that 3D supervision enriches patch-level geometric structure, which remains largely inaccessible to global pooling but can be effectively exploited by spatially structured probing mechanisms.

\begin{figure*}[ht]
        \centering
        \includegraphics[width=\linewidth]{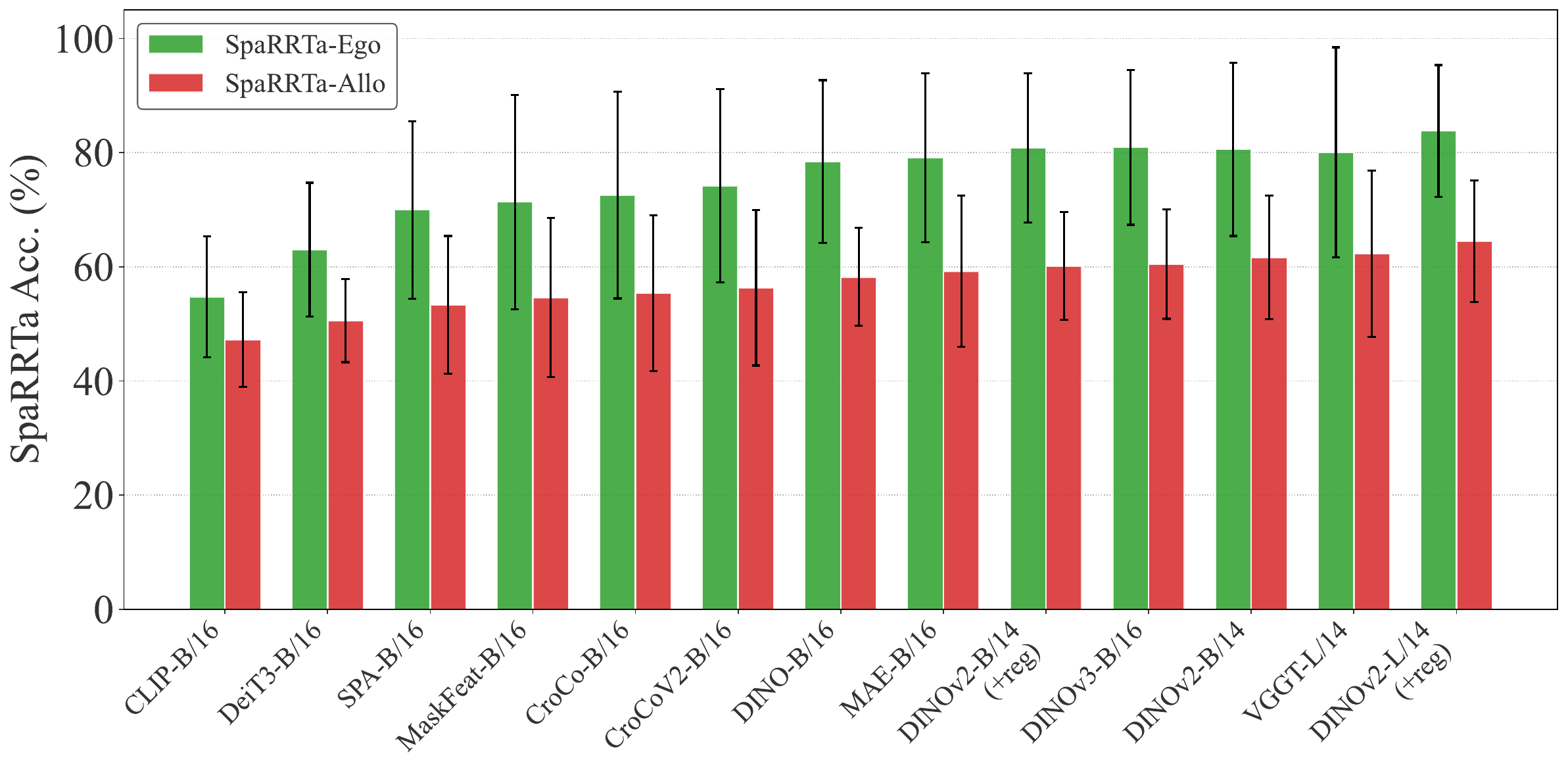}
        \caption{\textbf{Impact of Viewpoint Shift on Spatial Accuracy.} We report the mean accuracy aggregated across all probing methods and environments for each VFM. The consistent superiority of the egocentric bars (left) over the allocentric bars (right) confirms that resolving spatial relations relative to a non-camera viewpoint is significantly more challenging than \revision{identifying them based on} the camera view, regardless of the underlying model architecture.}
        \label{fig:easy_vs_hard_comparison}
            \vspace{-0.5cm}
\end{figure*}

\textbf{Allocentric spatial relation recognition is consistently more challenging than egocentric recognition.}
Across all evaluated models, probing strategies, and environments, accuracy on the allocentric variant is systematically lower than on the egocentric variant (see~\Cref{fig:easy_vs_hard_comparison}).
Egocentric relations can be resolved directly from the camera viewpoint, and performance for the strongest models approaches saturation.
In contrast, allocentric recognition requires \revision{a form of reasoning} about spatial relations from a viewpoint that does not coincide with the rendered image, introducing a substantial and persistent performance gap that remains challenging for all tested VFMs.

\textbf{Visual foundation models exhibit a largely consistent hierarchy of spatial recognition performance.}
Across task variants and probing strategies, we observe a stable ranking of models in terms of \our{} accuracy (see mean ranks in~\Cref{tab:ego_results} and~\Cref{tab:allo_results}).
In particular, CLIP~\cite{radford2021clip} and DeiT~\cite{touvron2022deit} consistently achieve low performance, indicating that their representations provide limited support for decoding spatial relations.
In contrast, the DINO family of models, especially DINO-v2 and DINOv3~\cite{oquab2023dinov2,siméoni2025dinov3} pre-trained on large-scale datasets, achieves strong and consistent performance across settings.
VGGT~\cite{wang2025vggt} further improves upon these results under spatially selective probing strategies, reflecting the benefits of explicit 3D supervision.
Notably, the Masked Autoencoder (MAE)~\cite{he2021masked} also achieves competitive performance despite being pretrained on a smaller-scale dataset and using a pixel-level reconstruction objective~\cite{balestriero2024how}.
Together, these results suggest that spatial relation information is more readily accessible in representations that preserve rich, structured patch-level features.

\begin{figure}[th]
        \centering
        \includegraphics[width=\linewidth]{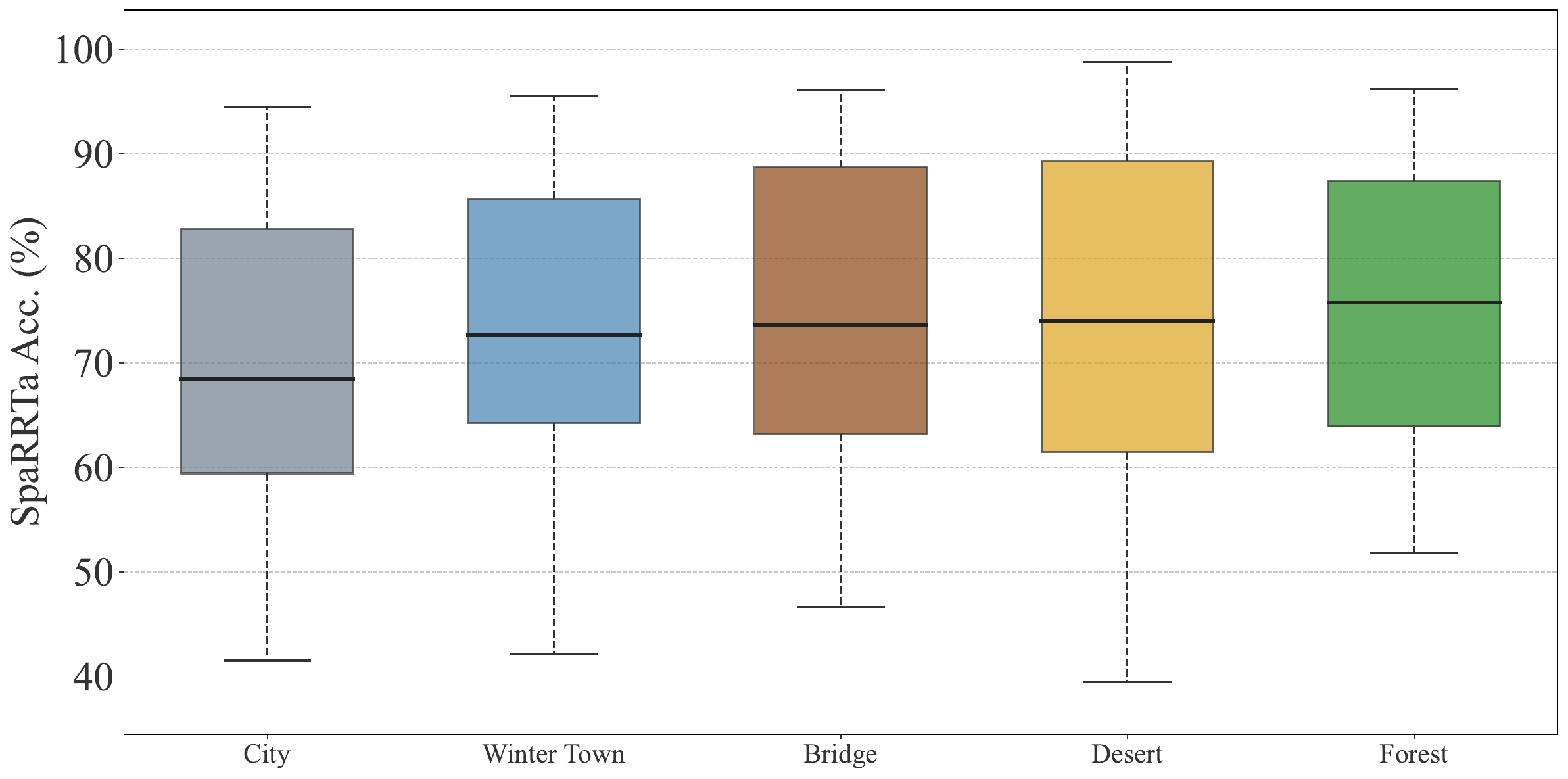}
        \caption{\textbf{Influence of Environmental Complexity on Model Performance.} We report the accuracy distribution of ViT Large backbones across five evaluation environments, averaged over all probing strategies. Performance peaks in visually homogeneous scenes (e.g., Desert, Forest) but degrades in cluttered scenes (e.g., City). This shows that visual noise makes it harder for models to understand spatial relations.}
        \label{fig:environment_difficulty_boxplot}
\end{figure}

\textbf{Environmental complexity significantly affects spatial relation recognition accuracy.}
As shown in~\Cref{fig:environment_difficulty_boxplot}, models achieve higher performance in environments with visually homogeneous backgrounds, such as Desert and Forest, and lower performance in scenes with increased semantic clutter, such as City and Winter Town.
This performance gap indicates that background complexity and visual noise can interfere with identifying the patches relevant for spatial relation \revision{recognition}
, making the task more challenging in cluttered environments.
The consistency of this trend across models suggests that sensitivity to environmental complexity is a general property of current visual representations rather than a model-specific artifact.

Additional validation experiments are provided in~\Cref{sec:additional_results} to assess the scope and robustness of the benchmark beyond the main protocol, in which the training, validation, and test images share the same environment and source--target--viewpoint object triple while differing in their spatial configurations. We test synthetic-to-real transfer using a real-world control set of 1,000 photographs of three Lego objects. Representative examples are shown in~\Cref{fig:lego_real_world_examples}, with the corresponding results reported in~\Cref{tab:real_world_holdout_control}. We analyze probe-data requirements by measuring how allocentric accuracy changes with the training-set size in~\Cref{fig:scaling_allo}. To examine generalization beyond a fixed semantic configuration, we also evaluate transfer to an unseen object triple in~\Cref{tab:cross_triple_transfer}. Additional controls study the effects of the viewpoint-object category and input resolution in~\Cref{tab:camel_viewpoint_control} and~\Cref{tab:full-res-vggt-dinov2}, respectively.

\subsection{Comparison of \our{} with \revision{3D geometric and semantic benchmarks}}
\label{sec:exp_correlation}

To validate that \our{} indeed measures
\revision{spatial relation recognition}
rather than proxy signals (e.g., low-level texture matching or high-level semantics), we correlate our results with established computer vision benchmarks.
We select two mid-level geometric tasks, camera pose estimation and monocular depth estimation on the NAVI dataset~\cite{jampani_navi_2023}, and two high-level semantic tasks, FGVC Aircraft~\cite{maji2013classificationaircraft} and Flowers102~\cite{Nilsback08} classification~\cite{kargin_visbench}. For external benchmarks, we strictly follow standard probing protocols. For classification, we use linear probing on frozen backbones~\cite{chen2024probingmidlevelvisioncapabilities}. For depth estimation, we attach a dense prediction transformer (DPT) head~\cite{ranftl_dense_prediction} as in~\cite{banani_probing_2024}, and for camera pose estimation, we use a 3-layer MLP regressor as in~\cite{zhu_spa_2024}. For \our{}, we use the mean accuracy across all five environments obtained by efficient probing~\cite{psomas2025attention}, as it provides the most robust estimate of a model's spatial capability.

\begin{figure*}[ht]
   \centering
   \includegraphics[width=\linewidth]{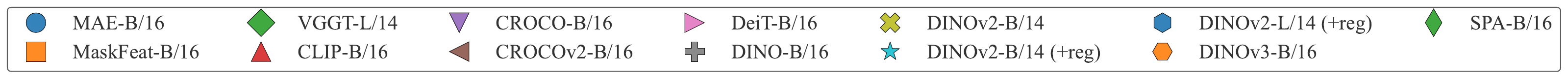}
   \begin{subfigure}[b]{0.48\textwidth}
       \centering
       \includegraphics[width=\linewidth]{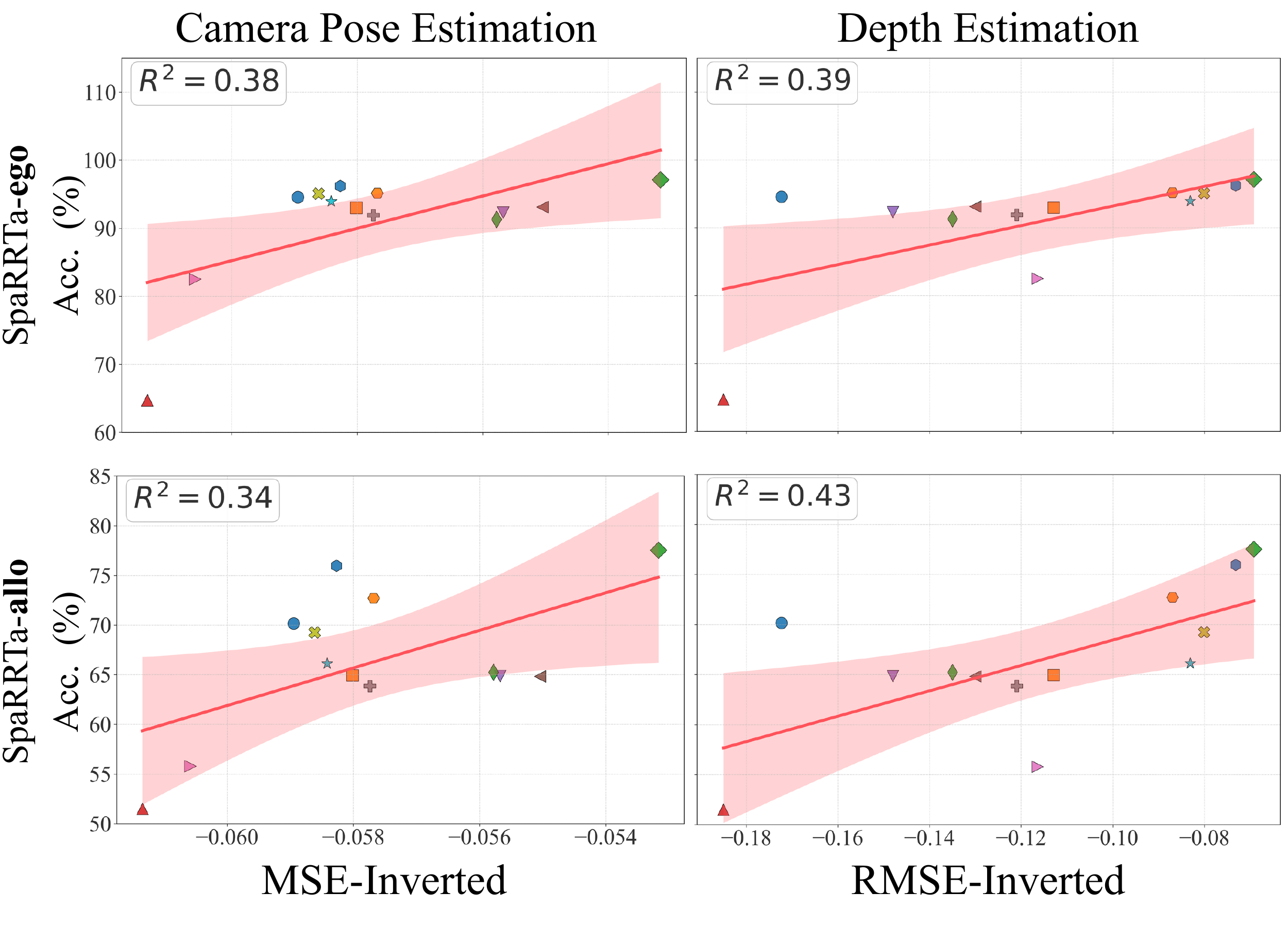}
       \caption{Correlation between \our{} performance and 3D mid-level vision tasks.}
       \label{fig:correlation_mid_level}
   \end{subfigure}
   \hfill
   \begin{subfigure}[b]{0.48\textwidth}
       \centering
       \includegraphics[width=\linewidth]{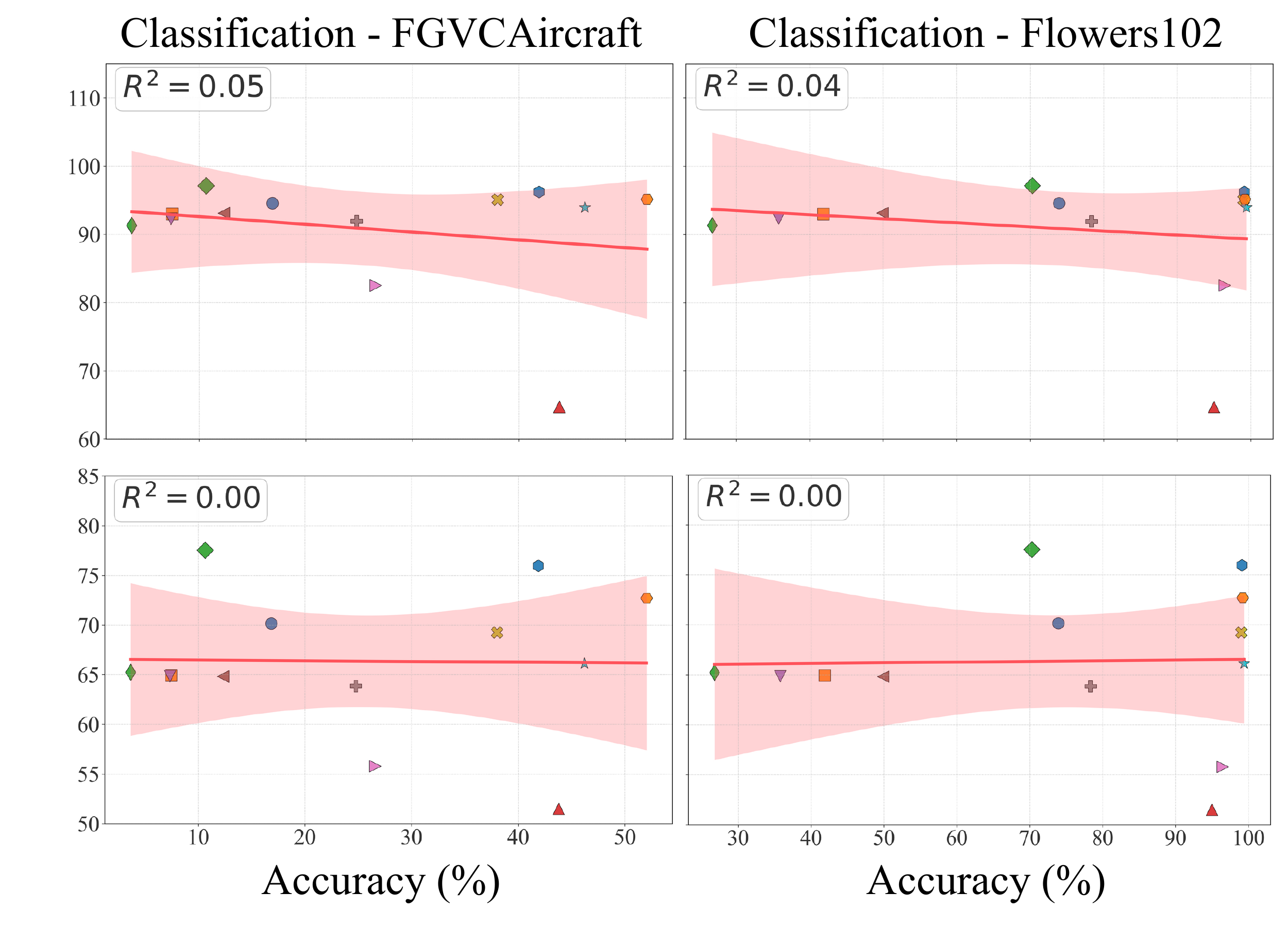}
       \caption{Correlation between \our{} performance and high-level semantic classification.}
       \label{fig:correlation_high_level}
   \end{subfigure}
   \caption{\textbf{Correlation of \our{} with 3D and Semantic Benchmarks.} We compare model accuracy on the \our{}-\textbf{ego} (top) and \our{}-\textbf{allo} (bottom) tasks against four external benchmarks: Camera Pose and Depth Estimation (left) and FGVC Aircraft and Flowers102 Classification (right). We observe a positive correlation ($R^2 \approx 0.40$) with 3D geometric tasks, but a negligible correlation ($R^2 \approx 0.00$) with fine-grained semantic classification. This confirms that \our{} targets latent spatial awareness and is orthogonal to the high-level semantic capabilities, validating \our{} as a distinct and complementary benchmark for VFM evaluation. Note: X-axis metrics (MSE/RMSE) on 3D tasks are inverted so that higher values indicate better performance.}
   \label{fig:model_rank_comparison}
\end{figure*}

As evident from~\Cref{fig:correlation_mid_level}, we observe a significant positive correlation between \our{} accuracy and performance on 3D tasks. Specifically, the allocentric task achieves a coefficient of determination of $R^2 = 0.43$ with depth estimation and $R^2 = 0.34$ with camera pose estimation. This strong relationship confirms that the abstract reasoning required by \our{} is grounded in the same latent geometric representations used for explicit 3D reconstruction. Models with a rich understanding of depth and viewpoint (e.g., VGGT and DINOv3) consistently excel at \our{}, reinforcing the benchmark's validity as a probe of spatial awareness. We further observe that models pre-trained with explicit multi-view consistency objectives, specifically SPA~\cite{zhu_spa_2024}, CroCo~\cite{weinzaepfel2022croco}, and CroCo v2~\cite{weinzaepfel2022crocov2}, exhibit higher performance on the camera pose estimation task than on our tasks and on the other tasks. We attribute this to their unique pre-training paradigms, which compel the network to internalize camera transformations. CroCo's cross-view completion objective requires the model to reconstruct masked patches from two different viewpoints of the same scene, effectively learning camera geometry as a latent variable. Similarly, SPA employs volumetric neural rendering to enforce consistency across multiple views from the same scene, directly embedding 3D spatial structure into the 2D representation. Consequently, these models exhibit a stronger inductive bias for geometric tasks, which explains their distinct advantage in regressing camera poses.

In contrast,~\Cref{fig:correlation_high_level} reveals the lack of correlation ($R^2 \approx 0.00$) between \our{} and fine-grained classification benchmarks. High performance in semantic tasks (e.g., identifying flower species) does not predict success in spatial relationships among objects. This result implies that \our{} targets a specific spatial skill set that does not emerge from strong semantic features.

\begin{figure}[t]
        \centering
        \includegraphics[width=\linewidth]{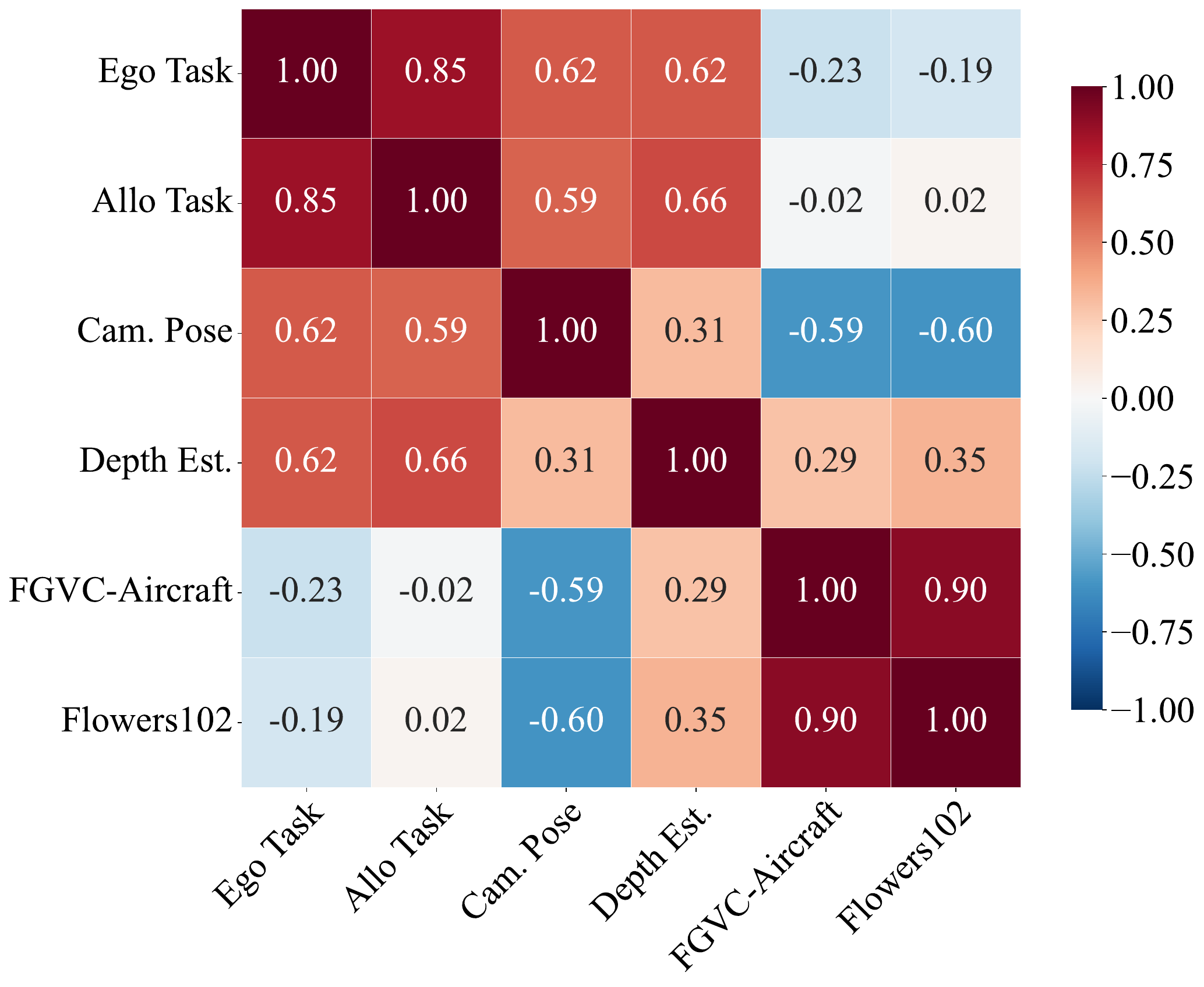}
        \caption{\textbf{Pearson Correlation Matrix Across Vision Benchmarks.} We visualize the pairwise correlation ($r$) between our proposed tasks (Ego, Allo) and external benchmarks. A distinct cluster emerges between \our{} and the 3D geometry tasks (Camera Pose, Depth Estimation), while correlations with semantic classification remain weak or negative. This structural separation underscores the importance of evaluating 3D awareness as an independent capability from semantic understanding.}
        \label{fig:correlation_matrix}
\end{figure}

The complete correlation matrix (\Cref{fig:correlation_matrix}) summarizes these findings. We observe a clear block structure in which spatial tasks (Ego, Allo, Pose, Depth) cluster together, distinct from the semantic classification block. This shows that existing semantic benchmarks are insufficient proxies for the spatial intelligence required by embodied agents. To build truly general vision systems, we must treat 3D awareness as a fundamental, distinct capability axis.

\subsection{Analysis of the mechanism of forming spatial representations} \label{sec:analysis_attention}

In this section, we analyze the mechanism of forming spatial representations in two VFMs that tend to achieve the best \our{} performance: DINO-v2 (with register tokens)~\cite{oquab2023dinov2,darcet_vision_2023} and VGGT~\cite{wang2025vggt}.
Apart from their potent spatial representations, the second reason we focus on these models is that the weights of DINO-v2 serve as an initialization for VGGT, which is subsequently trained with a 3D structure perception objective.
Therefore, the analysis of these models allows us to characterize not only which properties of representations inform spatial relation encoding but also the differences between the initial model pre-trained for semantic tasks and its version tuned for 3D perception.

To investigate why VGGT outperforms DINO-v2 on spatial tasks when using Efficient Probing, yet underperforms with Linear Probing, we analyze the internal attention mechanisms of both backbones.
We hypothesize that VGGT's 3D supervision forces the model to restructure how information flows between patches, shifting from object-centric feature extraction to relational scene encoding.

\begin{figure}[th]
    \centering
    \includegraphics[width=1.0\linewidth]{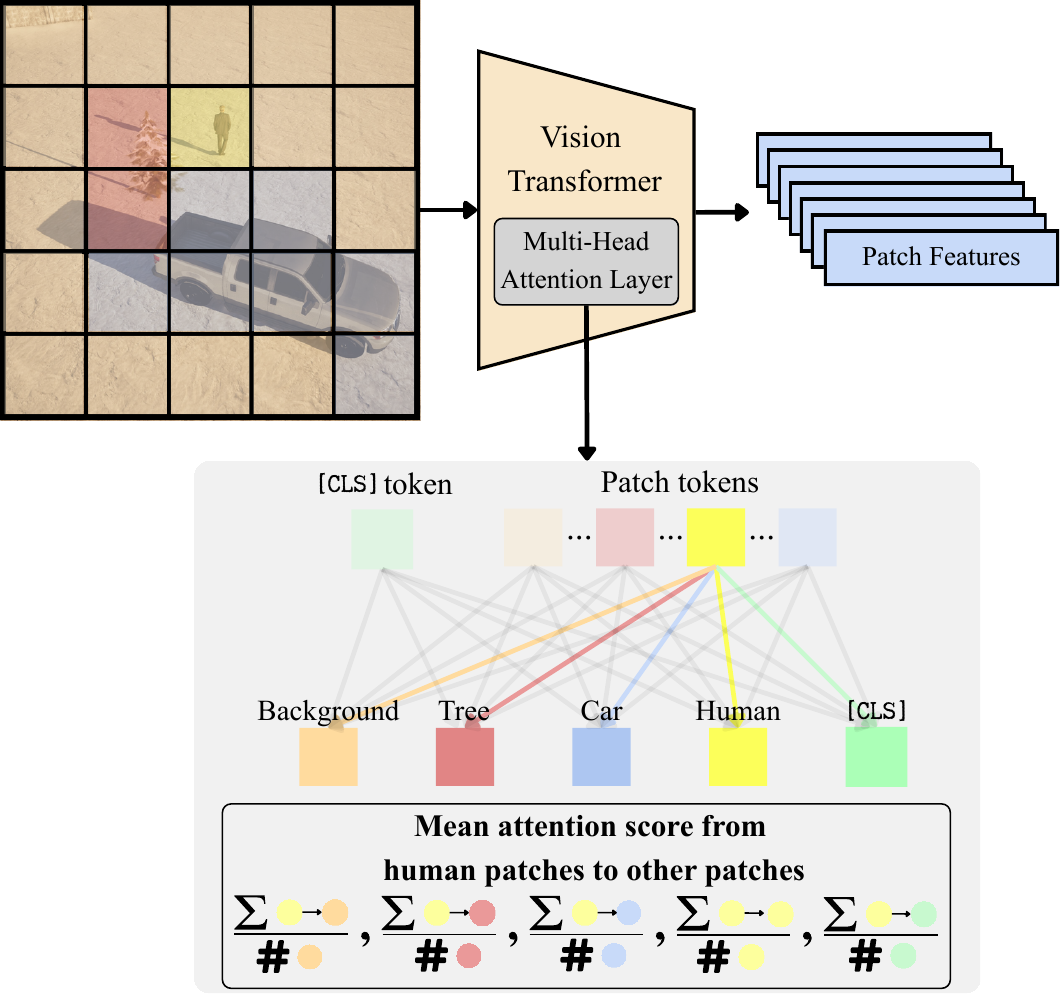}
    \caption{\textbf{Methodology for Attention Dynamics Analysis.} We illustrate the protocol for mapping patch tokens to semantic categories and calculating the inter-object attention scores. A sample input image is patchified and passed through the frozen ViT backbone to extract raw attention scores between tokens (top). We use ground-truth segmentation masks to assign semantic categories (Background, Tree, Car, Human) to these tokens. Finally, we aggregate patch tokens to compute the mean attention flow between objects (e.g., Human $\rightarrow$ Car, or Human $\rightarrow$ \cls{} token) using the formula displayed (bottom).}
    \label{fig:attention_dynamic_desc}
\end{figure}

\textbf{Methodology.} We conduct this analysis across the five evaluation environments and three objects (Tree, Truck, and Human), selecting 60 clear, non-occluded images per environment where the objects are strictly visible. Using ground-truth segmentation masks obtained from the UnrealCV API~\cite{qiu2017unrealcv}, we map every patch token to its corresponding object (Human, Tree, Truck, or Background). We then compute the mean attention weights from the source object's patches to all other regions across all heads in the frozen transformer blocks (see~\Cref{fig:attention_dynamic_desc}). We visualize the layer-wise evolution of these attention scores (averaged across heads and environments) on a logarithmic scale to account for the abundance of background patches.

\begin{figure*}[t]
    \centering
        \centering
        \includegraphics[width=\linewidth]{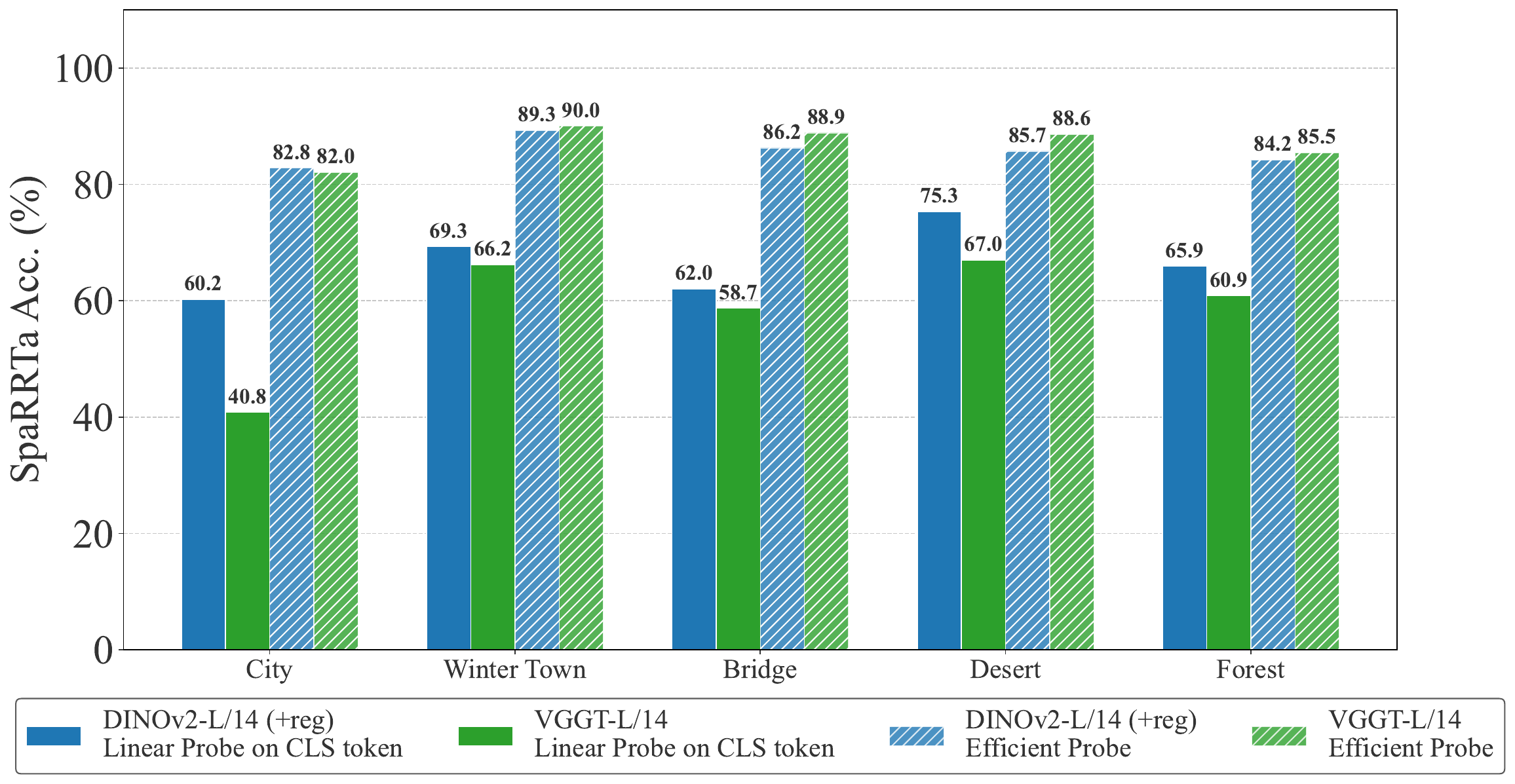}
        \caption{\textbf{Linear probing on \cls{} token vs. efficient probing.} We compare the spatial \revision{relation recognition}
        accuracy of DINO-v2 versus VGGT using two probing strategies. We report the mean accuracy averaged across both \our{}\textbf{-ego} and \our{}\textbf{-allo} tasks. DINO-v2 outperforms VGGT with linear probing on the global \cls{} token, but VGGT surpasses it with the efficient probe.}
    \label{fig:cls_vs_efficient_acc}
    \vspace{-0.3cm}
\end{figure*}

\begin{figure*}[ht!]
    \centering
    \begin{subfigure}[b]{0.60\textwidth}
        \centering
        \includegraphics[width=\linewidth]{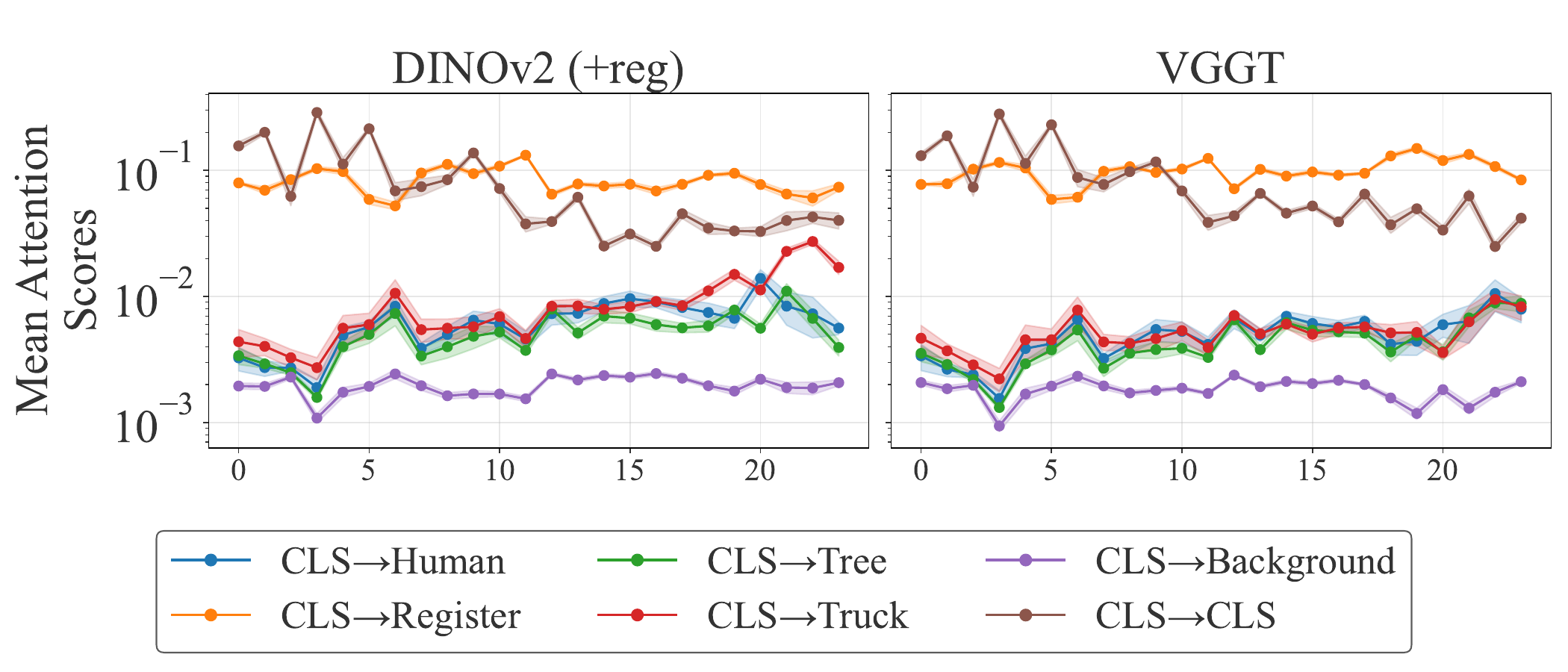}
    \end{subfigure}
    \begin{subfigure}[b]{0.365\textwidth}
        \centering
        \includegraphics[width=\linewidth]{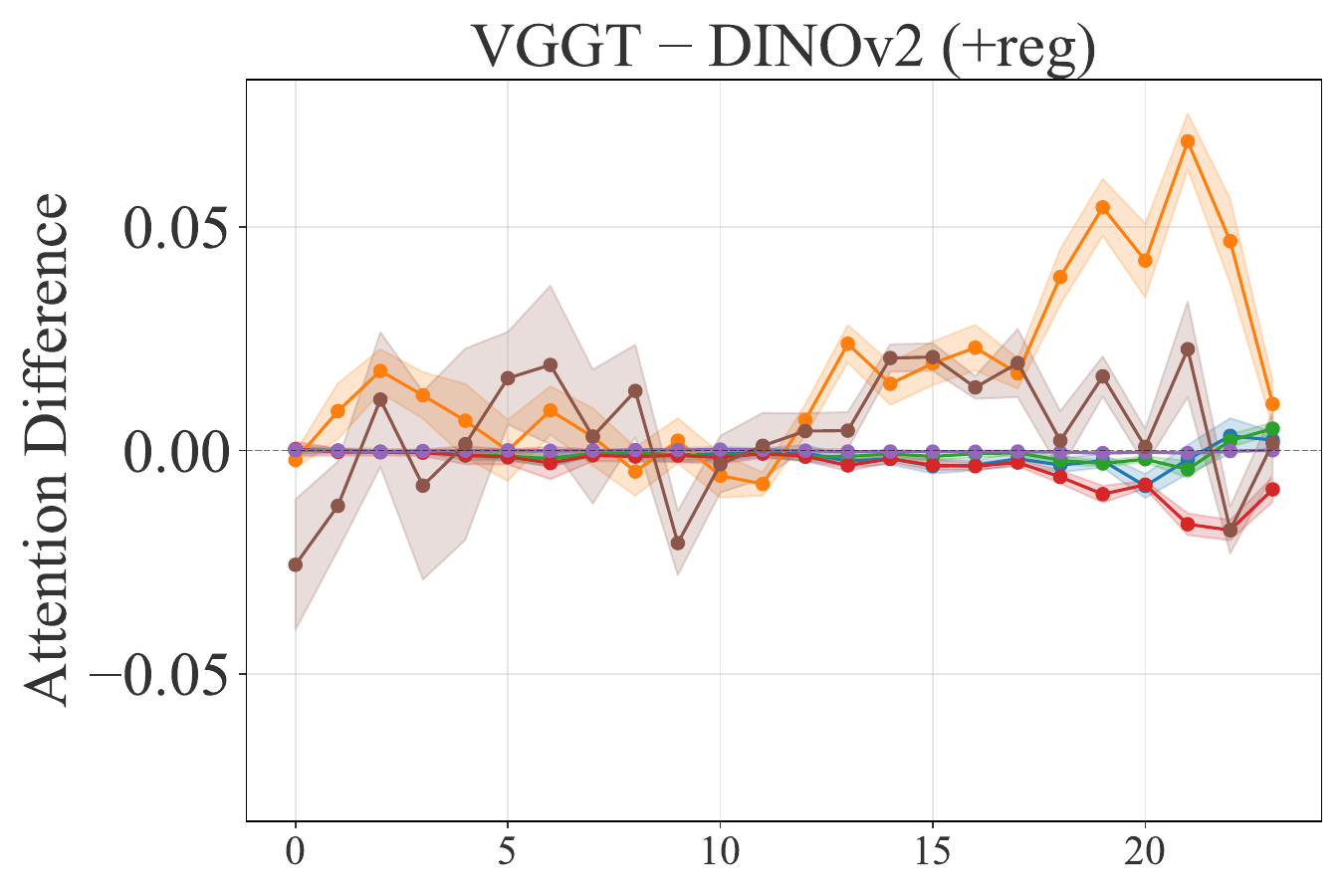}
    \end{subfigure}

    \begin{subfigure}[b]{0.60\textwidth}
        \centering
        \includegraphics[width=\linewidth]{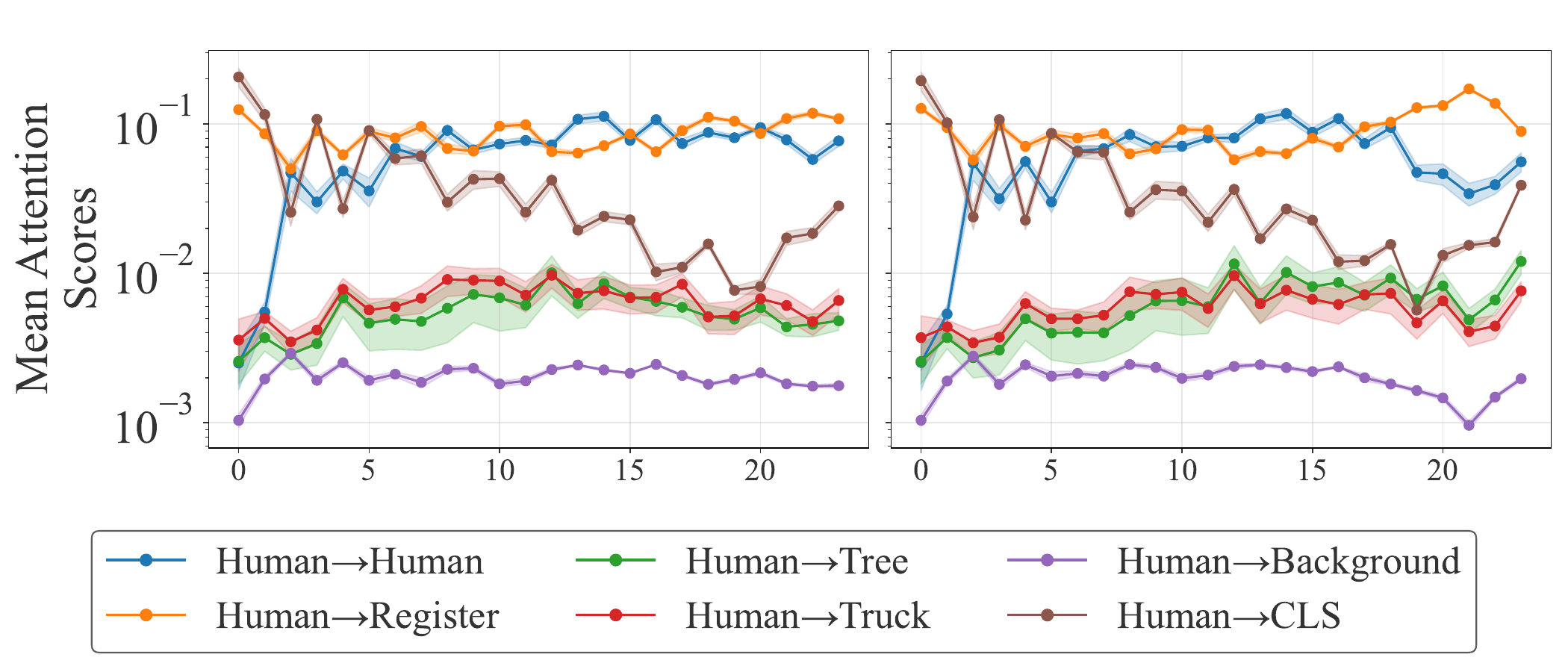}
    \end{subfigure}
    \begin{subfigure}[b]{0.365\textwidth}
        \centering
        \includegraphics[width=\linewidth]{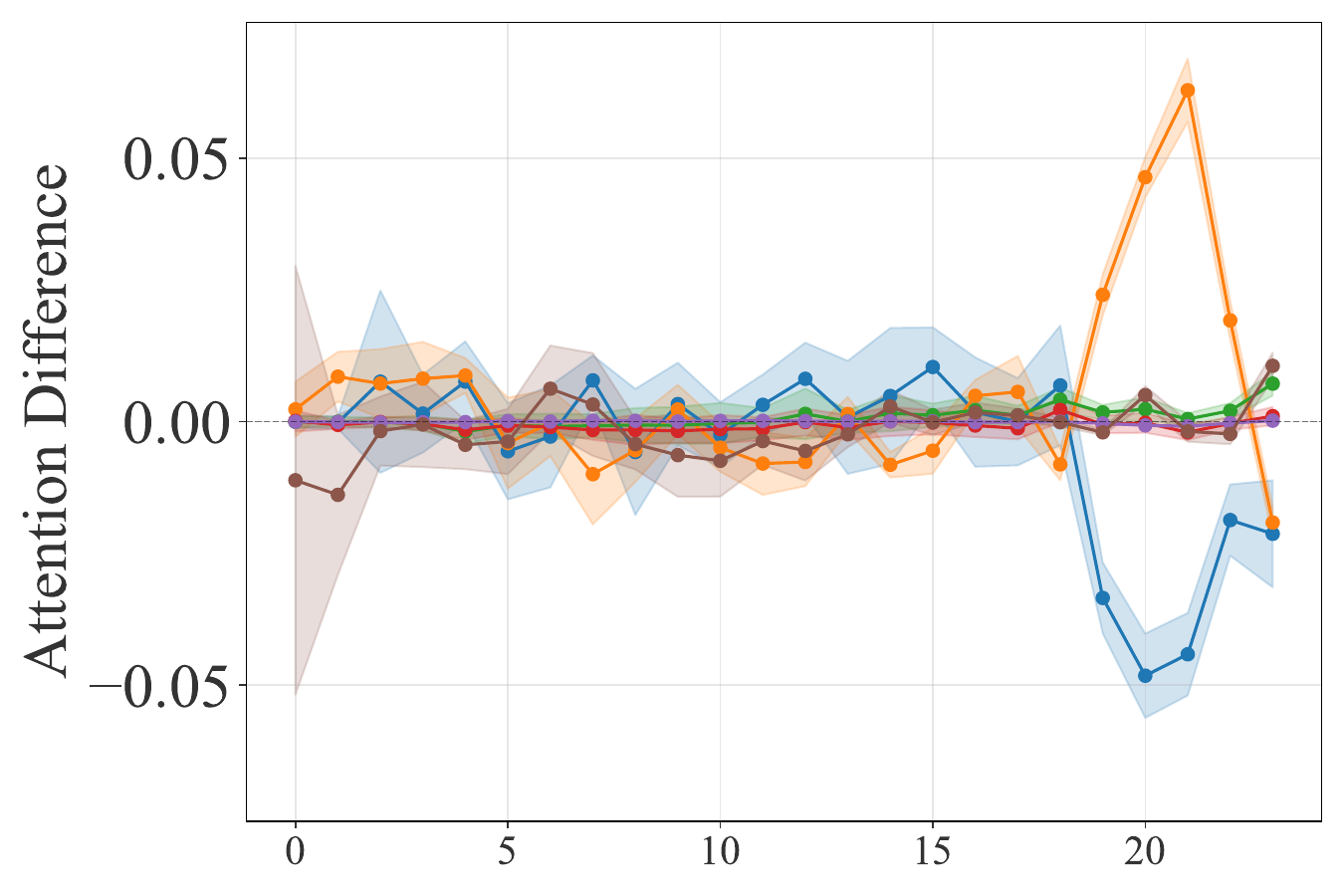}
    \end{subfigure}

    \caption{\textbf{Comparative Analysis of Attention Flow (DINO-v2 vs. VGGT).} We visualize the layer-wise mean attention scores averaged across heads and environments from the global \cls{} token (top row) and human patches (bottom row) to the rest of the patches (Human, Truck, Tree, Background, \cls{} and Register) for both DINO-v2 and VGGT backbones. The two columns on the left show absolute attention scores for DINO-v2 and VGGT, while the right column displays the differential (VGGT $-$ DINO-v2) to quantify the divergence in attention dynamics. A fundamental reorganization emerges across all token types in the deeper layers. In the bottom row, while DINO-v2 prioritizes attention from human patches to itself or to register tokens (blue and orange negative curves in the difference plot), VGGT consistently reallocates attention to other objects (green and red positive curves in the difference plot). This shows that VGGT's explicit 3D supervision forces the model to actively encode spatial dependencies between distinct entities.}
    \label{fig:diff_attn_dino_vggt}
    \vspace{-0.30cm}
\end{figure*}

\textbf{Model's attention behavior.} We observe that fine-tuning VGGT for 3D tasks induces a fundamental reorganization of information flow in the deep layers (specifically, layers 15--24). As visualized in the VGGT attention maps (\Cref{fig:diff_attn_dino_vggt}, bottom left), we observe a distinct relational shift. Attention from an object to itself (e.g., Human $\to$ Human) drops more sharply in the final layers of VGGT than in DINO-v2. At the same time, attention directed toward other objects (e.g., Human $\to$ Tree and Human $\to$ Truck) exhibits a sharper increase. This redistribution of attention is quantified in the difference plots (VGGT $-$ DINO-v2) presented in~\Cref{fig:diff_attn_dino_vggt} (bottom right). A clear structural divergence emerges: the difference in self-attention becomes negative, while the difference in cross-object attention becomes positive. This indicates that VGGT actively repurposes probability mass previously allocated to local self-refinement and redirects it to encode spatial dependencies between distinct entities. Further visualizations and analyses of the attention mechanism for the remaining object classes can be found in~\Cref{sec:extend_attention}.

\textbf{Implications for Probing Performance.} This structural reorganization provides an explanation for the probing results reported in~\Cref{sec:exp_main_results}. To validate this directly, we conducted an experiment comparing linear probing specifically on the \cls{} token versus efficient probing for VGGT and DINO-v2 (see~\Cref{fig:cls_vs_efficient_acc}). DINO-v2 outperforms VGGT when restricted to the global \cls{} token, yet VGGT surpasses when using the efficient probe. Why does the \cls{} token fail for VGGT? Our analysis in~\Cref{fig:diff_attn_dino_vggt} (Top Row) offers the answer. In DINO-v2, the \cls{} token maintains high attention to the objects (e.g., the Truck). This allows the single global vector to retain enough scene context for a linear probe to succeed. In contrast, VGGT's \cls{} token shifts its attention budget toward register tokens. This suggests that during 3D fine-tuning, VGGT offloads global geometric parameters to registers and distributes object-relative spatial details across the local patch tokens. Consequently, a linear probe on the single \cls{} token fails to access this distributed information, whereas efficient probing, which aggregates across the entire patch sequence via attention, successfully retrieves the rich spatial context, allowing VGGT to ultimately surpass DINO-v2 in \our{}.

\section{Conclusion}
\label{sec:conclusion}

Visual Foundation Models (VFMs) are a crucial component of embodied applications, encoding spatial information about an agent's surroundings.
However, good semantic recognition alone is insufficient for selecting appropriate actions in physical space.
In this paper, we introduced \ourfull{} (\our{}), a benchmark for evaluating whether a VFM represents spatial relations between objects in \revision{static} images.

Our results show that while modern VFMs contain non-trivial spatial information, this information is not uniformly accessible: spatial relations are primarily encoded at the patch level and are easily obscured by global pooling. As a consequence, standard linear evaluation substantially underestimates spatial awareness, particularly in settings that require perspective-dependent reasoning.
We further show that decoding relative (allocentric) spatial relations remains challenging across model families, and that explicit 3D supervision primarily improves patch-level geometric structure rather than global representations.

By isolating spatial relation recognition from semantic classification and metric estimation, \our{} provides a complementary evaluation axis that is not captured by existing benchmarks.
In the future, we plan to explore the fully controllable synthetic data engine used in \our{} for modeling \revision{agent} interactions with the environment and training spatially-aware, \revision{action-conditioned} world models.
We hope that \our{} will serve as a diagnostic tool for analyzing spatial representations, and as a guide for developing future vision models with stronger spatial awareness. \revision{Spatially-aware VFMs can then serve as a base for building more precise vision-action and vision-language-action models, simplifying the training regime and reducing the number of reinforcement learning episodes needed to achieve required performance on downstream embodied tasks.}

\revision{
\section*{Limitations}
\our{} focuses on static object-to-object directional relations in single images, under egocentric and allocentric viewpoints.
While these relations are a fundamental building block of spatial reasoning, \our{} does not evaluate other important aspects of spatial intelligence, such as metric distance estimation, containment, support, occlusion reasoning, object orientation, multi-object compositional reasoning, or temporal and action-conditioned understanding.

During the \our{} evaluation, probes are trained and tested within fixed semantic triples of source, target, and viewpoint objects.
As a result, the reported results demonstrate the recoverability of spatial relation information from frozen VFM representations within these predefined object configurations, but do not directly measure generalization to novel object combinations.

\our{} test cases are generated using a rendering engine, and this domain overlap may constitute an unfair advantage for models pretrained on synthetic data~\cite{wang2025vggt}.
However, our results do not suggest that synthetic pretraining alone explains performance: natural-image SSL models such as DINO-v2, DINOv3, and MAE remain highly competitive, while CroCo-style models do not consistently outperform them.
}

\section*{Acknowledgements}

This research has been supported by the flagship project
entitled "Artificial Intelligence Computing Center Core Facility" from the Priority Research Area DigiWorld under
the Strategic Programme Excellence Initiative at the Jagiellonian University.
The work of Turhan Can Kargin, Adam Pardyl, and Bartosz Zieli\'nski was supported by National Science Centre (Poland) grant number 2023/50/E/ST6/00469.
The research of Marcin Przewi\k{e}\.{z}likowski
was supported by the National Science Centre (Poland),
grant no. 2023/49/N/ST6/03268.
We gratefully acknowledge Polish high-performance computing infrastructure PLGrid (HPC Center: ACK Cyfronet AGH) for providing computer facilities and support within computational grant no. PLG/2025/018312.

\revision{
\section*{Data Availability Statement}
The complete \our{} benchmark dataset, including all generated synthetic images and geometric metadata, is publicly available at \href{https://huggingface.co/datasets/turhancan97/SpaRRTa}{huggingface.co/datasets/turhancan97/SpaRRTa}. To facilitate reproducibility and further research, we also release the full source code for the Unreal Engine data generation pipeline and the model evaluation framework at \href{https://github.com/gmum/SpaRRTa}{github.com/gmum/SpaRRTa}. Additional materials and experimental results are available from the corresponding author upon reasonable request.}

\bibliography{ref}

\appendix

\section{Dataset Generation Methodology}

\begin{figure*}[t]
    \centering
    \includegraphics[width=0.75\linewidth]{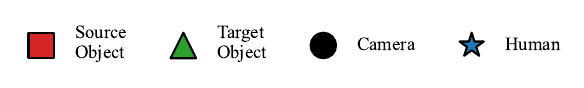}
    \begin{subfigure}[b]{0.27\textwidth}
        \centering
        \includegraphics[width=\linewidth]{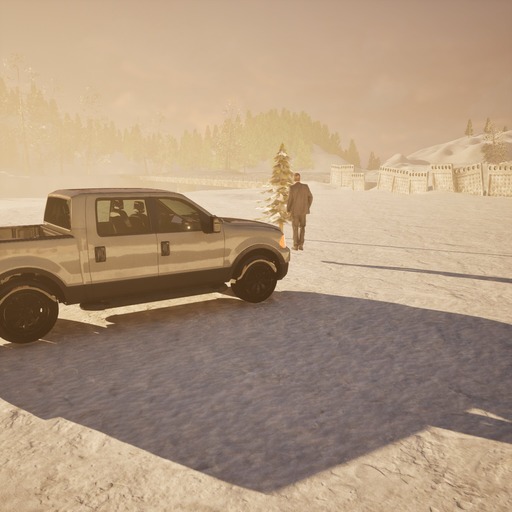}
    \end{subfigure}
    \hfill
    \begin{subfigure}[b]{0.30\textwidth}
        \centering
        \includegraphics[width=\linewidth]{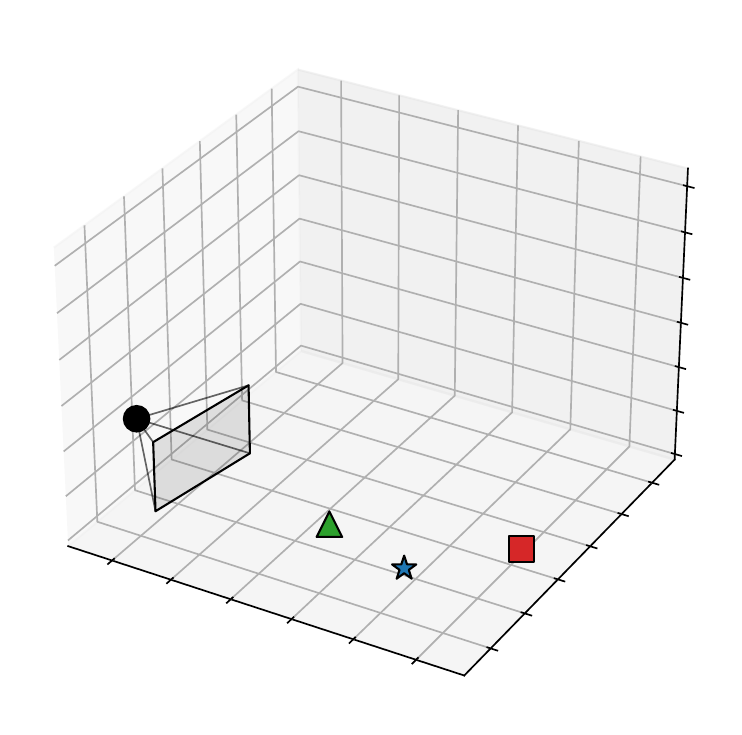}
    \end{subfigure}
    \hfill
    \begin{subfigure}[b]{0.30\textwidth}
        \centering
        \includegraphics[width=\linewidth]{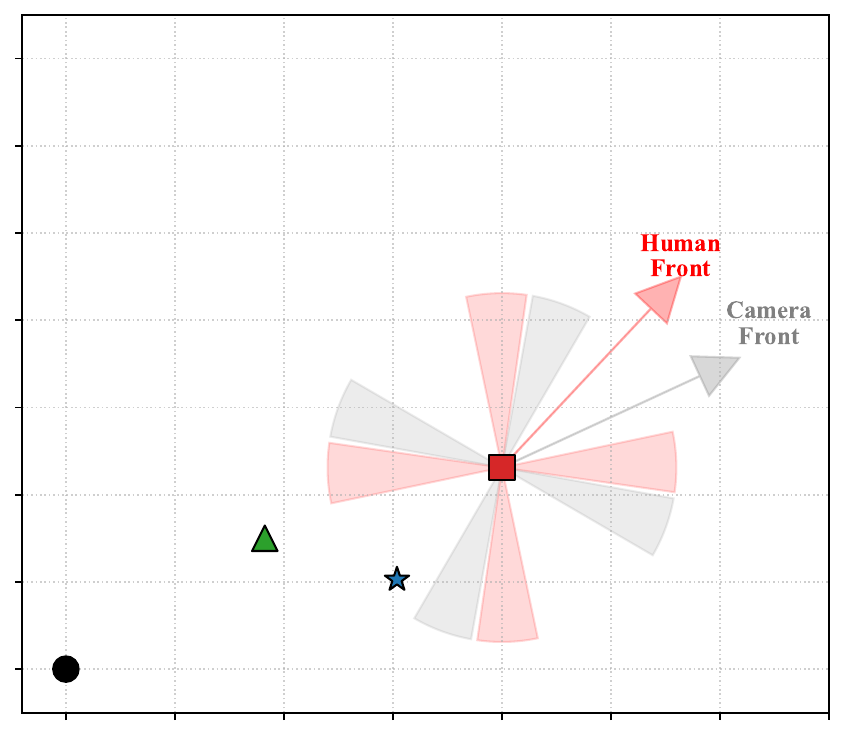}
    \end{subfigure}

    \vspace{0.25em}

    \begin{subfigure}[b]{0.27\textwidth}
        \centering
        \includegraphics[width=\linewidth]{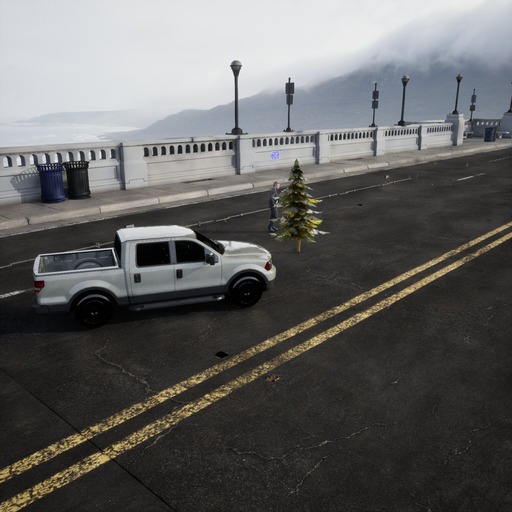}
        \caption{Input RGB image}
        \label{fig:scene_env}
    \end{subfigure}
    \hfill
    \begin{subfigure}[b]{0.30\textwidth}
        \centering
        \includegraphics[width=\linewidth]{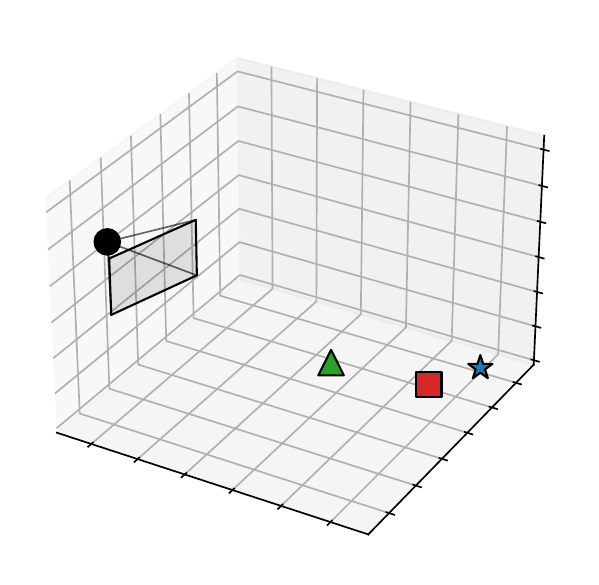}
        \caption{3D scene layout}
        \label{fig:scene_env_3d}
    \end{subfigure}
    \hfill
    \begin{subfigure}[b]{0.30\textwidth}
        \centering
        \includegraphics[width=\linewidth]{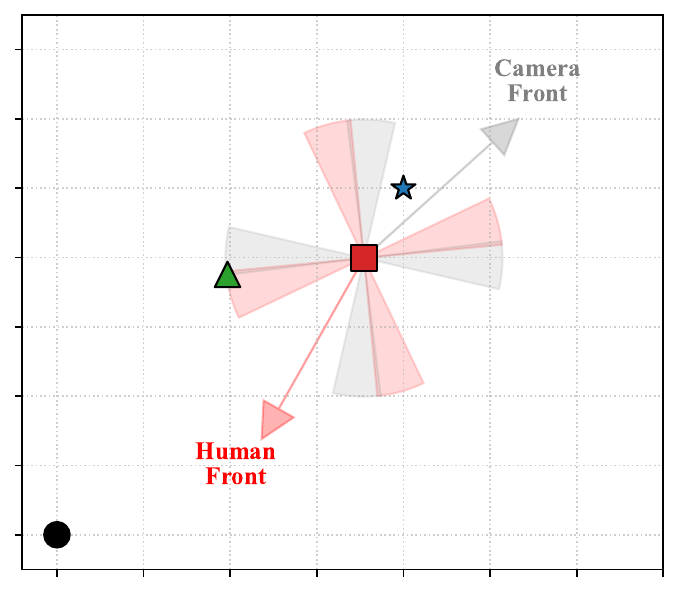}
        \caption{2D scene logic}
        \label{fig:scene_env_2d}
    \end{subfigure}

    \caption{\textbf{Visualization of Task Geometry and Ambiguity Filtering.} We present the geometric construction of \our{} samples, showing the rendered RGB input (\textbf{a}), the global 3D arrangement of assets (\textbf{b}), and the 2D logic for defining the egocentric (gray) and allocentric (red) spatial positions (\textbf{c}).}
    \label{fig:scene_examples_with_2d_3d}
\end{figure*}

\subsection{Core Design Principles and Controllability}\label{subsec:design_principles}

The design of \our{} is governed by the need to disentangle \revision{spatial relation recognition}
from low-level visual pattern matching while ensuring mathematically rigorous ground-truth labels. Unlike real-world datasets, where camera angles and object placements are biased by human photographers, our synthetic approach enables precise control over three critical axes: geometric validity, environmental diversity, and task complexity.

\textbf{Defining spatial relations and ambiguity.} As illustrated in~\Cref{fig:scene_examples_with_2d_3d}, the core challenge of \our{} lies in resolving spatial relations relative to a specific coordinate frame of the viewpoint. \Cref{fig:scene_env} shows views from the Unreal Engine environment with the source object (tree), target object (truck), and human agent. \Cref{fig:scene_env_3d} shows the three-dimensional spatial arrangement of the camera frustum, objects, and human agent in world coordinates. For the egocentric task (\Cref{fig:scene_env_2d}, gray arrow), the front direction is defined by the vector from the camera to the source object. For the allocentric task (\Cref{fig:scene_env_2d}, red arrow), the front is defined by the vector from the human agent to the source object. A fundamental challenge in spatial classification is defining boundaries between classes (e.g., when does Left become Front?). To ensure the benchmark's validity, we handle those geometric boundaries. We define ambiguity zones as conical regions (shaded red and gray in~\Cref{fig:scene_env_2d}) centered around the $45^\circ$, $135^\circ$, $225^\circ$, and $315^\circ$ diagonals relative to the viewpoint's forward vector. Any sample in which the target object falls within these zones is automatically filtered from the dataset. This guarantees that the ground-truth labels (left, right, front, back) are unambiguous and robust to minor variations in pose estimation, forcing the model to learn precise spatial boundaries rather than guessing on the margins.

\textbf{Asset curation and environmental diversity.} To test the robustness of spatial representations, we use five distinct environments in the \our{} benchmark. We curate specific asset sets for each environment to maintain semantic coherence, ensuring objects appear in natural contexts while preserving geometric consistency. Crucially, we select isotropic objects (rock, tree, traffic cone) as source objects. Their rotationally symmetric nature simplifies the definition of relative positions, ensuring that the spatial relation (e.g., left of the tree) is defined purely by the target's position relative to the source's center, minimizing ambiguity caused by the source object's own orientation. The assets used in our benchmark are detailed in Table~\ref{tab:asset_classes}.

\begin{table}[h]
    \centering
    \resizebox{\columnwidth}{!}{%
    \begin{tabular}{@{}lll@{}}
        \toprule
        \textbf{Environment} & \textbf{Asset Class} & \textbf{Source} \\
        \midrule
\multirow{3}{*}{\textbf{Global}} & Humans & Fab~\cite{asset_human} \\
        & Vehicle (Car and Taxi) & Fab~\cite{asset_vehicle} \\
        & Tree & Environment Pack~\cite{electric_dreams} \\
        \midrule
\multirow{4}{*}{\textbf{Forest}} & Brown Bear & Sketchfab~\cite{asset_bear} \\
        & Fox & Sketchfab~\cite{asset_fox} \\
        & Camping Tent & Sketchfab~\cite{asset_tent} \\
        & Rocks & Environment Pack~\cite{desert_env, electric_dreams} \\
        \midrule
\multirow{4}{*}{\textbf{Desert}} & Camel & Fab~\cite{asset_camel} \\
        & Barrel & Sketchfab~\cite{asset_barrel} \\
        & Cactus & Sketchfab~\cite{asset_cactus} \\
        & Rocks & Environment Pack~\cite{desert_env, electric_dreams} \\
        \midrule
\multirow{3}{*}{\textbf{Winter Town}} & Husky & Sketchfab~\cite{asset_husky} \\
        & Deer & Sketchfab~\cite{asset_deer} \\
        & Snowman & Sketchfab~\cite{asset_snowman} \\
        \midrule
        \multirow{2}{*}{\textbf{Bridge}} & Bicycle & Fab~\cite{asset_bike} \\
        & Trash Can & Environment Pack~\cite{bridge_env} \\
        \midrule
        \multirow{3}{*}{\textbf{City}} & Motorcycle & Sketchfab~\cite{asset_vespa} \\
        & Traffic Cone & Sketchfab~\cite{asset_traffic_cone} \\
        & Fire Hydrant & Sketchfab~\cite{asset_fire_hydrant} \\
        \bottomrule
    \end{tabular}
    }
    \caption{\textbf{\our{} Asset Registry.} We list the specific 3D assets used across environments. Sources denote the origin of the 3D model.}
    \label{tab:asset_classes}
\end{table}

\textbf{Dataset size and generalization.} To determine the optimal data scale, we analyzed model generalization across tasks. We found that \textbf{5,000 images} per environment were sufficient for models to generalize to the egocentric task. However, the allocentric task, which requires learning a more complex perspective transformation, exhibited signs of overfitting at this scale. Consequently, we increased the allocentric dataset size to \textbf{10,000 images} per environment, ensuring sufficient diversity for the model to learn the underlying geometric rules rather than memorize specific scene configurations.

\subsection{Rendering Pipeline Details}\label{subsec:render_details}

Our approach leverages the native Unreal Engine Python API~\cite{unreal_python_api} and the UnrealCV plugin~\cite{qiu2017unrealcv} to programmatically control object placement, physics validation, and camera logic. The pipeline outputs raw RGB renders, serializes the precise 6-DoF state (position: $x, y, z$ and rotation: $\phi, \theta, \psi$) of every task-critical object and the camera into JSON metadata files, and generates pixel-perfect instance masks of the source, target, and human objects.

\textbf{Terrain adaptation via raycasting.} We employ a rejection sampling strategy to ensure physical plausibility and visual clarity. We implement a physics-aware placement logic to handle uneven terrain (e.g., the Forest and Desert environments) without objects floating or clipping. For every object coordinate, the pipeline performs a vertical line trace to detect the exact Z-height of the landscape geometry. Objects are then spawned at the detected ground level, with their specific bounding box offsets applied to ensure flush contact.

\textbf{Visibility and occlusion control.} We enforce visibility constraints at the generation level to guarantee that all task-critical objects are captured. Candidate object positions are validated against the camera's viewing frustum by computing the angular deviation of each object relative to the camera's optical axis. We enforce a strict constraint where the maximum angular offset must remain within the camera's horizontal field of view to prevent objects from being cut off at the frame edges. Furthermore, to ensure appropriate scene composition, we reject configurations where objects appear too clustered or distant, forcing the camera to capture a meaningful spatial spread. Finally, we utilize AABB (Axis-Aligned Bounding Box)~\cite{schneider2003481} overlap checks to prevent inter-object collisions.

\textbf{Camera configuration and object placement.} We standardize the optical setup across all environments to prevent intrinsic camera parameters from becoming a confounding variable. All scenes are rendered using a simulated 50mm lens on a 50mm sensor width, resulting in a consistent horizontal field of view. The camera positioning follows a hierarchical stochastic process. Object coordinates are sampled around a randomly selected center point, with a distance limit to ensure they remain close enough to be distinguishable. The camera is then spawned within a broad area surrounding this group, with its height randomized relative to the average elevation of the objects. Finally, the pipeline computes the geometric centroid of the relevant scene objects (source, target, human agent) and dynamically locks the camera's rotation to center this cluster, ensuring the primary task elements remain the focal point.

\section{Additional Experimental Results}\label{sec:additional_results}

\subsection{Transfer to an Unseen Semantic Triple}
\label{sec:appendix_cross_triple_transfer}
\revision{
The main \our{} benchmark uses triple-specific probes, so the core results measure recoverability of spatial information within a fixed object configuration rather than transfer to unseen semantic combinations. To address this limitation, we added a transfer experiment to an unseen semantic triple in the Desert environment. In this experiment, we train the probe on the Camel, Cactus, and Human triple and test it on a new Camel, Barrel, and Human triple. Only the camel remains the same; the cactus is replaced by a barrel, and a different human agent is used. We report results for both the egocentric and allocentric settings. We conduct this test using Efficient Probing only, since it gives the most representative results in the main benchmark. We report both zero-shot transfer and few-shot adaptation using only samples from the new triple. In the zero-shot setting, the probe is trained on the source triple (Camel, Cactus, and Human) and evaluated directly on the new target triple (Camel, Barrel, and Human) without any target-side training. In the few-shot setting, the probe is adapted to a small labeled target set. For $K=10, 25, 50, 100$, we initialize the probe weights from the zero-shot solution and fine-tune them on $K$ target samples. For $K=500$, we additionally compare this transfer-initialized adaptation with training the same probe from randomly initialized weights on the same target set.

The results in Table~\ref{tab:cross_triple_transfer} show that initializing from the zero-shot probe improves adaptation over training the target probe from randomly initialized weights, especially in the low-shot regime. In particular, with only 25 target images, transfer-initialized adaptation already matches or nearly matches the performance of training with 500 images from randomly initialized probe weights, and in the allocentric setting it already exceeds that baseline for both backbones. The main takeaway is that the learned spatial structure is transferable: even a small amount of target data is enough to adapt the probe effectively once it is initialized from a related source triple. Overall, these results support the claim that \our{} is not only measuring recoverability within a fixed semantic configuration. Spatial structure learned from one triple can transfer to a different target triple and be adapted with limited target data.}

\begin{table*}[t]
\centering
\small
\setlength{\tabcolsep}{3pt}
\resizebox{0.98\textwidth}{!}{%
\begin{tabular}{llccccccc}
\toprule
\textbf{Backbone} & \textbf{Viewpoint} & \textbf{ZS} & \textbf{$K=10$} & \textbf{$K=25$} & \textbf{$K=50$} & \textbf{$K=100$} & \multicolumn{2}{c}{\textbf{$K=500$}} \\
\cmidrule(lr){8-9}
 &  &  &  &  &  &  & \textbf{Transfer} & \textbf{Random init.} \\
\midrule
\multirow{2}{*}{DINO-v2 reg (L/14)} & Egocentric & 61.04 & 68.40 & 81.95 & 84.66 & 88.65 & 93.56 & 82.42 \\
 & Allocentric & 44.89 & 40.37 & 49.54 & 53.39 & 52.32 & 54.98 & 47.14 \\
\midrule
\multirow{2}{*}{VGGT (L/14)} & Egocentric & 79.75 & 79.45 & 82.55 & 86.81 & 93.25 & 97.63 & 83.20 \\
 & Allocentric & 36.52 & 49.14 & 52.32 & 54.71 & 54.85 & 62.22 & 48.68 \\
\bottomrule
\end{tabular}
}
\caption{\revision{\textbf{Transfer to an unseen semantic triple in the Desert environment.} We train the probe on the Camel, Cactus, and Human triple and test it on a new Camel, Barrel, and Human triple while keeping the backbone frozen. Results are shown for Efficient Probing. \textbf{ZS} denotes zero-shot transfer. The columns for $K$ report target-only few-shot adaptation with transfer initialization from the zero-shot probe. Under $K=500$, we report both transfer-initialized adaptation and a non-transfer baseline, in which the same probe is trained from randomly initialized weights on the new target triple. This experiment shows that the benchmark is not limited to fixed semantic triples, since the spatial structure learned on one triple can transfer to a new object combination.}}
\label{tab:cross_triple_transfer}
\end{table*}

\subsection{Real-World Control Set and Probe Data Requirements}
\label{sec:appendix_scaling_realworld}

\begin{figure}[t]
\centering
\begin{subfigure}[t]{0.48\columnwidth}
    \centering
    \includegraphics[width=\linewidth]{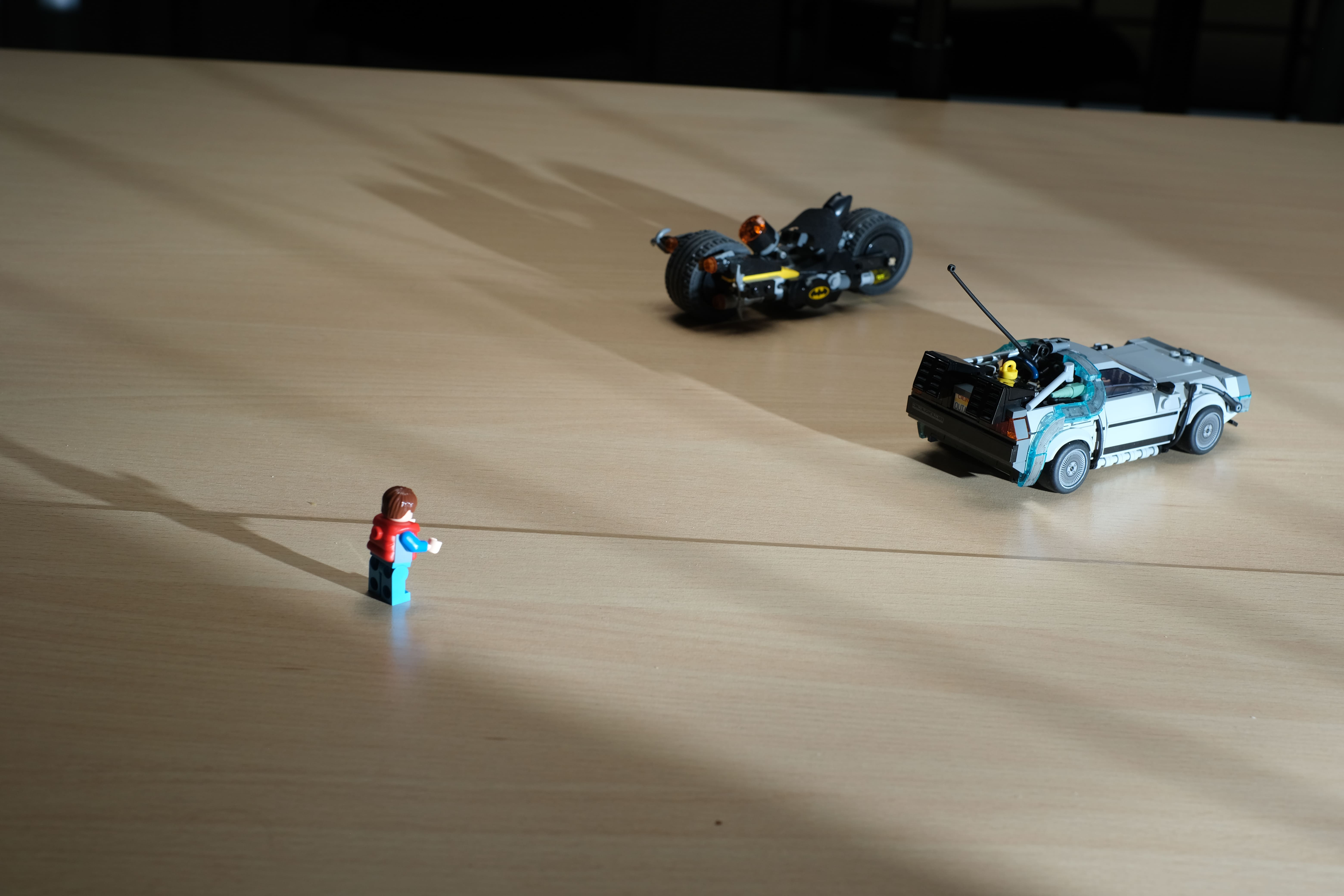}
    \caption{Left}
\end{subfigure}
\hfill
\begin{subfigure}[t]{0.48\columnwidth}
    \centering
    \includegraphics[width=\linewidth]{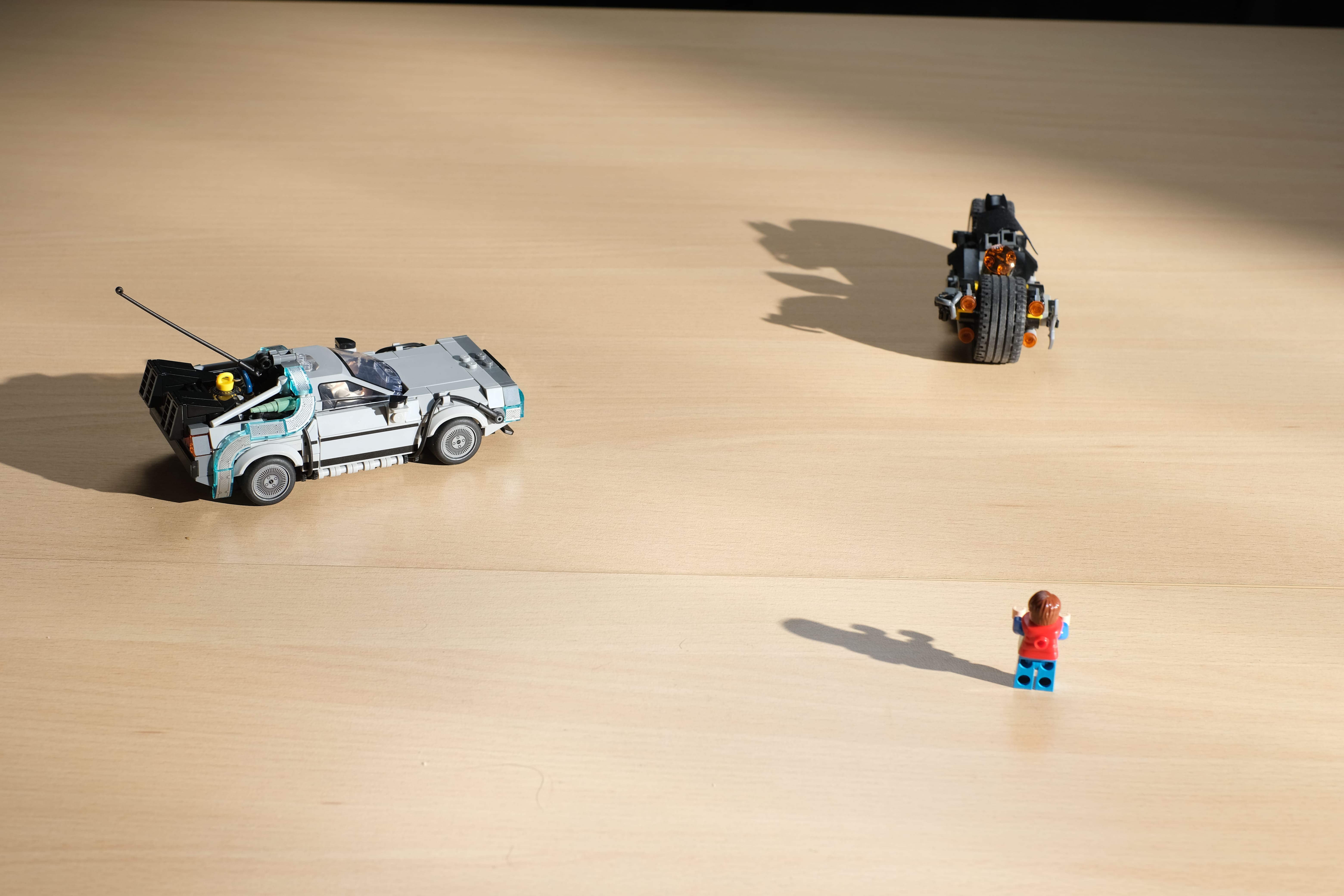}
    \caption{Right}
\end{subfigure}

\vspace{0.4em}

\begin{subfigure}[t]{0.48\columnwidth}
    \centering
    \includegraphics[width=\linewidth]{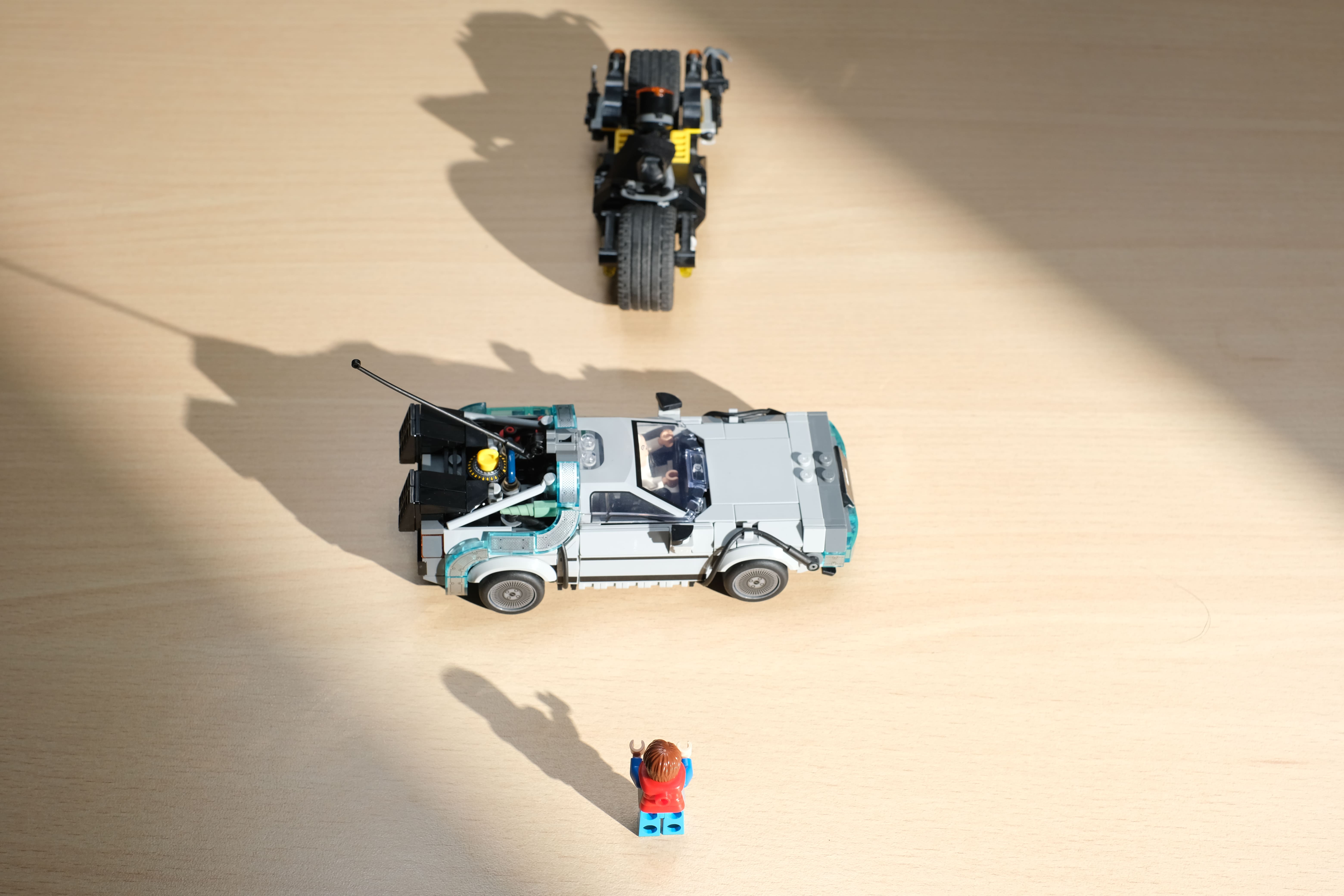}
    \caption{Front}
\end{subfigure}
\hfill
\begin{subfigure}[t]{0.48\columnwidth}
    \centering
    \includegraphics[width=\linewidth]{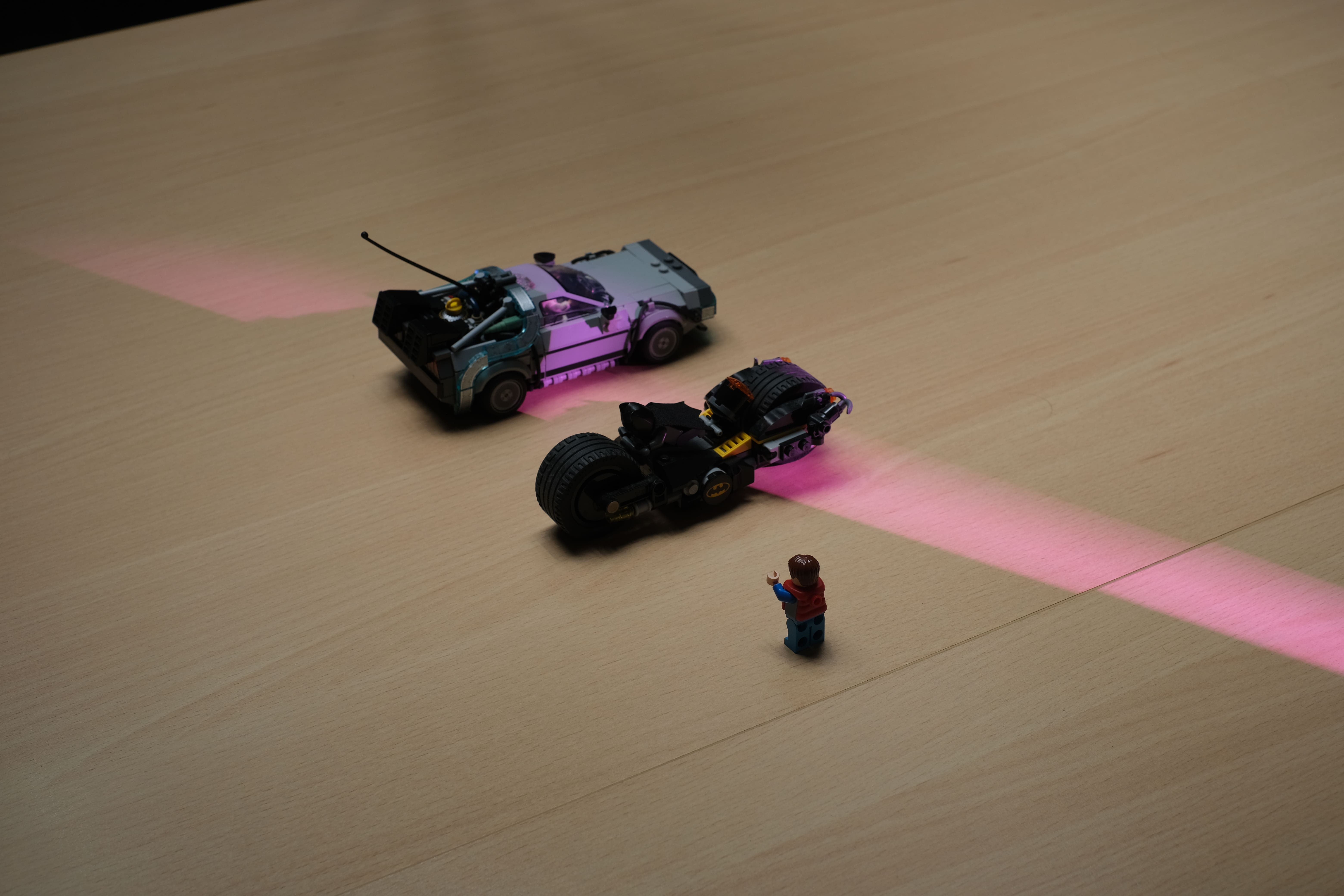}
    \caption{Back}
\end{subfigure}
\caption{\revision{\textbf{Examples from the real-world control set.} We show representative images for the four directional labels used in the real-world evaluation, which are all defined from the human viewpoint. The dataset uses three Lego objects under varied lighting and background conditions.}}
\label{fig:lego_real_world_examples}
\end{figure}

\revision{
Although \our{} uses photorealistic synthetic scenes with controlled annotations, a natural question is whether the recovered spatial signal transfers beyond the rendering domain.
To assess this, we collect a targeted real-world allocentric control set.
The dataset contains 1,000 real photographs of three Lego objects, with balanced directional labels defined from the human viewpoint (see~\Cref{fig:lego_real_world_examples}).
During acquisition, we vary illumination, background-light color, and related appearance factors, while keeping the object identities and allocentric relation structure controlled.
This setup allows us to test whether probes trained in the synthetic \our{} setting retain their spatial relation recognition ability on real images.
}

\revision{
Because the images were acquired sequentially by a human operator, random image-level splitting could introduce leakage by placing visually similar consecutive captures into different folds.
We therefore use a leakage-aware temporal split within each label block: the first 80\% of images are used for training, the next 10\% for validation, and the final 10\% for testing.
We evaluate two real-world adaptation regimes.
In the \textbf{transfer} setting, the probe is initialized from synthetic Unreal training and then adapted on the real-world training split.
In the \textbf{random initialization} setting, the same probe architecture is trained from randomly initialized weights on the real-world split.
}

\revision{
The results in~\Cref{tab:real_world_holdout_control} show that synthetic-to-real transfer improves over training the probe from random initialization on the real-world split.
The strongest transfer result is obtained by VGGT, followed closely by MAE, while the strongest randomly initialized result is also obtained by VGGT, followed by DINO.
These results indicate that the spatial information recovered by \our{} probes is not restricted to Unreal-rendered scenes, but can also support real-image allocentric relation recognition under controlled conditions.
}

\begin{figure}[t]
    \centering
    \includegraphics[width=\linewidth]{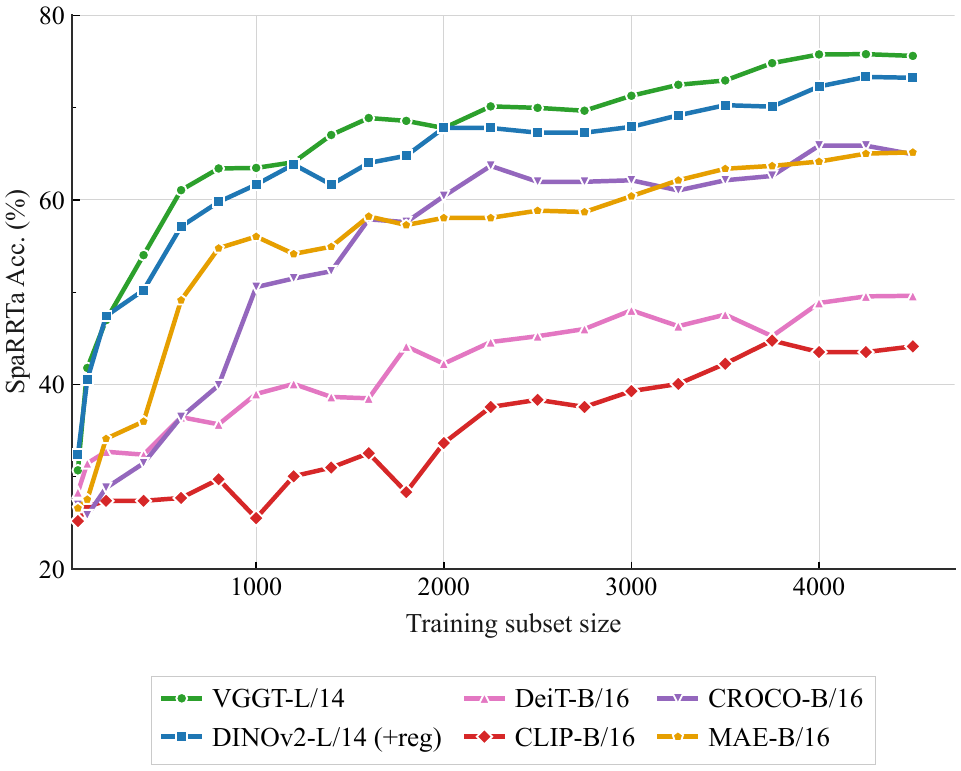}
    \caption{\revision{\textbf{Scaling behavior of probe accuracy in the allocentric task.} We vary the training subset size and report the resulting \our{} accuracy for representative models in the human-perspective setting.}}
    \label{fig:scaling_allo}
\end{figure}

\revision{
At the same time, the real-world control set highlights an important practical constraint of probe-based evaluation.
Because \our{} evaluates frozen visual representations through supervised probes, reliable model comparison depends on having sufficiently large train/validation/test splits.
To quantify this dependence, we perform a scaling analysis on the synthetic \our{} images, in which we vary the number of training examples used to train the probe and measure accuracy for six representative models on the allocentric task.
As shown in~\Cref{fig:scaling_allo}, performance continues to improve substantially as the training set grows.
In the low-data regime, accuracies are far from saturation and model rankings are less stable, indicating that small manually collected datasets can underestimate the spatial information available in frozen representations.
This motivates the use of synthetic generation in the main \our{} benchmark: it provides the scale, balance, and precise annotations needed for stable probe-based comparison, while the real-world control set verifies that the recovered spatial signal transfers beyond the rendering domain.
}

\begin{table}[t]
\centering
\small
\resizebox{\columnwidth}{!}{%
\begin{tabular}{lcc}
\toprule
\textbf{Model} & \textbf{Transfer} & \textbf{Random init.} \\
\midrule
DINO & \cellcolor[rgb]{0.819,0.931,0.750}61.90 & \cellcolor[rgb]{0.766,0.984,0.750}\underline{63.30} \\
DINO-v2 (B/14) & \cellcolor[rgb]{0.950,0.800,0.750}52.29 & \cellcolor[rgb]{0.929,0.821,0.750}44.12 \\
DINO-v2 reg (B/14) & \cellcolor[rgb]{0.925,0.825,0.750}54.13 & \cellcolor[rgb]{0.926,0.824,0.750}44.45 \\
DINO-v2 reg (L/14) & \cellcolor[rgb]{0.887,0.863,0.750}56.88 & \cellcolor[rgb]{0.898,0.852,0.750}47.71 \\
DINOv3 & \cellcolor[rgb]{0.797,0.953,0.750}63.49 & \cellcolor[rgb]{0.875,0.875,0.750}50.46 \\
VGGT (L/14) & \cellcolor[rgb]{0.750,1.000,0.750}\textbf{66.97} & \cellcolor[rgb]{0.750,1.000,0.750}\textbf{65.14} \\
SPA & \cellcolor[rgb]{0.862,0.888,0.750}58.72 & \cellcolor[rgb]{0.930,0.820,0.750}44.04 \\
CroCo & \cellcolor[rgb]{0.925,0.825,0.750}54.13 & \cellcolor[rgb]{0.914,0.836,0.750}45.87 \\
CroCo v2 & \cellcolor[rgb]{0.825,0.925,0.750}61.47 & \cellcolor[rgb]{0.938,0.812,0.750}43.12 \\
CLIP & \cellcolor[rgb]{0.962,0.788,0.750}51.38 & \cellcolor[rgb]{0.898,0.852,0.750}47.71 \\
DeiT & \cellcolor[rgb]{1.000,0.750,0.750}48.62 & \cellcolor[rgb]{0.945,0.805,0.750}42.20 \\
MAE & \cellcolor[rgb]{0.762,0.988,0.750}\underline{66.06} & \cellcolor[rgb]{0.930,0.820,0.750}44.04 \\
MaskFeat & \cellcolor[rgb]{0.937,0.813,0.750}53.21 & \cellcolor[rgb]{1.000,0.750,0.750}35.78 \\
\bottomrule
\end{tabular}%
}
\caption{\revision{\textbf{Real-world allocentric dataset evaluation.} We collect 1,000 real photographs with balanced directional labels for the human-viewpoint allocentric task using three Lego objects under varied lighting and background conditions. We report Efficient Probing only. \textbf{Transfer} initializes the probe from synthetic Unreal training and then adapts it on the real-world training split. \textbf{Random init.} trains the same probe from randomly initialized weights on the real-world training split.}}
\label{tab:real_world_holdout_control}
\end{table}

\subsection{Viewpoint-Object Semantics}
\label{sec:appendix_viewpoint_semantics}

\revision{In the allocentric setup, the viewpoint object is always a human. This may introduce a semantic bias, since differences in pretraining exposure to human-centric imagery could affect performance independently of the intended perspective-taking challenge. To test this, we conducted a control experiment in the Desert environment in which we replaced the human viewpoint object with a non-human object (camel) while keeping the allocentric geometric protocol unchanged.}

\revision{For this analysis, we evaluated models using only efficient probing, since it is the strongest probing strategy in the main benchmark and provides the most reliable estimate of allocentric spatial capability. The results are reported in Table~\ref{tab:camel_viewpoint_control}. We observe that the absolute performance changes for several models, which indicates that the semantic class of the viewpoint object affects task difficulty to some extent. However, the main conclusions remain stable: VGGT remains the strongest model, and the overall hierarchy between stronger and weaker model families is broadly preserved. This suggests that the allocentric challenge is not solely an artifact of using a human reference object, although viewpoint-object semantics remain a relevant benchmark design factor.}

\begin{table}[t]
  \centering
  \resizebox{\columnwidth}{!}{%
  \begin{tabular}{lcc}
    \toprule
    \textbf{Model} & \textbf{Human (\%)} & \textbf{Camel (\%)} \\
    \midrule
    DINO & 57.12 & 54.62 \\
    DINO-v2 (B/14) & 63.80 & 58.69 \\
    DINO-v2 reg (B/14) & 63.14 & 60.88 \\
    DINO-v2 reg (L/14) & \underline{70.30} & \underline{69.76} \\
    DINOv3 & 67.45 & 64.16 \\
    VGGT (L/14) & \textbf{77.74} & \textbf{73.54} \\
    SPA & 63.90 & 47.25 \\
    CroCo & 60.97 & 59.94 \\
    CroCo v2 & 60.64 & 51.17 \\
    CLIP & 50.72 & 43.51 \\
    DeiT & 48.18 & 46.17 \\
    MAE & 64.55 & 61.82 \\
    MaskFeat & 61.05 & 59.47 \\
    \bottomrule
  \end{tabular}
  }
  \caption{\revision{\textbf{Allocentric experiment in the Desert environment with a non-human viewpoint object.} We compare the original human-viewpoint setting from the main benchmark with a control setting in which the human viewpoint object is replaced by a camel, while keeping the geometric protocol unchanged. Results are reported only for efficient probing. The model ranking remains broadly consistent, which suggests that the allocentric performance gap is not explained solely by the use of a human viewpoint object.}}
  \label{tab:camel_viewpoint_control}
\end{table}

\subsection{Impact of Input Resolution}
\label{sec:appendix_resolution_control}

\revision{
In the main benchmark, all models are evaluated at a standardized input resolution of 224$\times$224. This improves comparability across backbones, but it may underestimate the absolute capability of models whose native or recommended inference resolution is substantially higher, such as VGGT.}

\revision{To assess this effect, we conducted a control experiment in the Desert environment using efficient probing, reevaluating VGGT and DINO-v2 at a higher input resolution of 518$\times$518 and comparing the results with the standard 224$\times$224 setting. The results are reported in Table~\ref{tab:full-res-vggt-dinov2}. We observe that higher resolution mainly improves performance in the allocentric setting, while the camera-viewpoint task is already close to saturation. These results show that input resolution is an important quantitative factor, especially for the more difficult allocentric task. At the same time, the main qualitative conclusions remain stable. VGGT remains the stronger model in this control, and the allocentric task remains substantially more challenging than the egocentric one. We therefore interpret the main 224$\times$224 benchmark as a standardized comparison protocol, while viewing higher-resolution evaluation as a complementary measure of absolute model capability.}

\begin{table}[t]
  \centering
  \resizebox{\columnwidth}{!}{%
  \begin{tabular}{llcc}
    \toprule
    \textbf{Image res.} & \textbf{Model} & \textbf{Ego} & \textbf{Allo}  \\
    \midrule
    518$\times$518 & VGGT (L/14) & 98.60 & 81.41 \\
    518$\times$518 & DINO-v2 reg (L/14) & 97.65 & 79.97 \\
    224$\times$224 & VGGT (L/14) & 98.78 & 77.74 \\
    224$\times$224 & DINO-v2 reg (L/14) & 96.70 & 70.30 \\
    \bottomrule
  \end{tabular}%
  }
  \caption{\revision{\textbf{Impact of the input resolution.} We compare efficient probing performance at 518$\times$518 and 224$\times$224 for VGGT and DINO-v2 in the Desert environment. Higher resolution improves performance mainly in the allocentric (human-viewpoint) setting, while the egocentric (camera-viewpoint) task is already near saturation.}}
  \label{tab:full-res-vggt-dinov2}
\end{table}

\section{Extended Attention Visualizations}\label{sec:extend_attention}

\subsection{Attention Dynamics across Object Categories}
\begin{figure*}[h!]
    \centering
    \begin{subfigure}[b]{0.60\textwidth}
        \centering
        \includegraphics[width=\linewidth]{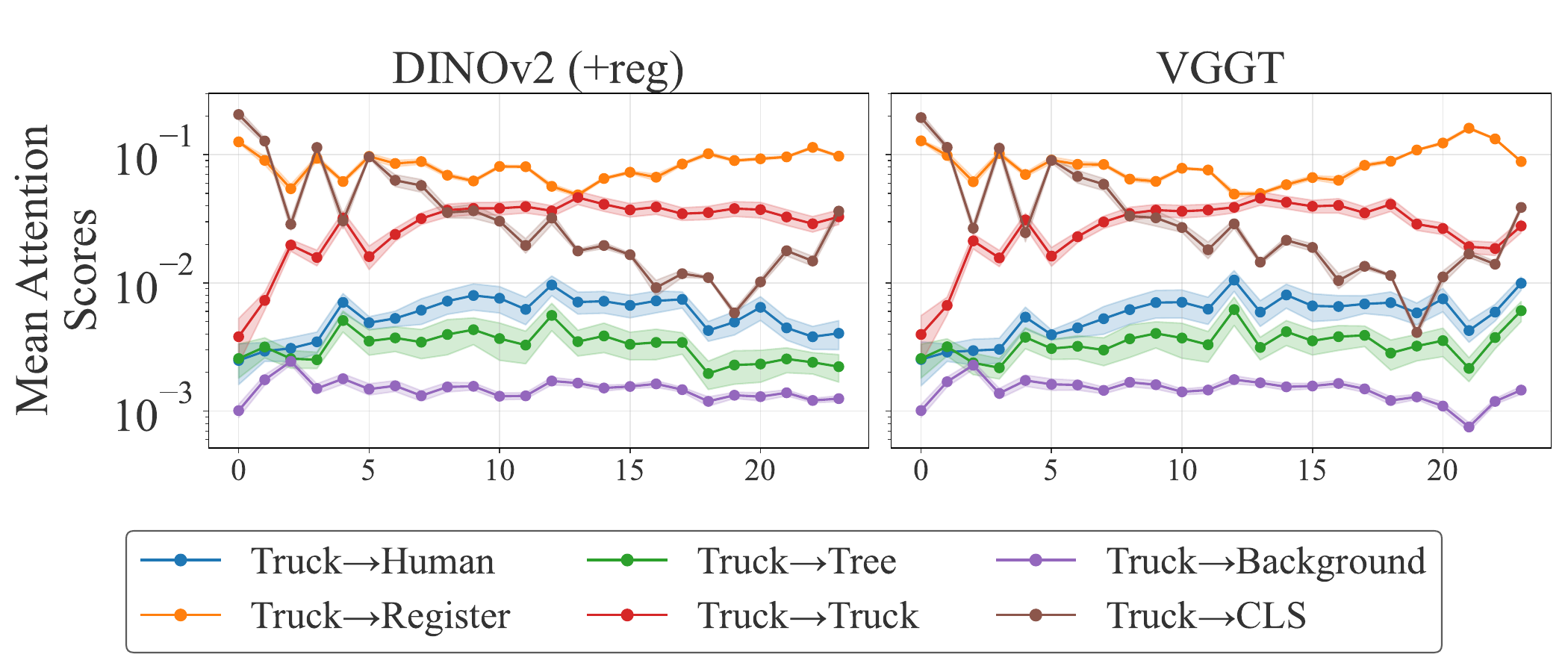}
        \label{fig:truck_agg}
    \end{subfigure}
    \begin{subfigure}[b]{0.365\textwidth}
        \centering
        \includegraphics[width=\linewidth]{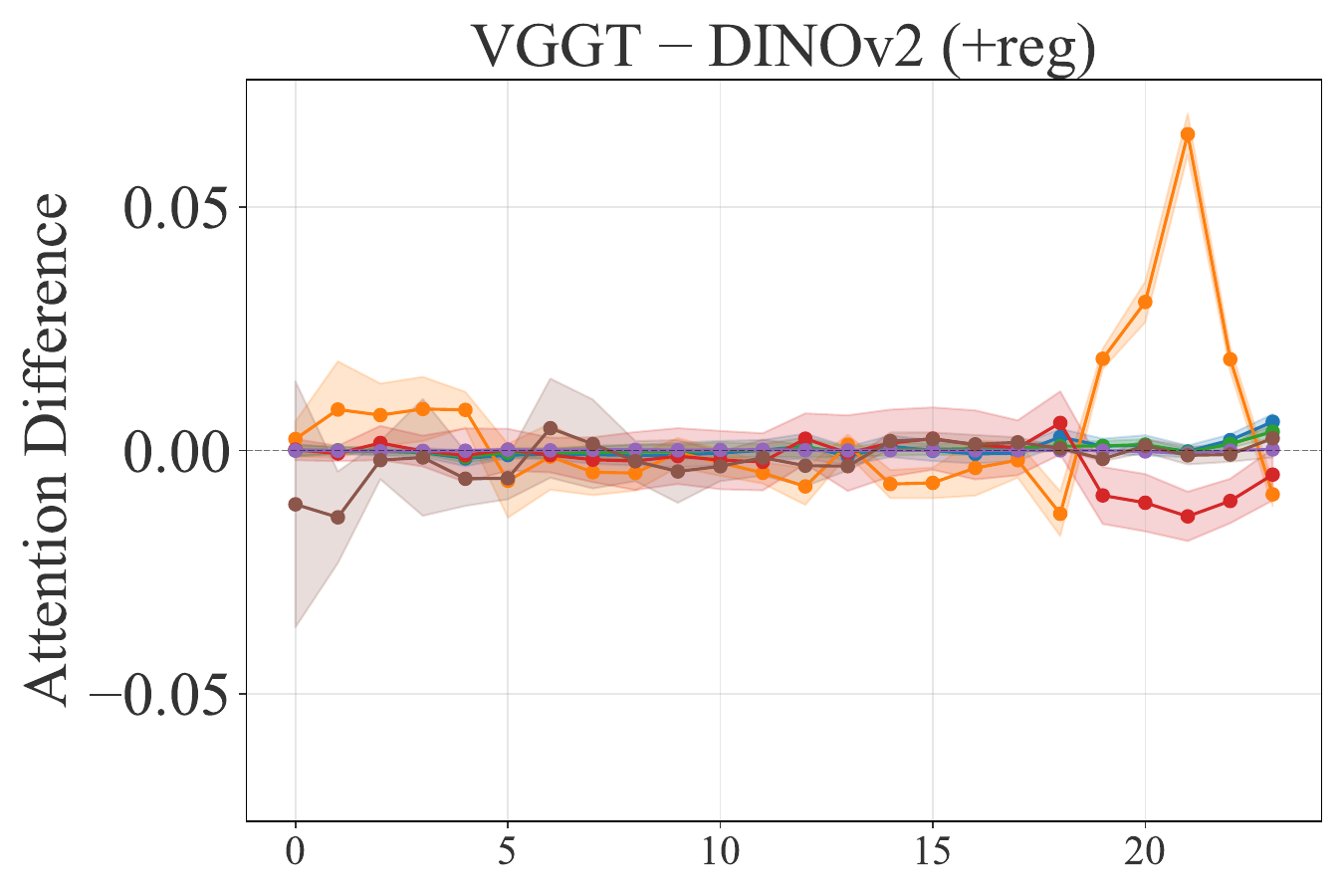}
        \label{fig:truck_diff}
    \end{subfigure}

    \vspace{0.2cm} 

    \begin{subfigure}[b]{0.60\textwidth}
        \centering
        \includegraphics[width=\linewidth]{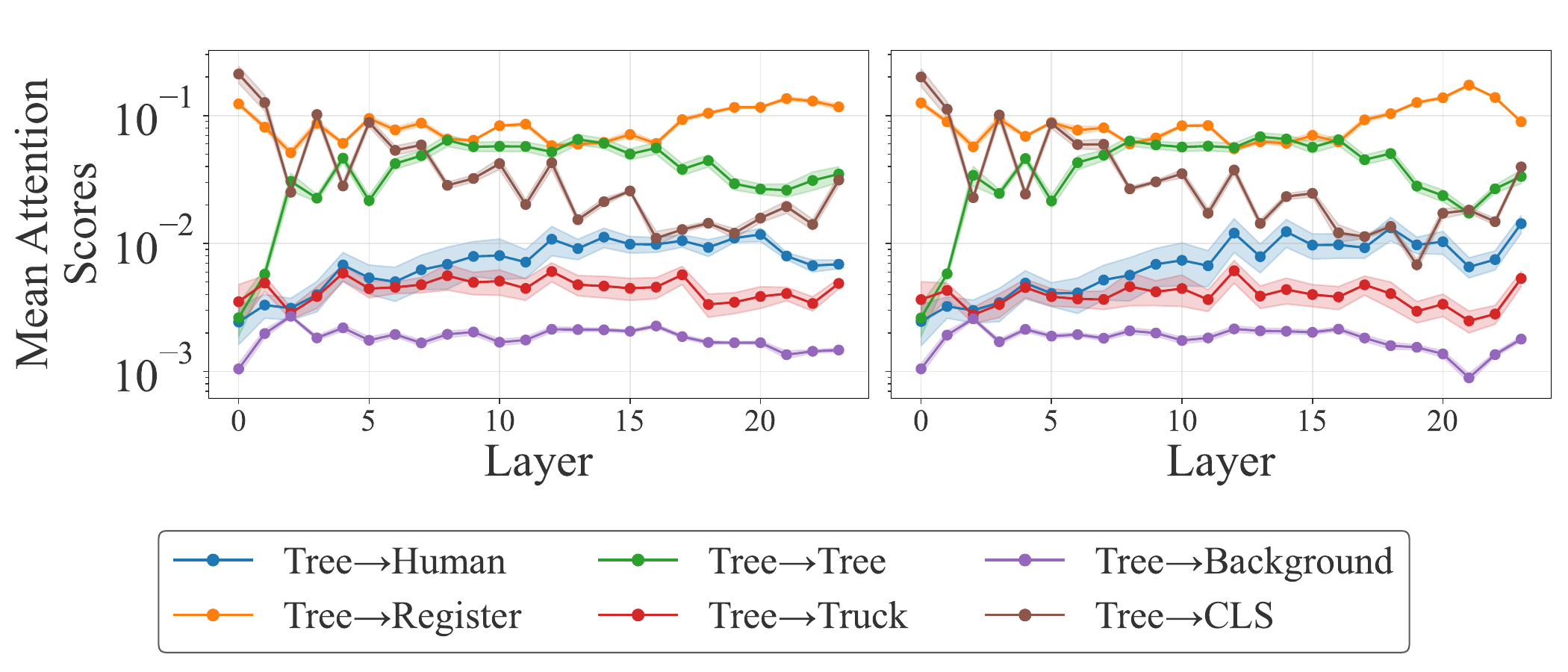}
        \label{fig:tree_agg}
    \end{subfigure}
    \begin{subfigure}[b]{0.365\textwidth}
        \centering
        \includegraphics[width=\linewidth]{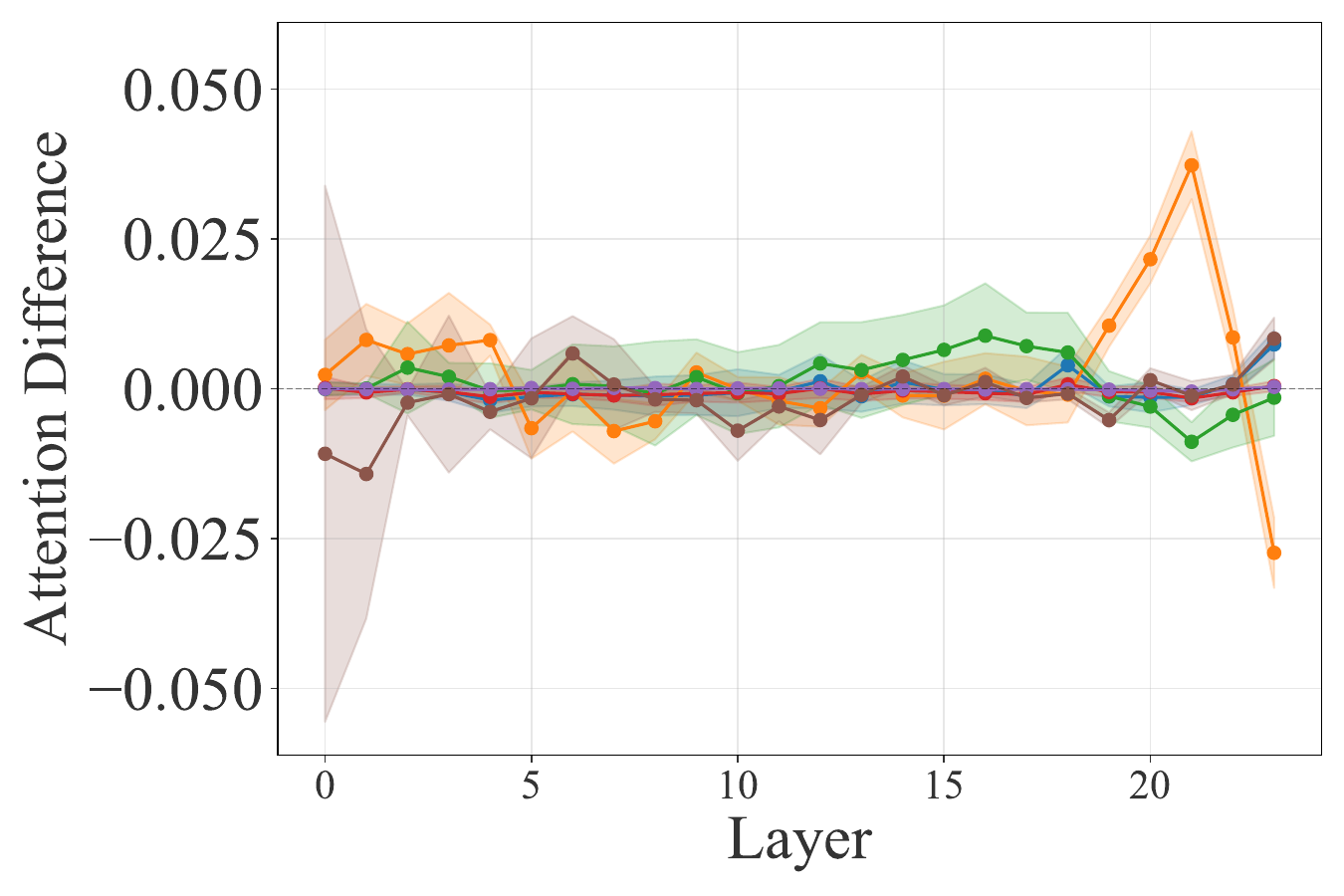}
        \label{fig:tree_diff}
    \end{subfigure}

    \caption{\textbf{Further Analysis of Attention Flow (DINO-v2 vs. VGGT).} We visualize the layer-wise mean attention scores from Truck (Top) and Tree (Bottom) tokens to the rest of the patches.}
    \label{fig:attention_other_objects}
\end{figure*}

In~\Cref{sec:analysis_attention}, we identified a relational reorganization of attention in the VGGT backbone, specifically in the global \cls{} token and the human patches. To verify that this mechanism is a general property rather than an artifact specific to the \cls{} token or human patches, we extend our analysis to other objects. \Cref{fig:attention_other_objects} visualizes the layer-wise attention dynamics for the Truck and Tree tokens. Consistent with the findings in~\Cref{sec:analysis_attention}, both object categories exhibit similar structural changes. As shown in~\Cref{fig:attention_other_objects} (right), VGGT consistently redirects the attention of object patches toward other scene entities (positive differences for Truck$\to$Human/Tree and negative differences for Truck$\to$Truck). This confirms that explicit 3D supervision forces the entire spatial representation to shift from object-centric feature extraction to relational scene encoding, regardless of the source token's semantic category.

\newpage
\subsection{Attention Maps of Efficient Probing}

\begin{figure}[t]
        \centering
        \includegraphics[width=\linewidth]{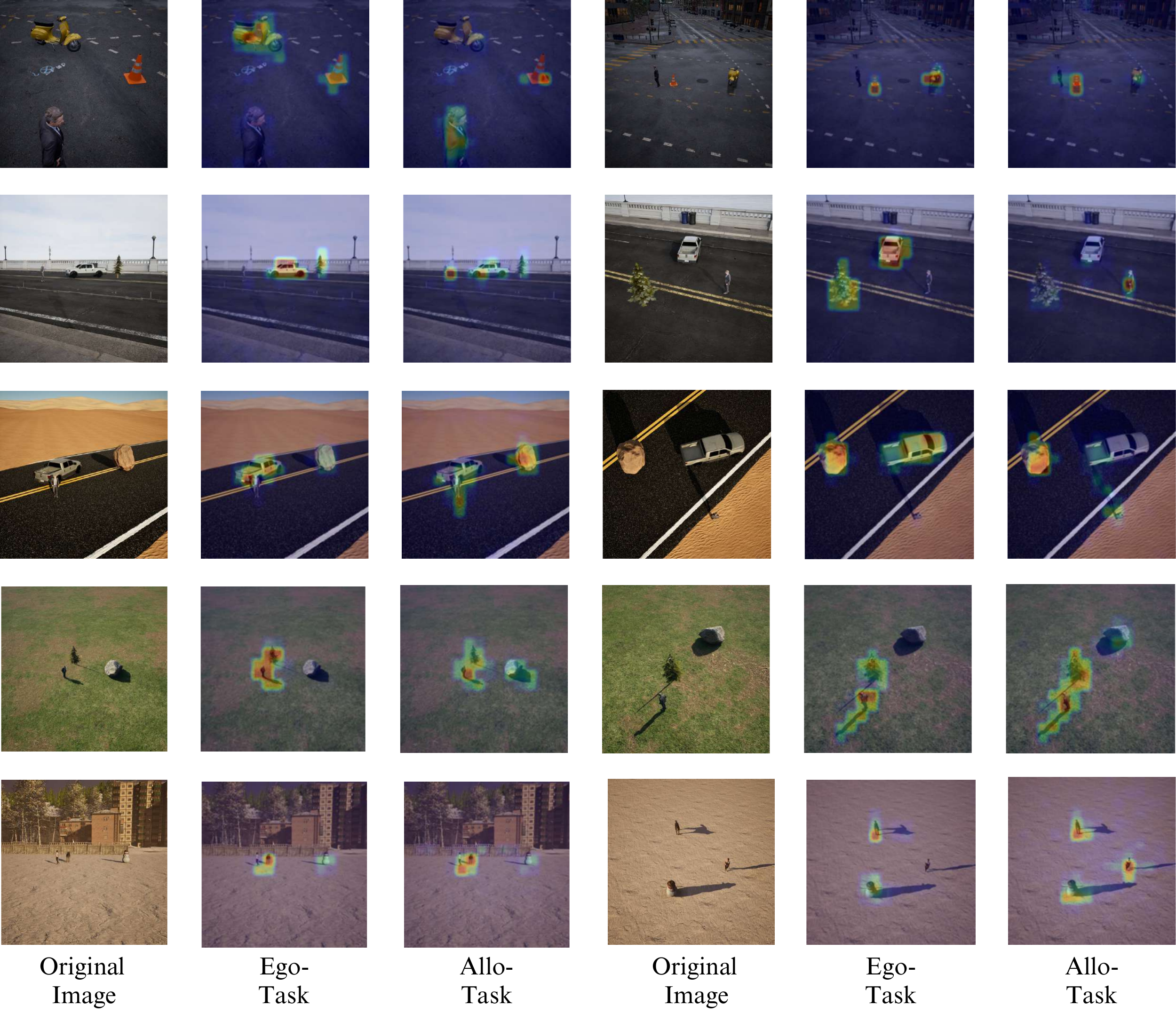}
        \caption{\textbf{Visualization of Task-Specific Attention Dynamics in Efficient Probing.} We visualize the mean attention maps (averaged across 4 learned queries) extracted from an efficient probing head attached to a frozen VGGT backbone.}
        \label{fig:EP_mean_attn}
\end{figure}

In~\Cref{fig:EP_mean_attn}, we visualize the mean attention maps generated by the efficient probing head attached to the frozen VGGT backbone to interpret how the model solves these spatial tasks. These visualizations show that the attention focus changes depending on the task. For the egocentric task (Columns 2 and 5), the model attends to the source and target objects while ignoring the other object. However, when the allocentric task (Columns 3 and 6) is solved, the model attends to the non-camera-viewpoint object, the source, and the target. This dynamic behavior confirms that the probe does not merely memorize texture but actively identifies and disentangles the specific geometric entities, viewpoint, source, and target, required to resolve the spatial query in the correct coordinate frame.

\begin{figure}[t]
        \centering
        \includegraphics[width=0.7\linewidth]{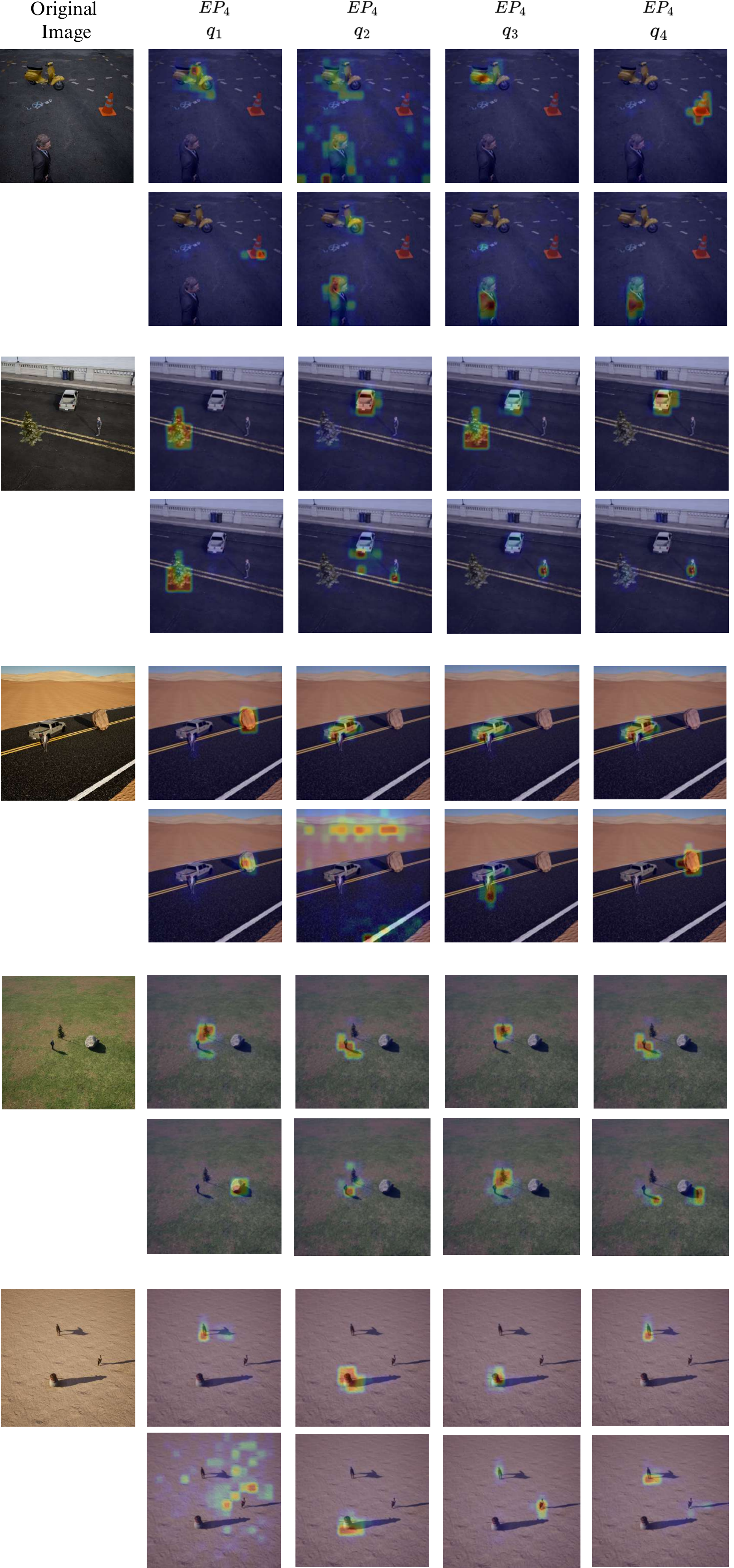}
        \caption{\textbf{Visualizing each head of multi-query attention.} We visualize the attention maps of the four learnable queries from an efficient probing head attached to a frozen VGGT backbone. In each environment, the first image in the top row shows the original image. The first row shows the attention maps for the four queries ($q_1 \dots q_4$) solving the egocentric tasks, and the second row shows the attention maps for the allocentric tasks.}
        \label{fig:EP_each_attn}
\end{figure}

In~\Cref{fig:EP_each_attn}, we visualize the individual attention maps of the four learnable queries ($q_1 \dots q_4$), to understand how efficient probing solves \revision{spatial relation recognition}. This breakdown reveals that rather than collapsing the scene into a single coarse, global point, the queries diversify to attend to distinct geometric entities. In the egocentric task, we observe that queries tend to split their focus between the source and target objects. Crucially, in the allocentric task, queries often split their attention between the non-camera viewpoint object and other objects. This behavior confirms that the multi-query architecture forces the model to decompose the scene into its functional components, viewpoint, source, and target, providing the downstream classifier with a structured, part-based representation of the geometry.

\newpage
\section{Layer-wise Probing: Where is Spatial Information Encoded?}\label{sec:layer-wise}
To understand how \revision{spatial relation recognition} emerges and evolves throughout the network depth, we evaluate the performance of intermediate layers as frozen features across both egocentric and allocentric tasks. For this analysis, we use VGGT and DINO-v2 (with register tokens), training an efficient probing head on the outputs of individual transformer blocks. To ensure that the observed trends reflect generalizable spatial skills rather than overfitting to specific visual domains, all reported results are averaged across the five diverse environments.

\begin{figure}[t]
        \centering
        \includegraphics[width=\linewidth]{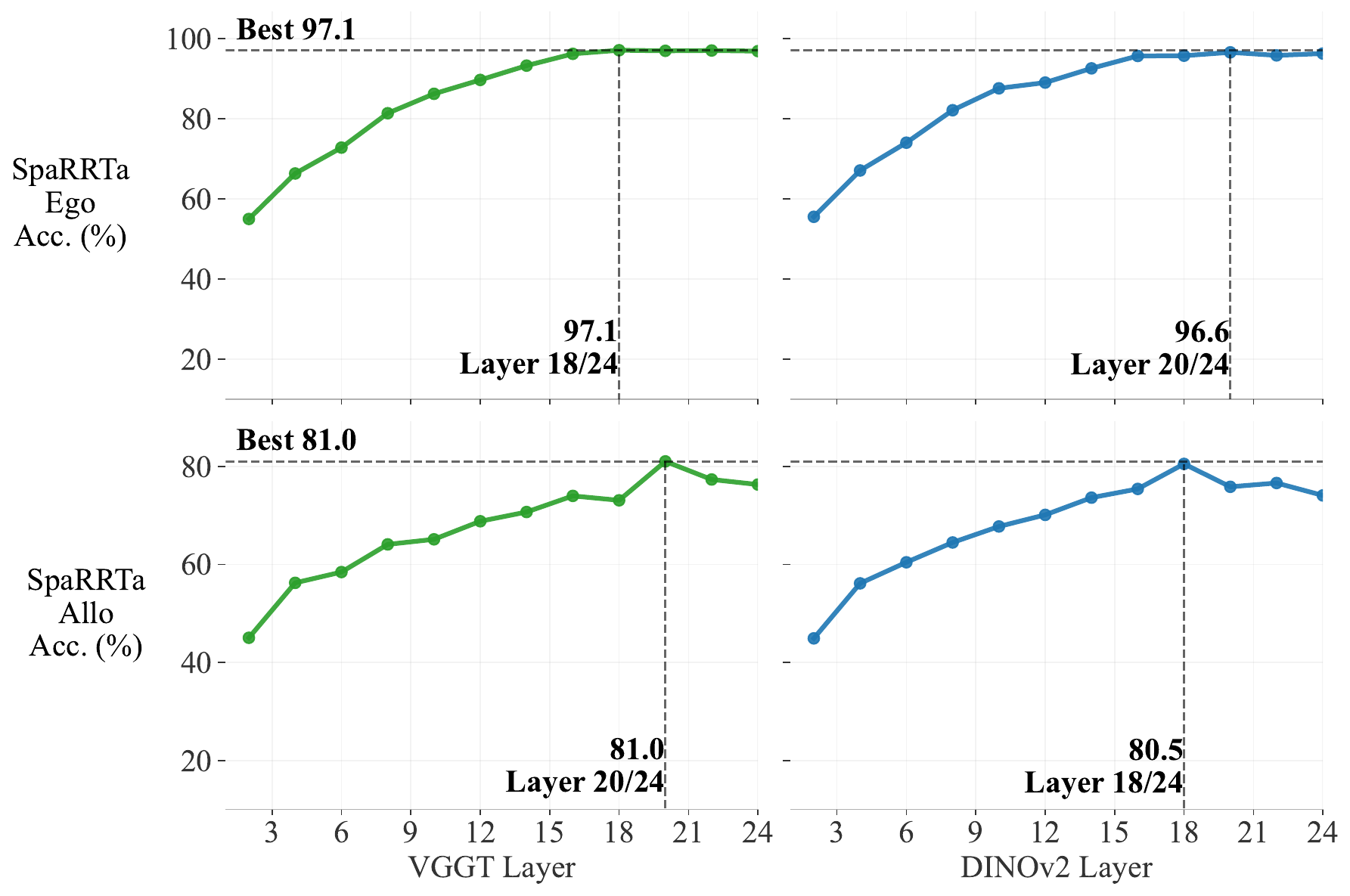}
        \caption{\textbf{Layer Analysis.} We evaluate the \revision{spatial relation recognition} capability of intermediate features extracted from frozen VGGT and DINO-v2 (with register tokens) ViT-Large backbones using efficient probing. Results are averaged across all five evaluation environments to ensure robustness against domain-specific bias. Vertical dashed lines denote the layer achieving peak performance.}
        \label{fig:layers}
\end{figure}

As illustrated in~\Cref{fig:layers}, we observe a consistent trend across both model architectures and task types: \revision{spatial relation recognition} accuracy improves steadily through the initial layers, plateaus, and eventually peaks in the late-intermediate layers (specifically Layers 18 and 20 for a 24-layer ViT-L), rather than at the final output layer. This behavior aligns with findings from the Perception Encoder~\cite{bolya_perception_2025} and recent architectural analyses in DINOv3~\cite{siméoni2025dinov3}. In standard VFMs, the final layers are often optimized for high-level semantic abstraction and invariance (e.g., classifying a "dog" regardless of its position). This objective implicitly encourages the model to discard precise geometric details in the final projection. Our results confirm that for tasks where geometry plays a governing role, such as the allocentric perspective-taking challenge, the most robust representations are located deeper in the network stack, before the last layer.

While the performance drop-off at the final layer is relatively minor, which makes it still a viable default choice for general use, these results suggest that downstream embodied agents requiring precise spatial manipulation or reasoning can achieve optimal performance by tapping into these late-intermediate layers. Specifically, extracting features from Layers 18--20 for a ViT-Large yields a measurable improvement in accuracy, providing a free performance boost without additional training cost.

\end{document}